\documentclass[a4paper,fleqn]{cas-sc}

\usepackage{enumitem}
\usepackage[authoryear]{natbib}
\usepackage[linesnumbered,ruled,vlined]{algorithm2e}
\usepackage{lineno}
\usepackage{float}
\usepackage{graphicx}
\usepackage{booktabs}
\usepackage{diagbox}
\usepackage{multirow}
\usepackage{threeparttable}
\usepackage{color,xcolor}
\usepackage{url}
\usepackage{microtype}
\usepackage{subfigure}
\usepackage{makecell}
\usepackage{bbm}
\usepackage{amssymb}
\usepackage{amsmath}
\usepackage{mathtools}
\usepackage{cases}
\usepackage{adjustbox}
\usepackage{pifont}
\usepackage{placeins}

\hypersetup{
    hypertex=true,
    colorlinks=true,
    linkcolor=red,
    anchorcolor=blue,
    urlcolor=magenta,
    filecolor=pink}

\newtheorem{definition}{Definition}[section]

\def\tsc#1{\csdef{#1}{\textsc{\lowercase{#1}}\xspace}}
\tsc{WGM}
\tsc{QE}
\tsc{EP}
\tsc{PMS}
\tsc{BEC}
\tsc{DE}

\begin{document}

\let\WriteBookmarks\relax
\def\floatpagepagefraction{1}
\def\textpagefraction{.001}

\shorttitle{Minimum Description Length based Granular-Ball Tree Regularization for Spectral Clustering}
\shortauthors{Xian et al.}

\title[mode=title]{Minimum Description Length based Granular-Ball Tree Regularization for Spectral Clustering}

\author[1,2]{Zeqiang Xian}
\ead{xianzeqiang@gnnu.edu.cn}

\author[1,2]{Caihui Liu}
\cormark[1]
\ead{liucaihui@gnnu.edu.cn}

\author[1,2]{Yong Zhang}
\ead{zhang_yong@gnnu.edu.cn}

\author[1,2]{Wenjing Qiu}
\ead{wenjingqiu@gnnu.edu.cn}

\affiliation[1]{
	organization={Department of Mathematics and Computer Science, Gannan Normal University},
	city={Ganzhou},
	postcode={341000},
	state={Jiangxi},
	country={China}
}

\affiliation[2]{
	organization={Key Laboratory of Data Science and Artificial Intelligence of Jiangxi Education Institutes, Gannan Normal University},
	city={Ganzhou},
	postcode={341000},
	state={Jiangxi},
	country={China}
}

\cortext[cor1]{Corresponding author}

\begin{abstract}
	This paper proposes Minimum Description Length based Granular-Ball Tree Regularized Spectral Clustering (MDL-GBTRSC), a fixed-$K$ method that uses granular balls as regional descriptors for sample-level affinity construction rather than as reduced graph nodes. A reciprocal-neighborhood continuity term is incorporated into an MDL-based refinement criterion to construct a granular-ball hierarchy, and the structural scales of the terminal regions are transferred to the original sample graph before spectral partitioning. Experiments on 18 real and 20 synthetic datasets, comprising 109,969 samples in total, compare MDL-GBTRSC with nine competing methods. On the 17 real datasets completed by all methods, MDL-GBTRSC achieves mean ARI and NMI values of $0.5998$ and $0.6613$, compared with the next-highest means of $0.5792$ and $0.6285$, respectively. On the 20 synthetic datasets, it achieves mean ARI and NMI values of $0.9618$ and $0.9688$, compared with the next-highest means of $0.7921$ and $0.8306$, and is best or tied on 15 datasets for both metrics. These results indicate that MDL-selected regional structure provides useful local information for affinity construction while preserving the resolution of the original spectral graph. Source code and experimental materials are available at \href{https://github.com/AnonymousUser0816/MDL-GBTRSC}{https://github.com/AnonymousUser0816/MDL-GBTRSC}.
\end{abstract}


\maketitle

\section{Introduction}
\label{sec:introduction}

Cluster analysis is a fundamental task in unsupervised learning and information processing. It aims to organize unlabeled observations into groups so that samples within the same group share stronger similarity than samples from different groups~\citep{hu2026interpretable}. Classical clustering methods, such as $k$-means~\citep{McQueen1967KMeans} and hierarchical clustering~\citep{Johnson1967Hierarchical}, have been widely used because of their simple formulations and broad applicability. However, the effectiveness of these methods is often affected by assumptions about cluster compactness, distance-based similarity, or hierarchical separability. Such assumptions may not hold when the data contain non-convex structures, heterogeneous local densities, or weak boundaries between neighboring groups.

Spectral clustering provides a graph-based formulation for handling such data structures. It constructs an affinity graph over samples, derives a spectral embedding from the graph Laplacian, and then obtains the final partition in the embedded space~\citep{ng2001spectral,von2007tutorial,Ding2024SurveySC}. In this pipeline, the affinity graph is not only an intermediate data representation. It determines how local sample relations are encoded before spectral embedding and therefore directly affects the subsequent partition. Classical $k$-nearest-neighbor graphs and RBF graphs encode sample relations through a predefined neighborhood order or kernel scale. These constructions are straightforward, but a fixed graph scale may be inadequate when local density and geometric structure vary across the data space. A graph that misses important within-cluster links may fragment a true cluster, whereas a graph that retains inappropriate cross-cluster links may blur cluster boundaries.

Recent spectral clustering studies have attempted to address graph construction from different perspectives. Direct spectral clustering~\citep{Kong2025DirectSC} reduces the separation between graph construction and clustering. Structured doubly stochastic graph-based clustering~\citep{Wang2025SDSGC} introduces structural constraints into graph learning. Self-constrained spectral clustering~\citep{SelfConstrainedSC} and stacked adaptive graph learning~\citep{Li2024SCnetAGL} further show that affinity relations can be updated during clustering rather than being fixed in advance. These studies improve the adaptability of spectral graphs. However, when local continuity and boundary-crossing relations are difficult to distinguish, the learned graph still depends on the structural evidence extracted from local neighborhoods.

Another line of research focuses on the computational cost of sample-level spectral clustering. Since traditional spectral clustering requires an affinity matrix over all samples and eigendecomposition of the graph Laplacian, its direct application to large-scale datasets can be expensive. Spectral clustering with linear embedding~\citep{Gao2024SCLE}, approximate spectral embedding representation~\citep{Zhou2025ASER}, restarted block-diagonal spectral clustering~\citep{RestartedLargeScaleSC}, and RESKM~\citep{RESKM} reduce the cost of spectral computation through approximation or embedding strategies. Anchor- and bipartite-graph-based methods reduce graph size by introducing representative structures, as in self-adapted bipartite graph learning~\citep{Yang2023FSBGL}, structured optimal bipartite graph clustering~\citep{StructuredOptimalBipartiteGraph}, discrete and balanced spectral clustering~\citep{Wang2023DBSC}, fast co-clustering with adaptive anchor graph learning~\citep{FastDirectCoClusteringAAGL}, and GMM-enhanced anchor-based spectral clustering~\citep{Zhang2025GMMSC}. These methods improve scalability, but the reduced graph remains dependent on whether the representatives preserve local geometry and boundary information.

The above studies indicate that both graph learning and graph reduction require an appropriate description of local data structure. Several methods refine spectral graphs by introducing local structural information. Subspace-distance based spectral clustering uses local subspace relations in similarity estimation~\citep{Naseri2025SMDS}. Diffusion-based spectral clustering updates affinities through neighborhood propagation~\citep{Zhu2025DPSC}. Stratified multi-density spectral clustering incorporates density heterogeneity into graph construction~\citep{Yue2023SMDSC}. Refined $k$NN graph construction~\citep{RefinedKNNGraphSC}, split--merge strategies on $k$NN graphs~\citep{SplitMergeKNNGraph}, and three-way-decision based spectral clustering~\citep{ThreeWayDecisionSC} further handle uncertain local relations before spectral embedding. These methods enrich graph construction beyond raw pairwise distances, but most of them still operate mainly at the edge or neighborhood level. The structural properties of local regions, including coverage, scale, dispersion, and consistency, are less directly incorporated into the original sample-level affinity graph.

Granular-ball computing provides a region-level representation for local data organization. A granular ball is described by the samples it covers, together with its center and radius~\citep{Xia2024GBCSurvey,Xia2019_GBClassifiers}. This representation has been adopted in clustering and spectral clustering. GBCT models clustering structures through relations among granular balls~\citep{Xia2025GBCT}. Granular-ball spectral clustering constructs similarities on generated balls rather than on all original samples~\citep{Xie2023GBSC}. Granular-ball manifold clustering~\citep{Cheng2024GBUSC} and pseudo-label based granular-ball division~\citep{Cheng2025FSCPLGB} use granular balls as adaptive representatives. Micro-cluster based spectral clustering organizes local density and manifold information through region-level units~\citep{Xu2025MDMSC}. These methods show that local regions can serve as useful computational units. In most existing granular-ball or micro-cluster based spectral clustering methods, however, the generated regions mainly serve as reduced graph nodes, anchors, or intermediate representatives. Their internal structural information is not systematically transferred back to the original sample-level affinity graph.

The criterion used to generate local regions further affects how these regions can support graph construction. Compactness, density, pseudo-label purity, and heuristic splitting rules are commonly used in granular-ball generation. These criteria may produce meaningful regions, but they do not always provide a unified model-selection view for deciding whether a region should be retained or refined. The Minimum Description Length (MDL) principle compares alternative descriptions according to their total code lengths and has been used to formulate granular-ball generation as a local model-selection problem~\citep{Xian2026MDLGBG}. Boundary-aware MDL granular-ball classification extends this description-length view to supervised boundary modeling~\citep{xian2026MDLGBC}. For spectral clustering, region refinement should additionally consider graph continuity, because a split that improves local geometric description may separate reciprocal neighborhood relations that are important for the spectral graph.

\subsection{Research Objective and Main Contributions}
\label{subsec:objective_contributions}

This work studies how adaptive region-level structure can be incorporated into a sample-level spectral graph without replacing the original observations by anchors or coarse representatives. To this end, we propose Minimum Description Length based Granular-Ball Tree Regularized Spectral Clustering (MDL-GBTRSC), which learns a granular-ball hierarchy under reciprocal-neighborhood continuity and transfers the structural scales of its terminal regions to the affinities among individual samples.

The main contributions are summarized as follows:
\begin{itemize}
	\item A sample-level spectral clustering framework is developed in which granular-ball regions provide local structural information rather than serving as graph nodes or anchors.
	\item A graph-regularized description-length criterion combines local geometric fit with reciprocal-neighborhood continuity to determine granular-ball refinement.
	\item The coverage and dispersion of terminal granular balls are converted into local structural scales and transferred to the sample-level affinity graph.
	\item Experiments on real and synthetic benchmarks evaluate the resulting sample-level graph and examine the roles of graph continuity, leaf-scale regularization, graph connectivity, and computational scale.
\end{itemize}

The remainder of this paper is organized as follows. Section~\ref{sec:related_work} reviews related work. Section~\ref{sec:preliminaries} introduces the notation and background used by the method. Section~\ref{sec:methodology} presents MDL-GBTRSC, followed by complexity analysis in Section~\ref{sec:complexity}. Section~\ref{sec:experiments} reports the experiments, Section~\ref{sec:discussion} discusses the main findings and limitations, and Section~\ref{sec:conclusion} concludes the paper.

\section{Related Work}
\label{sec:related_work}

Research related to this work can be broadly understood from three perspectives according to how local structure is incorporated into spectral clustering: adapting relations among individual samples, reducing the spectral graph through representative structures, and explicitly modeling local regions. These directions provide different ways of describing local geometry and are reviewed below.

\subsection{Adaptive Graph Learning for Spectral Clustering}
\label{subsec:related_adaptive_graph}

The quality of spectral clustering depends strongly on whether the affinity graph reflects the underlying geometry of the data. Conventional $k$-nearest-neighbor and kernel graphs construct affinities according to predefined neighborhood orders or similarity scales, whereas adaptive graph-learning methods estimate or refine these relations from the data. Direct spectral clustering~\citep{Kong2025DirectSC} couples graph construction with cluster assignment, and structured doubly stochastic graph-based clustering~\citep{Wang2025SDSGC} introduces structural constraints into affinity learning. Self-constrained spectral clustering~\citep{SelfConstrainedSC} uses clustering information to regulate graph refinement, while stacked adaptive graph learning~\citep{Li2024SCnetAGL} progressively updates graph relations through multiple representation stages. Although their optimization formulations differ, these methods share the objective of adapting the sample-level graph to the structural information revealed during clustering.

Local geometry can also be incorporated through explicit modification of neighborhood relations or pairwise similarities. Subspace-distance based spectral clustering measures affinities in locally estimated subspaces~\citep{Naseri2025SMDS}, diffusion-based spectral clustering propagates neighborhood information to refine sample relations~\citep{Zhu2025DPSC}, and stratified multi-density spectral clustering adapts graph construction to heterogeneous density structures~\citep{Yue2023SMDSC}. Refined $k$NN graphs~\citep{RefinedKNNGraphSC}, split--merge neighborhood strategies~\citep{SplitMergeKNNGraph}, and three-way-decision based spectral clustering~\citep{ThreeWayDecisionSC} further modify uncertain or ambiguous local connections before spectral embedding. These methods enrich the affinity graph through sample-level or neighborhood-level evidence, with local structure primarily represented through the resulting pairwise relations.

\subsection{Representative-Based and Large-Scale Spectral Clustering}
\label{subsec:related_scalable_sc}

The computational cost of spectral clustering has motivated a parallel line of research on approximate and structured spectral formulations. Spectral clustering with linear embedding~\citep{Gao2024SCLE} reduces the cost of obtaining cluster representations through a linearized formulation, while approximate spectral embedding representation~\citep{Zhou2025ASER} estimates spectral information from a compact representation. Restarted block-diagonal spectral clustering~\citep{RestartedLargeScaleSC} and RESKM~\citep{RESKM} exploit structured or approximate computations to reduce the cost associated with conventional eigendecomposition. These approaches mainly improve the computational realization of spectral clustering while preserving sample-level relations as an important source of structural information.

Another widely used strategy is to construct the spectral problem around a smaller set of representative points or anchors. Self-adapted bipartite graph learning~\citep{Yang2023FSBGL} and structured optimal bipartite graph clustering~\citep{StructuredOptimalBipartiteGraph} model relations between samples and representative nodes through bipartite structures. Discrete and balanced spectral clustering~\citep{Wang2023DBSC} combines representative graph learning with discrete assignment and balance constraints. Fast co-clustering with adaptive anchor graph learning~\citep{FastDirectCoClusteringAAGL} and GMM-enhanced anchor-based spectral clustering~\citep{Zhang2025GMMSC} further construct adaptive representative structures for efficient clustering. By reducing the number of graph nodes involved in the spectral problem, these methods provide an effective route to large-scale clustering, while the quality of the resulting representation depends on how faithfully the selected representatives preserve local geometry and connectivity.

\subsection{Granular and Region-Based Structure Modeling}
\label{subsec:related_gbc_mdl}

Granular-ball computing provides a region-level representation in which a subset of observations is described by its covered samples together with geometric quantities such as its center and radius~\citep{Xia2024GBCSurvey,Xia2019_GBClassifiers}. The representation is adaptive because relatively homogeneous regions can remain coarse, whereas more complex regions can be progressively refined. In clustering, GBCT models relations among granular balls to recover cluster structures~\citep{Xia2025GBCT}. Granular-ball spectral clustering constructs similarities among generated balls before spectral partitioning~\citep{Xie2023GBSC}, while granular-ball manifold clustering~\citep{Cheng2024GBUSC} and pseudo-label based granular-ball division~\citep{Cheng2025FSCPLGB} employ granular balls as adaptive representative units. Related micro-cluster based spectral clustering methods also summarize local density and manifold information through region-level structures~\citep{Xu2025MDMSC}. These studies show that adaptive regions provide an explicit way of describing local structure beyond individual pairwise relations.

The construction of such regions requires a principle for determining whether a local subset should remain intact or be further refined. Existing granular-ball and micro-cluster methods employ criteria based on compactness, density, purity, neighborhood consistency, or related structural quantities. Description-length-based granular-ball learning provides a model-selection perspective in which alternative local representations are compared according to their coding costs. MDL-based granular-ball generation applies this principle to unsupervised region construction~\citep{Xian2026MDLGBG}, while boundary-aware MDL granular-ball classification extends the same general idea to supervised boundary modeling~\citep{xian2026MDLGBC}. Under this perspective, retain and split representations of the same observations can be evaluated within a common model-selection objective.

\subsection{Discussion and Research Motivation}
\label{subsec:discussion_motivation}

The studies reviewed above improve spectral clustering from complementary perspectives, including adaptive affinity learning, representative-based graph construction, and region-level structure modeling. These approaches differ mainly in how local structure enters the clustering process: it may be encoded directly in pairwise affinities, summarized by representative nodes, or described through adaptive local regions. Comparatively less attention has been given to using adaptive regions as structural descriptors while retaining the original samples as vertices of the spectral graph. This perspective motivates the formulation considered in this work, which connects region-level structural modeling with sample-level affinity construction without replacing the original observations by coarse representatives.

\section{Preliminaries}
\label{sec:preliminaries}

\subsection{Problem Setting and Notation}
Let $X=[x_1,\ldots,x_n]^\top\in\mathbb R^{n\times d}$ denote an unlabeled dataset with $n$ observations and $d$ features. Given the prescribed number of clusters $K$, the clustering task is to obtain a partition $\Pi=\{C_1,\ldots,C_K\}$ of the index set $V=\{1,\ldots,n\}$. Throughout the paper, $W=[w_{ij}]$ denotes a sample-level affinity matrix, $D=\operatorname{diag}(d_1,\ldots,d_n)$ with $d_i=\sum_jw_{ij}$ is the degree matrix, and $G=(V,E)$ is the graph induced by the nonzero affinities.

\subsection{Granular-Ball Representation and Divisive Construction}
For a nonempty subset $S\subseteq V$, the corresponding granular ball is denoted by $B(S)=(X_S,c_S,r_S)$, where $X_S=\{x_i:i\in S\}$,
\begin{equation}
	c_S=\frac{1}{|S|}\sum_{i\in S}x_i,\qquad
		r_S=\max_{i\in S}\|x_i-c_S\|_2.
	\label{eq:prelim_ball}
\end{equation}
Its average within-ball dispersion is
\begin{equation}
	\sigma_S^2=\frac{1}{|S|d}\sum_{i\in S}\|x_i-c_S\|_2^2.
	\label{eq:prelim_dispersion}
\end{equation}
A granular-ball representation of the data consists of a collection $\mathcal B=\{B(S_b)\}_{b=1}^{m}$ whose covered index sets form a disjoint partition of $V$, i.e., $V=\dot\bigcup_{b=1}^{m}S_b$. Divisive granular-ball construction starts from the root region $B(V)$. At each stage, an active subset $S$ is either retained or replaced by disjoint child subsets $S_L$ and $S_R$ satisfying $S=S_L\dot\cup S_R$. Repeating this operation produces a rooted hierarchy, and the terminal subsets define the final granular-ball collection. Different granular-ball methods mainly differ in the criterion used to evaluate a region and in the rule used to generate its candidate children.

\subsection{Local Description-Length Model Selection}
Description length provides a model-selection criterion for comparing alternative representations of the same observations. For a local sample set $X_S$ and a candidate model $M$, a two-part description has the generic form
\begin{equation}
L(X_S,M)= L(M)+ L(X_S\mid M),
	\label{eq:prelim_mdl}
\end{equation}
where the two terms describe model complexity and residual data, respectively. Alternative representations of the same local sample set can therefore be compared through their total description lengths. This principle provides the model-selection basis for the retain-or-refine decision introduced in Section~\ref{subsec:graph_regularized_tree_induction}.

\subsection{Spectral Clustering and Graph Connectivity}
For a nonnegative symmetric affinity matrix $W$, normalized spectral clustering is based on the symmetric normalized Laplacian~\citep{ng2001spectral,von2007tutorial}
\begin{equation}
	L_{\mathrm{sym}}=I-D^{-1/2}WD^{-1/2}.
	\label{eq:prelim_laplacian}
\end{equation}
The eigenvectors associated with the smallest eigenvalues of $L_{\mathrm{sym}}$ define a spectral representation of the samples from which a discrete $K$-way partition is obtained. The multiplicity of the zero eigenvalue equals the number of connected components of the underlying graph. Thus, connected components identify groups already separated by graph topology, whereas a connected component containing multiple clusters requires further spectral subdivision.

\section{Methodology}
\label{sec:methodology}
\begin{figure}
	\centering
	\includegraphics[width=\columnwidth]{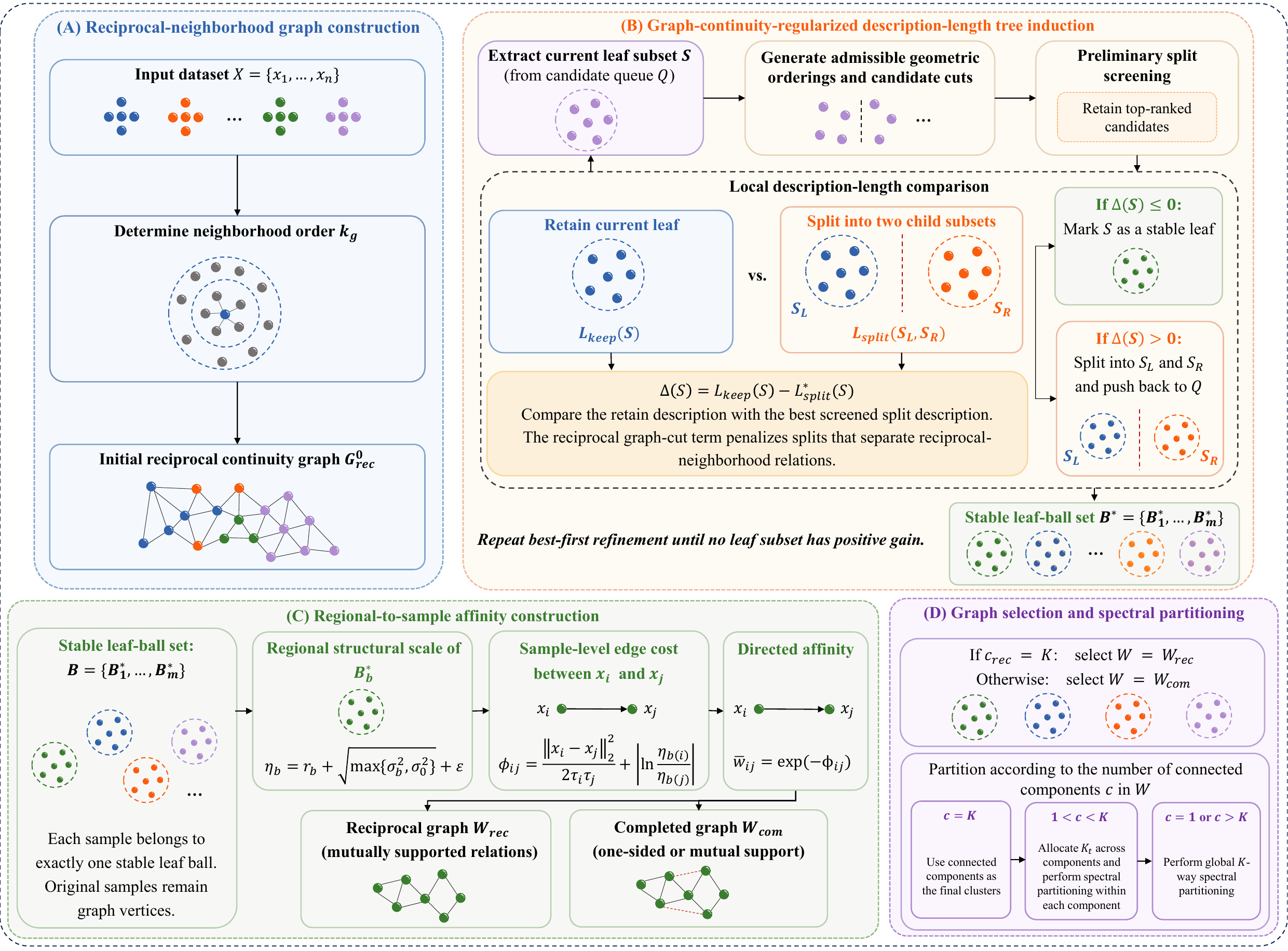}
	\caption{Framework of MDL-GBTRSC.
	(A) Reciprocal-neighborhood relations provide local continuity information for subsequent refinement.
	(B) Admissible geometric cuts are screened before the retained candidates are evaluated by the graph-continuity-regularized local description-length criterion, producing the stable leaf-ball set.
	(C) Regional structural scales from the stable leaf balls are transferred to affinities among the original samples, from which reciprocal and completed graphs are constructed.
	(D) The final graph is selected using the prescribed cluster number $K$ and partitioned according to its connected-component state by direct assignment, component-wise spectral subdivision, or global spectral partitioning.
	}
	\label{fig:framework}
\end{figure}

MDL-GBTRSC constructs a granular-ball hierarchy on the original sample set and transfers the structural scales of its terminal regions to a sample-level affinity graph. The method consists of four stages: reciprocal-neighborhood graph construction, graph-continuity-regularized description-length refinement, regional-to-sample affinity construction, and connectivity-aware spectral partitioning. Figure~\ref{fig:framework} summarizes the resulting data flow.

\subsection{Reciprocal Continuity Graph}
\label{subsec:reciprocal_continuity_graph}
For each sample $x_i$, let $N_g(i)$ denote its $k_g$-nearest-neighbor set, where
\begin{equation}
		k_g=\min\left\{n-1,\max\left(2,\left\lceil\frac{\ln(n+1)}{\ln2}+\sqrt d\right\rceil\right)\right\}.
	\label{eq:graph_order}
\end{equation}
Let $\delta_{i,k_g}$ be the distance from $x_i$ to its $k_g$-th nearest neighbor and define $\tau_i=\max\{\delta_{i,k_g},\varepsilon_{\rm mach}\}$. For $j\in N_g(i)$, the directed self-tuning affinity is
\begin{equation}
	\bar a_{ij}
	=
	\exp\left(-\frac{\|x_i-x_j\|_2^2}{2\tau_i\tau_j}\right).
	\label{eq:base_affinity}
\end{equation}
Before logarithmic coding, affinities are restricted to the interval $[\varepsilon_{+},1-\varepsilon_{\rm mach}]$, where $\varepsilon_{+}$ denotes a positive machine-level lower bound. The reciprocal affinity is
\begin{equation}
	a_{ij}^{\mathrm{rec}}
	=
	\begin{cases}
		\min\{\bar a_{ij},\bar a_{ji}\}, & j\in N_g(i),\ i\in N_g(j),\\
		0, & \text{otherwise}.
	\end{cases}
	\label{eq:reciprocal_affinity}
\end{equation}
The corresponding undirected graph is denoted by $G_{\mathrm{rec}}^0=(V,E_{\mathrm{rec}}^0)$. Because an edge is retained only when the neighborhood relation is supported from both endpoints, $a_{ij}^{\mathrm{rec}}$ measures mutually supported local continuity and is used below when evaluating whether a regional refinement separates strongly connected samples.

\subsection{Local Description-Length Model}
\label{subsec:granular_ball_coding}
A local region can be approximately isotropic or concentrated around a lower-dimensional direction. MDL-GBTRSC therefore evaluates each subset under two competing Gaussian descriptions that share the empirical center $c_S$. For $S\subseteq V$, let $n_S=|S|$ and let $\sigma_S^2$ be the within-ball dispersion in Eq.~\eqref{eq:prelim_dispersion}. The local description combines Gaussian residual codelengths with BIC-type asymptotic parameter codelengths. Accordingly, the MDL formulation used here is a BIC-style local description-length comparison over the deterministic retain/refine candidate family defined below. To define the logarithmic variance terms on degenerate local configurations, let $\mathcal D_+$ contain the positive squared nearest-neighbor distances in the dataset and let $\bar v=d^{-1}\sum_{j=1}^d\operatorname{Var}(X_{:j})$. Define
\begin{equation}
		\sigma_0^2=
		\max\left\{\varepsilon_{\rm mach},
		\begin{cases}
			\operatorname{median}(\mathcal D_+)/d, & \mathcal D_+\neq\varnothing,\\[2pt]
			\bar v/\max\{n,1\}, & \mathcal D_+=\varnothing,
		\end{cases}\right\},
		\qquad
		\ell(v)=\ln\max\{v,\sigma_0^2\}.
	\label{eq:variance_resolution}
\end{equation}
The same machine-level floors are used for zero distances, zero empirical variances, and logarithmic edge costs throughout the algorithm. A zero-diameter subset omits the diameter ordering introduced below, while constant directions remain finite through $\ell(v)$. These rules do not introduce dataset-specific tuning parameters.

\subsubsection{Isotropic description}
The isotropic description models the residuals around $c_S$ with a common scalar dispersion:
\begin{equation}
	L_{\mathrm{iso}}(S)
	=
	\frac{n_Sd}{2}\left[\ln(2\pi)+1+\ell(\sigma_S^2)\right]
	+\frac{d+1}{2}\ln n_S.
	\label{eq:iso_code}
\end{equation}
The first term describes the residual data and the second accounts for the center and scalar dispersion parameters.

\subsubsection{Subspace-adaptive description}
An anisotropic region is described through a principal subspace and its orthogonal complement. Let
\begin{equation}
	M_S=\sum_{i\in S}(x_i-c_S)(x_i-c_S)^\top
	\label{eq:local_scatter}
\end{equation}
have eigenvalues $\lambda_1\ge\cdots\ge\lambda_d\ge0$. For $q\in\{0,\ldots,d\}$, define
\begin{equation}
	\sigma_{\parallel,q}^{2}
	=\frac{\sum_{\ell=1}^{q}\lambda_\ell}{n_Sq},\quad q>0,
	\label{eq:parallel_variance}
\end{equation}
\begin{equation}
	\sigma_{\perp,q}^{2}
	=\frac{\sum_{\ell=q+1}^{d}\lambda_\ell}{n_S(d-q)},\quad q<d.
	\label{eq:perp_variance}
\end{equation}
The residual codelengths of the two components are
\begin{equation}
	\Psi_{\parallel}(S,q)
	=
	\begin{cases}
		\dfrac{n_Sq}{2}\left[\ln(2\pi)+1+\ell(\sigma_{\parallel,q}^{2})\right], & q>0,\\
		0, & q=0,
	\end{cases}
	\label{eq:parallel_code}
\end{equation}
\begin{equation}
	\Psi_{\perp}(S,q)
	=
	\begin{cases}
		\dfrac{n_S(d-q)}{2}\left[\ln(2\pi)+1+\ell(\sigma_{\perp,q}^{2})\right], & q<d,\\
		0, & q=d.
	\end{cases}
	\label{eq:perp_code}
\end{equation}
and the corresponding subspace description is
\begin{equation}
	L_{\mathrm{sub}}(S,q)
	=\Psi_{\parallel}(S,q)+\Psi_{\perp}(S,q)
	+\frac{k_{\mathrm{sub}}(q)}{2}\ln n_S+\ln(d+1),
	\label{eq:subspace_code_q}
\end{equation}
\begin{equation}
	k_{\mathrm{sub}}(q)=d+\mathbb I(q>0)+\mathbb I(q<d)+q(d-q).
	\label{eq:subspace_param_count}
\end{equation}
Here $d$ parameters specify the center, the indicator terms count the active variance components, and $q(d-q)$ is the dimension of the Grassmann representation of the local subspace. The term $\ln(d+1)$ accounts for the choice of $q$. Thus
\begin{equation}
	L_{\mathrm{sub}}(S)=\min_{0\le q\le d}L_{\mathrm{sub}}(S,q).
	\label{eq:subspace_code}
\end{equation}
For $d>1$, the local granular-ball description is
\begin{equation}
	L_{\mathrm{ball}}(S)=\min\{L_{\mathrm{iso}}(S),L_{\mathrm{sub}}(S)\}+\ln2,
	\label{eq:leaf_ball_code}
\end{equation}
where $\ln2$ accounts for the choice between the isotropic and subspace-adaptive descriptions. For $d=1$, $L_{\mathrm{ball}}(S)=L_{\mathrm{iso}}(S)$.

\subsection{Graph-Regularized Granular-Ball Refinement}
\label{subsec:graph_regularized_tree_induction}
The hierarchy is initialized with the complete index set $V$. At each active node, two local representations are compared: retaining $S$ as one region or replacing it by two child regions. The split description includes the two child models together with the cost of the separated reciprocal relations and the identity of the selected refinement. For a current subset $S$, retaining the region has codelength
\begin{equation}
	L_{\mathrm{keep}}(S)=L_{\mathrm{ball}}(S)+L_{\mathrm{choice}},
	\label{eq:keep_code}
\end{equation}
where $L_{\mathrm{choice}}=\ln2$. For an admissible binary refinement $S\to(S_L,S_R)$,
\begin{equation}
	L_{\mathrm{split}}(S_L,S_R)
	=L_{\mathrm{ball}}(S_L)+L_{\mathrm{ball}}(S_R)
	+L_{\mathrm{cut}}(S_L,S_R)+L_{\mathrm{choice}}+L_{\mathrm{cand}}(S).
	\label{eq:split_code}
\end{equation}
The graph term measures the reciprocal continuity separated by the refinement:
\begin{equation}
	L_{\mathrm{cut}}(S_L,S_R)
	=\sum_{\substack{i\in S_L,\,j\in S_R\\\{i,j\}\in E_{\mathrm{rec}}^0}}
	-\ln(1-a_{ij}^{\mathrm{rec}}).
	\label{eq:cut_code}
\end{equation}

Both child subsets must contain at least $n_{\min}(S)$ samples. Following the adaptive local-support rule used in MDL-based granular-ball generation~\citep{Xian2026MDLGBG},
\begin{equation}
	n_{\min}(S)=
	\begin{cases}
		1, & n_S\le3,\\[3pt]
		\min\left\{\left\lfloor\dfrac{n_S}{2}\right\rfloor,
		\max\left[2,\left\lceil\min\left(\dfrac{\sqrt{n_S}}{\ln\sqrt{d+2}},d+2\right)\right\rceil\right]\right\}, & n_S>3.
	\end{cases}
	\label{eq:n_min}
\end{equation}

Candidate binary refinements are generated from four one-dimensional geometric score functions. Let $u_1$ be a leading eigenvector of the local scatter matrix $M_S$. Let $a$ be a sample farthest from $c_S$ and $b$ a sample farthest from $x_a$, with ties resolved by sample index. When $\|x_b-x_a\|_2>0$, define $v_{\rm diam}=(x_b-x_a)/\|x_b-x_a\|_2$. Let $j^\star$ denote a feature of maximum empirical variance in $S$, with ties resolved by the smallest feature index. The four scores are
\begin{equation}
		\begin{aligned}
			o_{\rm pc}(i)&=u_1^\top x_i, &
			o_{\rm diam}(i)&=v_{\rm diam}^\top x_i,\\
			o_{\rm var}(i)&=x_{ij^\star}, &
			o_{\rm rad}(i)&=\|x_i-c_S\|_2^2.
	\end{aligned}
	\label{eq:candidate_orderings}
\end{equation}
Each nonconstant score induces an ordering of the samples in $S$; equal scores are ordered by sample index. The diameter ordering is omitted when the local empirical diameter is zero, and orderings that induce the same sample sequence are retained only once. Denote the resulting set of distinct orderings by $\mathcal O_S$, so that $|\mathcal O_S|\le4$. For any $o\in\mathcal O_S$, cutting the ordered sample sequence after position $h$ produces a binary refinement $(S_L^{o,h},S_R^{o,h})$. The admissible cut positions are
\begin{equation}
	\mathcal H_S=\{h:n_{\min}(S)\le h\le n_S-n_{\min}(S)\},
	\label{eq:admissible_cut_positions}
\end{equation}
with selection codelength
\begin{equation}
	L_{\mathrm{cand}}(S)=\ln(|\mathcal O_S|+1)+\ln(|\mathcal H_S|+1).
	\label{eq:candidate_code}
\end{equation}
If several candidate refinements have the same complete codelength, ties are resolved lexicographically by the fixed ordering sequence in Eq.~\eqref{eq:candidate_orderings} and then by cut position. If several current leaves have the same positive gain, the leaf with the smallest sample index is refined first.

Evaluating the full subspace-adaptive description at every admissible cut can be unnecessarily expensive. The cut positions are therefore evaluated in two stages. An isotropic score
\begin{equation}
	\widetilde L_{\mathrm{split}}(S_L,S_R)
	=L_{\mathrm{iso}}(S_L)+L_{\mathrm{iso}}(S_R)+L_{\mathrm{choice}}+L_{\mathrm{cand}}(S)
	\label{eq:screening_split_code}
\end{equation}
ranks the admissible positions within each ordering. Let
\begin{equation}
	\rho_S=\max\{2,\lceil\sqrt{n_S}\rceil\}.
	\label{eq:screening_budget}
\end{equation}
For each ordering $o$, let $\mathcal C_{S,o}$ contain the $\min\{|\mathcal H_S|,\rho_S\}$ admissible cuts with the smallest preliminary costs, and define $\mathcal C_S=\bigcup_{o\in\mathcal O_S}\mathcal C_{S,o}$. The full criterion in Eq.~\eqref{eq:split_code} is evaluated on $\mathcal C_S$. The best refinement cost and its gain over retention are
\begin{equation}
	L_{\mathrm{split}}^\ast(S)
	=\min_{(S_L,S_R)\in\mathcal C_S}L_{\mathrm{split}}(S_L,S_R),
	\label{eq:best_split}
\end{equation}
\begin{equation}
	\Delta(S)=L_{\mathrm{keep}}(S)-L_{\mathrm{split}}^\ast(S),
	\label{eq:mdl_gain}
\end{equation}
with $L_{\mathrm{split}}^\ast(S)=+\infty$ when no admissible refinement exists.

\begin{definition}[Stable leaf ball]
	A current subset $S$ is stable if no admissible candidate in $\mathcal C_S$ yields a positive description-length gain, equivalently $\Delta(S)\le0$. The granular ball $B(S)$ associated with such a terminal subset is called a stable leaf ball.
\end{definition}
Among current leaves with positive gain, the subset with the largest $\Delta(S)$ is refined first. Each accepted refinement replaces one nonempty leaf by two nonempty children and therefore increases the number of leaves by one; since the leaves form a partition of $n$ samples, the procedure terminates after finitely many accepted refinements. The resulting terminal subsets $\mathcal S^\star=\{S_1^\star,\ldots,S_m^\star\}$ form a partition of $V$, with corresponding stable leaf balls $\mathcal B^\star=\{B(S_1^\star),\ldots,B(S_m^\star)\}$. These terminal regions provide the local structural scales used in the sample-level affinity graph.

\subsection{Leaf-Scale Affinity Regularization}
\label{subsec:leaf_scale_affinity_regularization}
Let $b(i)$ denote the stable leaf ball containing sample $i$. Each leaf scale combines the spatial extent of the covered region with its average residual dispersion. For $B(S_b^\star)$, define
\begin{equation}
	\eta_b=r_{S_b^\star}+\sqrt{\max\{\sigma_{S_b^\star}^2,\sigma_0^2\}}+\varepsilon_{\rm mach}.
	\label{eq:leaf_coding_scale}
\end{equation}
For $j\in N_g(i)$, the sample-level edge cost is
\begin{equation}
	\phi_{ij}
	=\frac{\|x_i-x_j\|_2^2}{2\tau_i\tau_j}
	+\left|\ln\frac{\eta_{b(i)}}{\eta_{b(j)}}\right|,
	\label{eq:tree_regularized_cost}
\end{equation}
The first term in Eq.~\eqref{eq:tree_regularized_cost} measures geometric proximity under the self-tuning neighborhood scales, whereas the second measures the discrepancy between the structural scales of the two terminal regions. Nearby samples embedded in regions of comparable scale therefore retain stronger affinity than equally distant samples lying across a pronounced scale transition. The directed affinity is
\begin{equation}
	\bar w_{ij}=\exp(-\phi_{ij}).
	\label{eq:directed_tree_affinity}
\end{equation}
Symmetrization gives a reciprocal graph
\begin{equation}
	w_{ij}^{\mathrm{rec}}
	=\begin{cases}
		\min\{\bar w_{ij},\bar w_{ji}\}, & \bar w_{ij}>0\ \text{and}\ \bar w_{ji}>0,\\
		0, & \text{otherwise},
	\end{cases}
	\label{eq:reciprocal_tree_affinity}
\end{equation}
and a completed graph
\begin{equation}
	w_{ij}^{\mathrm{com}}=\max\{\bar w_{ij},\bar w_{ji}\}.
	\label{eq:completed_tree_affinity}
\end{equation}
The reciprocal graph retains only mutually supported local relations, while the completed graph also preserves a relation supported from one endpoint. They therefore provide a conservative and a connectivity-preserving realization of the same leaf-regularized neighborhood affinities.

\subsection{Graph Selection and Spectral Partitioning}
\label{subsec:graph_selection_partitioning}
Let $c_{\rm rec}$ and $c_{\rm com}$ be the numbers of connected components in the reciprocal and completed graphs. When the reciprocal graph already contains exactly $K$ connected components, its topology directly matches the prescribed partition order and it is retained. Otherwise, the completed graph is used to preserve the broader set of leaf-regularized local relations before spectral subdivision. Let $W$ denote the selected affinity matrix and $c$ its number of connected components.

If $c=K$, the connected components directly define the partition. If $c=1$, the $K$ eigenvectors associated with the smallest eigenvalues of the normalized Laplacian in Eq.~\eqref{eq:prelim_laplacian} are discretized by deterministic QR-based spectral rounding. For $1<c<K$, the disconnected components remain separated and the $K$ clusters are distributed among them by
\begin{equation}
		(K_1^\star,\ldots,K_c^\star)
		\in\arg\max_{\substack{1\le K_t\le n_t\\\sum_{t=1}^cK_t=K}}
		\sum_{t=1}^c n_tH_{K_t},
		\qquad
		H_q=\sum_{r=1}^q\frac1r,
	\label{eq:component_budget_allocation}
\end{equation}
where $n_t$ is the size of component $t$. The concavity of $H_q$ gives a size-weighted allocation with diminishing marginal gain: after assigning one cluster to each component, additional clusters are assigned according to the marginal value $n_t/(K_t+1)$. Components are indexed by their smallest sample index, which also resolves equal marginal values. Components with $K_t^\star>1$ are partitioned spectrally. If $c>K$, a global $K$-way spectral partition is applied to the selected graph. The completed graph has positive degree at every vertex for $n>1$ because max symmetrization retains each directed neighborhood relation. If the reciprocal graph is selected with $c_{\rm rec}=K$, any singleton component is retained directly as one component cluster.

\subsection{Algorithm}
\label{subsec:algorithm}

Algorithm~\ref{alg:mdl_gbtrsc} summarizes the complete procedure of
	MDL-GBTRSC, including reciprocal-continuity construction, hierarchical
	granular-ball refinement, leaf-scale affinity regularization, and the final
	spectral partition.

\begin{algorithm}[t]
	\DontPrintSemicolon
	\caption{MDL-GBTRSC}
	\label{alg:mdl_gbtrsc}
	
	\KwIn{Data $X=\{x_i\}_{i=1}^{n}\subset\mathbb R^d$ and cluster number $K$}
	\KwOut{Clustering labels $\hat y$}
	
	Construct the directed neighborhood sets $N_g(i)$ and local scales $\tau_i$\;
	Build the initial reciprocal continuity graph $G_{\rm rec}^{0}$\;
	
	Initialize the leaf partition $\mathcal S\leftarrow\{V\}$\;
	\For{each current leaf $S\in\mathcal S$}{
		Compute the retain description length $L_{\rm keep}(S)$\;
		Generate admissible geometric orderings and cut positions for $S$\;
		Rank candidate cuts by the preliminary split criterion and retain the top $\rho_S$ cuts per ordering\;
		Evaluate the complete description length of the retained candidates\;
		Record the best split $(S_L^\star,S_R^\star)$ and gain
			$\Delta(S)=L_{\rm keep}(S)-L_{\rm split}^\star(S)$\;
	}
	
	\While{$\max_{S\in\mathcal S}\Delta(S)>0$}{
		Select $S^\star=\arg\max_{S\in\mathcal S}\Delta(S)$\;
		Replace $S^\star$ by $(S_L^\star,S_R^\star)$ in $\mathcal S$\;
		Evaluate the admissible refinements and gains of the two new leaves\;
	}
	Set $\mathcal B^\star=\{B(S):S\in\mathcal S\}$ as the stable leaf-ball set\;
	
	\For{each $B_b\in\mathcal B^\star$}{
		Compute the leaf structural scale $\eta_b$\;
	}
	Regularize neighborhood affinities using sample distances and the structural scales of their leaf balls\;
	Construct the reciprocal graph $W_{\rm rec}$ and completed graph $W_{\rm com}$\;
	
	\eIf{$W_{\rm rec}$ contains exactly $K$ connected components}{
		Set $W\leftarrow W_{\rm rec}$\;
	}{
		Set $W\leftarrow W_{\rm com}$\;
	}
	
	Let $\{\mathcal V_t\}_{t=1}^{c}$ be the connected components of $W$\;
	\eIf{$c=K$}{
		Use the connected components directly as the $K$ clusters\;
	}{
		\eIf{$1<c<K$}{
			Allocate $K_t$ clusters to the components and perform spectral partitioning within each component\;
		}{
			Perform deterministic $K$-way spectral partitioning on $W$\;
		}
	}
	\Return{$\hat y$}\;
\end{algorithm}

\section{Complexity Analysis}
\label{sec:complexity}

Let $n$ and $d$ denote the number of samples and features, respectively, and let $k_g$ be the neighborhood order defined in Eq.~\eqref{eq:graph_order}. Let $\mathcal T$ denote the set of granular-ball nodes evaluated during refinement, with $n_S=|S|$ for $S\in\mathcal T$. The sparse neighborhood representation contains $m_g=O(nk_g)$ retained relations.

Construction of the reciprocal continuity graph requires the $k_g$-nearest neighbors of each sample. Using brute-force exact search as a conservative worst-case reference gives $O(n^2d)$ time. Once the neighbors have been obtained, the local scales, directed affinities, and reciprocal relations can be computed in $O(nk_g)$ time. The implementation retains the data matrix together with the neighbor indices, distances, and sparse graph quantities, requiring $O(nd+nk_g)$ storage. From Eq.~\eqref{eq:graph_order}, $k_g=O(\log n+\sqrt d)$ before the natural upper bound $n-1$.

During granular-ball refinement, the principal computation is the local description in Eq.~\eqref{eq:leaf_ball_code}. For a node $S$, the isotropic description requires $O(n_Sd)$ operations. For the subspace-adaptive description, let $Z_S$ denote the centered local data matrix. The nonzero eigenvalues of $Z_S^\top Z_S$ and $Z_SZ_S^\top$ coincide, allowing the smaller of the feature-space and sample-space eigensystems to be used. The resulting cost of a local granular-ball description is
\begin{equation}
		T_{\mathrm{ball}}(S)
		=
		O\!\left(
		n_Sd+
		\min\left\{
		n_Sd^2+d^3,\;
		n_S^2d+n_S^3
		\right\}
		\right).
	\label{eq:complexity_ball}
\end{equation}

At most four distinct orderings in $\mathcal O_S$ are considered at each node. Their geometric scores can be formed in $O(n_Sd)$ time, and sorting them requires $O(n_S\log n_S)$. The cumulative statistics used by the preliminary criterion in Eq.~\eqref{eq:screening_split_code} are also obtained in $O(n_Sd)$ time. For a fixed ordering, the graph-cut costs of all admissible prefix refinements can be accumulated by scanning the reciprocal neighbors of the samples in $S$. Since each reciprocal degree is at most $k_g$, this step requires $O(n_Sk_g)$ time. The complete description-length criterion is evaluated for at most $\rho_S=\max\{2,\lceil\sqrt{n_S}\rceil\}$ candidates per ordering. The two child descriptions associated with each candidate have the same asymptotic upper bound as Eq.~\eqref{eq:complexity_ball}. Omitting the constant factor associated with the finite ordering family, the tree-induction cost is therefore
\begin{equation}
		T_{\mathrm{tree}}
		=
		O\!\left(
		\sum_{S\in\mathcal T}
		\left[
		n_S\!\left(d+\log n_S+k_g\right)
		+
		\rho_S T_{\mathrm{ball}}(S)
		\right]
		\right).
	\label{eq:complexity_tree}
\end{equation}

The effect of the hierarchy can be made explicit under fixed $d$. In this setting, $T_{\mathrm{ball}}(S)=O(n_S)$ and $k_g=O(\log n)$. For an approximately balanced hierarchy of logarithmic depth, the non-model-evaluation terms in Eq.~\eqref{eq:complexity_tree} sum to $O(n\log^2 n)$, whereas
\begin{equation}
		\sum_{S\in\mathcal T}
		\rho_S T_{\mathrm{ball}}(S)
		=
		O\!\left(
		\sum_{S\in\mathcal T}n_S^{3/2}
		\right)
		=
		O(n^{3/2}).
	\label{eq:complexity_balanced}
\end{equation}
Thus, the tree-induction cost is $O(n^{3/2})$ in the balanced case. A maximally unbalanced admissible hierarchy can be bounded conservatively by node sizes $n,n-1,\ldots$, for which the non-model-evaluation terms are $O(n^2\log n)$ and
\begin{equation}
		\sum_{S\in\mathcal T}n_S^{3/2}
		=
		O(n^{5/2}),
	\label{eq:complexity_worst}
\end{equation}
giving the corresponding upper bound $O(n^{5/2})$. If the complete description were evaluated at every admissible cut rather than at the screened candidates, the local model-evaluation terms would instead be $O(n^2)$ and $O(n^3)$ for balanced and maximally unbalanced hierarchies, respectively. The empirical comparison with exhaustive evaluation is reported in Section~\ref{subsec:split_search_analysis}.

After the hierarchy has stabilized, the leaf scales in Eq.~\eqref{eq:leaf_coding_scale} are transferred to the stored neighborhood relations. The geometric term in Eq.~\eqref{eq:tree_regularized_cost} reuses the previously computed neighbor distances, so constructing the leaf-regularized reciprocal and completed graphs requires $O(nk_g)$ time. Connected-component analysis then requires $O(n+m_g)$. If the selected graph already has $K$ connected components, these components directly define the partition. Otherwise, spectral decomposition is performed globally or within the components that require further subdivision, as described in Section~\ref{subsec:graph_selection_partitioning}. For a component with $N$ samples, $m$ retained relations, and a $q$-dimensional spectral embedding, the eigensolver cost depends on the sparse numerical solver and its convergence behavior, while the deterministic QR-based discretization requires $O(Nq^2+q^3)$ time.

Consequently, for fixed $d$ and $K$, the non-spectral part of MDL-GBTRSC has the reference bounds
\begin{equation}
		T_{\mathrm{nonsp}}
		=
		\begin{cases}
			O(n^2), & \text{approximately balanced hierarchy},\\[3pt]
			O(n^{5/2}), & \text{maximally unbalanced hierarchy},
		\end{cases}
	\label{eq:complexity_summary}
\end{equation}
where the quadratic term in the balanced case follows from conservative exact neighborhood construction. The spectral cost is incurred only for the graph components requiring further subdivision.

The memory requirement also depends on the learned hierarchy. Besides the $O(nd+nk_g)$ data and neighborhood storage, each evaluated granular-ball node retains its sample membership and $O(d)$ local statistics. The hierarchy therefore occupies
\begin{equation}
		M_{\mathrm{tree}}
		=
		O\!\left(
		\sum_{S\in\mathcal T}n_S
		+
		|\mathcal T|d
		\right).
	\label{eq:complexity_tree_memory}
\end{equation}
The largest temporary workspace required by a local subspace evaluation is
\begin{equation}
		M_{\mathrm{local}}
		=
		O\!\left(
		\max_{S\in\mathcal T}
		\left[
		n_Sd+\min\{d^2,n_S^2\}
		\right]
		\right).
	\label{eq:complexity_local_memory}
\end{equation}
Together with an $O(nK)$ spectral embedding, the retained storage of the sparse formulation is
\begin{equation}
		M_{\mathrm{sparse}}
		=
		O\!\left(
		nd
		+
		nk_g
		+
		\sum_{S\in\mathcal T}n_S
		+
		|\mathcal T|d
		+
		M_{\mathrm{local}}
		+
		nK
		\right).
	\label{eq:complexity_memory}
\end{equation}
For fixed $d$ and $K$, this gives $O(n\log n)$ retained storage for an approximately balanced hierarchy and an $O(n^2)$ upper bound for a maximally unbalanced hierarchy. Dense eigendecomposition, when used for modest graph sizes or as a numerical fallback, additionally requires $O(n^2)$ matrix storage. The runtime and peak-memory experiments in Section~\ref{subsec:scalability} complement these bounds by examining the effects of both $n$ and $d$ under the same implementation setting.

\section{Experiments}
\label{sec:experiments}

The empirical evaluation examines both the clustering performance of MDL-GBTRSC and the structural role of its granular-ball hierarchy. Eighteen real and twenty synthetic datasets are evaluated under a common fixed-$K$ protocol. The experiments compare clustering accuracy, statistical significance, runtime, and synthetic-data partitions, and then examine the learned graph structure, split screening, component ablations, sensitivity to the prescribed cluster number, and computational scalability. Unless stated otherwise, every method receives the same reference cluster number $K$, and the reference labels are used only to obtain this common model order and to compute ARI and NMI.

\subsection{Experimental Protocol}
\label{subsec:experimental_settings}

All experiments were conducted in Python 3.12 on a Windows 11 workstation equipped with an AMD Ryzen 9 8940HX CPU, an NVIDIA GeForce RTX 5070 Ti GPU with 12 GB memory, and 32 GB system memory. Each feature was Min--Max scaled to $[0,1]$, with constant features mapped to zero, and the same externally normalized data were supplied to every compared method. The source code, benchmark scripts, and experimental outputs are available at \href{https://github.com/AnonymousUser0816/MDL-GBTRSC}{https://github.com/AnonymousUser0816/MDL-GBTRSC}.

The comparison covers the main graph-construction alternatives relevant to the proposed formulation. SC-kNN~\citep{von2007tutorial} and SC-RBF~\citep{ng2001spectral} instantiate conventional neighborhood and Gaussian sample graphs, STSC-FK uses a self-tuning local affinity~\citep{STSC-FK}, N-DSC represents direct adaptive graph learning~\citep{Kong2025DirectSC}, and FastSC provides a recent fast spectral construction~\citep{Cheng2025FSCPLGB}. GBCT~\citep{Xia2025GBCT} and GBSC~\citep{Xie2023GBSC} provide granular-ball comparisons, whereas MDMSC~\citep{Xu2025MDMSC} and GMM-SC~\citep{Zhang2025GMMSC} construct the spectral problem through micro-clusters or representative structures. All methods are rerun on the same 38 datasets after the common preprocessing step. The method-specific settings in Table~\ref{tab:parameter_settings} are fixed across datasets and follow the released implementations or the configurations adopted by the corresponding methods. No per-dataset setting is selected from reference labels; consequently, the reported comparison concerns fixed reproducible configurations rather than label-oracle parameter optima.

\begin{table}[htbp]
	\centering
	\caption{Parameter settings of the compared methods.}
	\label{tab:parameter_settings}
	\scriptsize
	\setlength{\tabcolsep}{4pt}
	\renewcommand{\arraystretch}{1.1}
	\begin{threeparttable}
		\begin{tabular}{l>{\arraybackslash}p{0.72\columnwidth}}
			\toprule
			Method & Parameter setting \\
			\midrule
			MDL-GBTRSC & No dataset-specific tunable parameter \\
			STSC-FK & Self-tuning local scale; no additional parameter \\
			N-DSC & $\lambda=1$ \\
			FastSC & Symmetric unweighted $10$-NN graph, $t=15$ \\
			SC-kNN & $k=10$ \\
			SC-RBF & $\gamma=1$ \\
			GBCT & Minimum ball size $=2$, sparsity coefficient $=0.2$ \\
			MDMSC & $k=4$, $\lambda_{\rm ratio}=1.5$, $\beta=16$ \\
			GBSC & Control parameter $=0.1$ \\
			GMM-SC & $\tau=0.8$, $M=250$, $s=15$ \\
			\bottomrule
		\end{tabular}
	\end{threeparttable}
\end{table}

Under the adopted implementations, MDL-GBTRSC, STSC-FK, and N-DSC are deterministic. MDL-GBTRSC consequently requires no random seed, and its spectral branches use the deterministic QR-based discretization described in Section~\ref{subsec:graph_selection_partitioning}. For the remaining methods, 20 runs are performed with seeds $0,\ldots,19$. SC-kNN and SC-RBF use discretization-based spectral label assignment with the corresponding trial seed; in MDMSC only the random state of the internal K-means step is varied; FastSC, GBSC, and GMM-SC receive the same trial seed in their stochastic steps. The detailed clustering tables report the sample mean and sample standard deviation over these 20 runs, whereas deterministic methods are represented by their unique outputs. For runtime evaluation, each method is run 20 times on every dataset and the mean elapsed time is reported. When a single stochastic partition is required for qualitative visualization, the seed-0 result is used; deterministic methods are shown by their unique partitions.

Numerical edge cases were additionally checked using duplicate samples, zero-variance features, repeated rows, an all-identical one-cluster dataset, and near-machine-spacing coordinates. Under the numerical rules in Section~\ref{subsec:granular_ball_coding}, all tests returned finite labels and finite stored graph quantities.

Clustering quality is evaluated by the Adjusted Rand Index (ARI) and normalized mutual information (NMI), both of which are invariant to permutations of the numerical cluster labels. Tables~\ref{tab:dataset_info} and~\ref{tab:synthetic_dataset_details} give the sample size, feature dimension, and reference class count of the real\footnote{\url{https://archive.ics.uci.edu/}} and synthetic benchmarks\footnote{\url{https://github.com/milaan9/Clustering-Datasets/tree/master}}\footnote{\url{https://github.com/wylbdthxbw/GBC}}; for the synthetic data, a concise description of the geometric structure is also included to make the controlled test cases explicit. A run that cannot be completed under the common memory constraint is recorded as memory exceeded (ME). Statistical tests that require a common result matrix are computed only from datasets completed by every method in that comparison.

\begin{table}[htbp]
	\centering
	\caption{Real benchmark datasets used in the experiments.}
	\label{tab:dataset_info}
	\scriptsize
	\setlength{\tabcolsep}{2.5pt}
	\renewcommand{\arraystretch}{1.05}
	\begin{tabular}{lp{0.25\columnwidth}rrr}
		\toprule
		Abbr. & DataSet & Samples & Features & Classes \\
		\midrule
		BTissue & Breast Tissue & 106 & 9 & 6 \\
		Iris & Iris & 150 & 4 & 3 \\
		Wine & Wine & 178 & 13 & 3 \\
		Seeds & Seeds & 210 & 8 & 3 \\
		Glass & Glass Identification & 214 & 9 & 6 \\
		Thyroid & New Thyroid & 215 & 5 & 3 \\
		Libras & Libras & 360 & 90 & 15 \\
		WDBC & Breast Cancer Wisconsin & 569 & 30 & 2 \\
		Banknote & Banknote Authentication & 1372 & 4 & 2 \\
		Segment & Image Segmentation & 2310 & 19 & 7 \\
		Rice & Rice Cammeo and Osmancik & 3810 & 7 & 2 \\
		PageBlocks & Page Blocks Classification & 5473 & 10 & 5 \\
		Digits & Optical Handwritten Digits & 5620 & 64 & 10 \\
		Landsat & Landsat Satellite & 6435 & 36 & 6 \\
		Isolet & Isolet & 7797 & 617 & 26 \\
		PenDigits & Pen-Based Digits & 10992 & 16 & 10 \\
		Bean & Dry Bean & 13611 & 16 & 7 \\
		HTRU2 & HTRU2 & 17898 & 8 & 2 \\
		\bottomrule
	\end{tabular}
\end{table}

\begin{table}[htbp]
	\centering
	\caption{Synthetic datasets used in the experiments.}
	\label{tab:synthetic_dataset_details}
	\scriptsize
	\setlength{\tabcolsep}{4pt}
	\renewcommand{\arraystretch}{1.1}
	\begin{threeparttable}
		\begin{tabular}{lrrr>{\arraybackslash}p{0.55\columnwidth}}
			\toprule
			DataSet & Samples & Features & Classes & Main challenge \\
			\midrule
			zelnik6 & 238 & 2 & 3 & nested clusters, non-convex structure \\
			zelnik1 & 299 & 2 & 3 & nested clusters, non-convex structure \\
			blobs & 300 & 2 & 3 & overlapping Gaussian-like clusters \\
			zelnik2 & 303 & 2 & 3 & uneven cluster sizes, background cluster \\
			3-spiral & 312 & 2 & 3 & intertwined spiral structure, non-convex clusters \\
			db2 & 315 & 2 & 4 & nonlinear separability, shape diversity \\
			jain & 373 & 2 & 2 & density sparsity, non-convex structure \\
			zelnik4 & 622 & 2 & 4 & uneven cluster sizes, background cluster \\
			target & 770 & 2 & 6 & nested clusters, uneven cluster sizes, non-convex structure \\
			aggregation & 788 & 2 & 7 & touching boundaries, bridging structures, shape diversity \\
			chainlink & 1000 & 3 & 2 & nonlinear separability, non-convex structure \\
			fourty & 1000 & 2 & 40 & many Gaussian-like clusters \\
			wingnut & 1016 & 2 & 2 & sparse density distribution \\
			N & 1016 & 2 & 4 & nested clusters, non-convex structure \\
			A & 1735 & 2 & 6 & nonlinear separability, shape diversity \\
			complex8 & 2551 & 2 & 8 & nonlinear separability, shape diversity \\
			diamond9 & 3000 & 2 & 9 & touching boundaries, overlapping clusters \\
			cure-t2-4k & 4200 & 2 & 6 & noise contamination, bridging structures, uneven cluster sizes \\
			banana & 4811 & 2 & 2 & banana-shaped clusters, nonlinear separability \\
			cluto-t8-8k & 8000 & 2 & 8 & noise contamination, nonlinear separability, shape diversity \\
			\bottomrule
		\end{tabular}
	\end{threeparttable}
\end{table}

\subsection{Overall Clustering Performance}
\label{subsec:comparative_performance}

\subsubsection{Real benchmark results}
\label{subsec:real_results}

Table~\ref{tab:ari_nmi_comparison} reports the ARI and NMI values on the 18 real datasets. MDL-GBTRSC obtains mean ARI/NMI values of $0.5933/0.6431$ over all 18 datasets. Since STSC-FK and N-DSC exceed the available memory on HTRU2, the common ten-method comparison is based on the remaining 17 datasets. On these complete cases, MDL-GBTRSC reaches mean ARI/NMI values of $0.5998/0.6613$ and average ranks of $2.94/2.94$, both the best overall values in the comparison. N-DSC has the second-highest complete-case means, $0.5792$ for ARI and $0.6285$ for NMI, whereas SC-kNN and STSC-FK achieve better average ranks than N-DSC. The difference between mean values and ranks indicates substantial variation in the relative ordering of the strongest methods across datasets.

At the dataset level, MDL-GBTRSC gives the highest ARI on Iris and Wine and the highest NMI on Wine, Glass, Segment, and Landsat. On Wine, it reaches $0.9309/0.9088$, exceeding the corresponding scores of STSC-FK, N-DSC, SC-kNN, MDMSC, and the remaining methods. Segment and Landsat exhibit a metric-dependent ordering: MDL-GBTRSC obtains the highest NMI values, $0.6852$ and $0.6723$, whereas N-DSC gives the highest ARI on both datasets. Banknote shows a larger reversal, with N-DSC reaching $0.8648/0.7865$ compared with $0.6011/0.5567$ for MDL-GBTRSC. STSC-FK is strongest on PageBlocks, N-DSC leads on Digits and PenDigits, and GBSC obtains the largest ARI on Bean. These dataset-level differences account for the relatively close average ranks among the leading methods on the real suite.

Banknote also shows pronounced sensitivity to graph construction. For MDL-GBTRSC, the reciprocal graph contains 10 connected components although $K=2$, while the completed graph is connected and is partitioned globally. In the ablation study, removing graph continuity and leaf-scale regularization changes its ARI by $-0.142$ and $-0.183$, respectively. Together with the wide spread of scores among the compared methods, this indicates that the representation of local connectivity and regional scale has a substantial effect on the Banknote partition.

Across the 17 complete real datasets, MDL-GBTRSC has higher ARI than STSC-FK, N-DSC, SC-kNN, and MDMSC on 10, 10, 9, and 13 datasets, respectively; for NMI the corresponding counts are 10, 13, 9, and 11. The same competitive pattern appears in the Holm-corrected tests, where the differences from STSC-FK, N-DSC, SC-kNN, and MDMSC are not significant on the real suite.

\begin{table}[htbp]
	\centering
	\scriptsize
	\setlength{\tabcolsep}{2.0pt}
	\renewcommand{\arraystretch}{1.06}
	\caption{Dataset-level comparison of ARI and NMI on the real benchmark datasets.}
	\label{tab:ari_nmi_comparison}
	\begin{threeparttable}
		\resizebox{\columnwidth}{!}{%
			\begin{tabular}{llcccccccccc}
				\toprule
				\multirow{2}{*}{Dataset} & \multirow{2}{*}{Metric}
				& \multirow{2}{*}{Ours}
				& \multicolumn{5}{c}{Spectral baselines}
				& \multirow{2}{*}{\makecell{GBCT\\(TNNLS 2025)}}
				& \multirow{2}{*}{\makecell{MDMSC\\(AAAI 2025)}}
				& \multirow{2}{*}{\makecell{GBSC\\(TKDE 2023)}}
				& \multirow{2}{*}{\makecell{GMM-SC\\(TNNLS 2025)}} \\
				\cmidrule(lr){4-8}
				& & & STSC-FK & N-DSC & FastSC & SC-kNN & SC-RBF & & & & \\
				\midrule
				\multirow{2}{*}{BTissue} & ARI & 0.3718 & 0.3731 & \textbf{0.4202} & 0.3725$\pm$0.0307 & 0.4113$\pm$0.0076 & 0.2573$\pm$0.0350 & 0.1971$\pm$0.0621 & 0.3519$\pm$0.0151 & 0.3073$\pm$0.1064 & 0.2971$\pm$0.0142 \\
				& NMI & 0.5207 & 0.5376 & \textbf{0.5398} & 0.4913$\pm$0.0114 & 0.5386$\pm$0.0031 & 0.4827$\pm$0.0323 & 0.4615$\pm$0.0537 & 0.5238$\pm$0.0062 & 0.4285$\pm$0.1062 & 0.5285$\pm$0.0167 \\
				\midrule
				\multirow{2}{*}{Iris} & ARI & \textbf{0.9038} & 0.8176 & 0.6412 & 0.6482$\pm$0.1333 & 0.7445$\pm$0.0000 & 0.6207$\pm$0.0000 & 0.2507$\pm$0.2852 & 0.7920$\pm$0.0802 & 0.6537$\pm$0.0000 & 0.8883$\pm$0.1132 \\
				& NMI & 0.8851 & 0.8038 & 0.6736 & 0.7275$\pm$0.0688 & 0.7777$\pm$0.0000 & 0.6448$\pm$0.0000 & 0.3265$\pm$0.3707 & 0.8238$\pm$0.0441 & 0.7490$\pm$0.0000 & \textbf{0.8958}$\pm$0.0613 \\
				\midrule
				\multirow{2}{*}{Wine} & ARI & \textbf{0.9309} & 0.8806 & 0.7125 & 0.7418$\pm$0.1647 & 0.8685$\pm$0.0000 & 0.9149$\pm$0.0000 & 0.2130$\pm$0.2231 & 0.7414$\pm$0.0000 & 0.5522$\pm$0.0000 & 0.8665$\pm$0.0430 \\
				& NMI & \textbf{0.9088} & 0.8494 & 0.6968 & 0.7677$\pm$0.0972 & 0.8529$\pm$0.0000 & 0.8926$\pm$0.0000 & 0.3075$\pm$0.2425 & 0.7528$\pm$0.0000 & 0.6430$\pm$0.0000 & 0.8506$\pm$0.0347 \\
				\midrule
				\multirow{2}{*}{Seeds} & ARI & 0.6896 & 0.7331 & 0.6256 & 0.5681$\pm$0.1736 & 0.6871$\pm$0.0103 & 0.7218$\pm$0.0064 & 0.4385$\pm$0.2171 & \textbf{0.7556}$\pm$0.0145 & 0.7338$\pm$0.0000 & 0.5766$\pm$0.0322 \\
				& NMI & 0.6696 & 0.7049 & 0.6350 & 0.5983$\pm$0.1283 & 0.6793$\pm$0.0001 & 0.6929$\pm$0.0052 & 0.5089$\pm$0.1809 & \textbf{0.7249}$\pm$0.0116 & 0.7069$\pm$0.0000 & 0.6162$\pm$0.0153 \\
				\midrule
				\multirow{2}{*}{Glass} & ARI & 0.2207 & 0.1732 & 0.2021 & 0.1428$\pm$0.0157 & 0.2189$\pm$0.0000 & 0.1742$\pm$0.0032 & 0.1358$\pm$0.0686 & \textbf{0.2234}$\pm$0.0449 & 0.1426$\pm$0.0003 & 0.1841$\pm$0.0178 \\
				& NMI & \textbf{0.3939} & 0.3804 & 0.3651 & 0.2823$\pm$0.0119 & 0.3601$\pm$0.0000 & 0.3404$\pm$0.0069 & 0.2721$\pm$0.0727 & 0.3701$\pm$0.0290 & 0.2824$\pm$0.0012 & 0.3461$\pm$0.0279 \\
				\midrule
				\multirow{2}{*}{Thyroid} & ARI & 0.8273 & \textbf{0.8597} & 0.7452 & 0.2525$\pm$0.1563 & 0.8007$\pm$0.0000 & 0.6592$\pm$0.0000 & 0.2621$\pm$0.0770 & 0.7997$\pm$0.0000 & 0.7022$\pm$0.0000 & 0.7801$\pm$0.0955 \\
				& NMI & 0.7574 & \textbf{0.7796} & 0.6481 & 0.3439$\pm$0.1147 & 0.7021$\pm$0.0000 & 0.6245$\pm$0.0000 & 0.2924$\pm$0.0821 & 0.7219$\pm$0.0000 & 0.7082$\pm$0.0000 & 0.6988$\pm$0.0709 \\
				\midrule
				\multirow{2}{*}{Libras} & ARI & 0.3538 & 0.3534 & 0.3027 & 0.2809$\pm$0.0266 & \textbf{0.3627}$\pm$0.0022 & 0.3121$\pm$0.0165 & 0.1428$\pm$0.0584 & 0.3440$\pm$0.0219 & 0.1048$\pm$0.0112 & 0.3146$\pm$0.0240 \\
				& NMI & 0.6171 & 0.6186 & 0.6018 & 0.5547$\pm$0.0228 & 0.6325$\pm$0.0010 & 0.6024$\pm$0.0113 & 0.3628$\pm$0.1063 & \textbf{0.6360}$\pm$0.0157 & 0.3353$\pm$0.0172 & 0.5937$\pm$0.0182 \\
				\midrule
				\multirow{2}{*}{WDBC} & ARI & 0.7735 & 0.7430 & 0.7373 & 0.7821$\pm$0.0225 & \textbf{0.8115}$\pm$0.0000 & 0.6933$\pm$0.0000 & 0.0588$\pm$0.1793 & 0.7423$\pm$0.0000 & 0.7243$\pm$0.0000 & 0.6878$\pm$0.0207 \\
				& NMI & 0.6670 & 0.6251 & 0.6221 & 0.6975$\pm$0.0348 & \textbf{0.7101}$\pm$0.0000 & 0.6143$\pm$0.0000 & 0.0867$\pm$0.1584 & 0.6410$\pm$0.0000 & 0.6106$\pm$0.0000 & 0.5748$\pm$0.0203 \\
				\midrule
				\multirow{2}{*}{Banknote} & ARI & 0.6011 & 0.1419 & \textbf{0.8648} & 0.0748$\pm$0.1079 & 0.0009$\pm$0.0029 & 0.0523$\pm$0.0000 & -0.0033$\pm$0.0074 & -0.0053$\pm$0.0002 & 0.0268$\pm$0.0000 & -0.0015$\pm$0.0000 \\
				& NMI & 0.5567 & 0.2433 & \textbf{0.7865} & 0.0813$\pm$0.0841 & 0.0108$\pm$0.0073 & 0.0379$\pm$0.0000 & 0.0068$\pm$0.0024 & 0.0304$\pm$0.0193 & 0.0797$\pm$0.0000 & 0.0017$\pm$0.0001 \\
				\midrule
				\multirow{2}{*}{Segment} & ARI & 0.5174 & 0.4736 & \textbf{0.5756} & 0.2789$\pm$0.0711 & 0.4797$\pm$0.0048 & 0.4150$\pm$0.0268 & 0.3163$\pm$0.0776 & 0.1218$\pm$0.0363 & 0.4783$\pm$0.0133 & 0.4771$\pm$0.0477 \\
				& NMI & \textbf{0.6852} & 0.6644 & 0.6564 & 0.4634$\pm$0.0630 & 0.6789$\pm$0.0106 & 0.5727$\pm$0.0179 & 0.5655$\pm$0.0390 & 0.3909$\pm$0.0684 & 0.6340$\pm$0.0186 & 0.6153$\pm$0.0255 \\
				\midrule
				\multirow{2}{*}{Rice} & ARI & 0.6920 & 0.6938 & 0.5986 & 0.3520$\pm$0.2172 & \textbf{0.6981}$\pm$0.0000 & 0.6781$\pm$0.0000 & 0.0011$\pm$0.0053 & 0.6767$\pm$0.0017 & 0.6824$\pm$0.0000 & 0.5129$\pm$0.0476 \\
				& NMI & 0.5791 & 0.5809 & 0.5182 & 0.3070$\pm$0.1803 & \textbf{0.5854}$\pm$0.0000 & 0.5643$\pm$0.0000 & 0.0142$\pm$0.0083 & 0.5719$\pm$0.0012 & 0.5694$\pm$0.0000 & 0.4355$\pm$0.0361 \\
				\midrule
				\multirow{2}{*}{PageBlocks} & ARI & 0.2908 & \textbf{0.4283} & 0.0901 & 0.0704$\pm$0.0598 & 0.0771$\pm$0.0000 & 0.0883$\pm$0.0066 & 0.3004$\pm$0.1711 & 0.1121$\pm$0.0024 & 0.2673$\pm$0.0000 & 0.2432$\pm$0.0413 \\
				& NMI & 0.2342 & \textbf{0.3188} & 0.1557 & 0.1083$\pm$0.0440 & 0.1454$\pm$0.0000 & 0.1159$\pm$0.0081 & 0.1927$\pm$0.1028 & 0.0844$\pm$0.0021 & 0.1690$\pm$0.0000 & 0.2615$\pm$0.0362 \\
				\midrule
				\multirow{2}{*}{Digits} & ARI & 0.8043 & 0.5270 & \textbf{0.8609} & 0.7253$\pm$0.0546 & 0.8279$\pm$0.0202 & 0.6249$\pm$0.0003 & 0.0398$\pm$0.1241 & 0.8333$\pm$0.0000 & 0.2846$\pm$0.0112 & 0.5619$\pm$0.0415 \\
				& NMI & 0.8624 & 0.6791 & 0.8936 & 0.8322$\pm$0.0251 & 0.8927$\pm$0.0101 & 0.7739$\pm$0.0001 & 0.1123$\pm$0.1885 & \textbf{0.9029}$\pm$0.0000 & 0.4593$\pm$0.0051 & 0.6888$\pm$0.0251 \\
				\midrule
				\multirow{2}{*}{Landsat} & ARI & 0.5893 & 0.5807 & \textbf{0.6267} & 0.4138$\pm$0.0891 & 0.5530$\pm$0.0005 & 0.5260$\pm$0.0002 & 0.0884$\pm$0.0986 & 0.5673$\pm$0.0206 & 0.4813$\pm$0.0027 & 0.4468$\pm$0.0028 \\
				& NMI & \textbf{0.6723} & 0.6219 & 0.6388 & 0.5333$\pm$0.0611 & 0.6548$\pm$0.0001 & 0.5990$\pm$0.0004 & 0.2132$\pm$0.1600 & 0.6572$\pm$0.0018 & 0.5737$\pm$0.0027 & 0.5486$\pm$0.0056 \\
				\midrule
				\multirow{2}{*}{Isolet} & ARI & 0.4834 & 0.4166 & \textbf{0.5411} & 0.3999$\pm$0.0241 & 0.4875$\pm$0.0001 & 0.2025$\pm$0.0138 & 0.0168$\pm$0.0198 & 0.4412$\pm$0.0264 & 0.1494$\pm$0.0176 & 0.4751$\pm$0.0267 \\
				& NMI & 0.7498 & 0.6431 & 0.7463 & 0.7210$\pm$0.0160 & \textbf{0.7712}$\pm$0.0000 & 0.6241$\pm$0.0182 & 0.2228$\pm$0.1462 & 0.7534$\pm$0.0090 & 0.4179$\pm$0.0150 & 0.7107$\pm$0.0111 \\
				\midrule
				\multirow{2}{*}{PenDigits} & ARI & 0.5805 & 0.6335 & \textbf{0.7715} & 0.5055$\pm$0.0859 & 0.5901$\pm$0.0078 & 0.4926$\pm$0.0083 & 0.0970$\pm$0.0612 & 0.4144$\pm$0.0234 & 0.4248$\pm$0.0001 & 0.5499$\pm$0.0465 \\
				& NMI & 0.7923 & 0.7828 & \textbf{0.8474} & 0.6765$\pm$0.0509 & 0.7785$\pm$0.0048 & 0.6721$\pm$0.0047 & 0.3714$\pm$0.1276 & 0.7004$\pm$0.0084 & 0.5969$\pm$0.0003 & 0.6863$\pm$0.0277 \\
				\midrule
				\multirow{2}{*}{Bean} & ARI & 0.5655 & 0.6060 & 0.5310 & 0.3702$\pm$0.0754 & 0.6295$\pm$0.0001 & 0.4613$\pm$0.0018 & 0.1267$\pm$0.0773 & 0.6254$\pm$0.0206 & \textbf{0.6679}$\pm$0.0001 & 0.6113$\pm$0.0179 \\
				& NMI & 0.6907 & 0.7290 & 0.6585 & 0.5428$\pm$0.0544 & \textbf{0.7325}$\pm$0.0001 & 0.5753$\pm$0.0049 & 0.2315$\pm$0.0835 & 0.7313$\pm$0.0070 & 0.7275$\pm$0.0000 & 0.7150$\pm$0.0119 \\
				\midrule
				\multirow{2}{*}{HTRU2} & ARI & 0.4838 & ME & ME & 0.2887$\pm$0.2177 & 0.4854$\pm$0.0000 & 0.5406$\pm$0.0000 & 0.0130$\pm$0.0181 & 0.5272$\pm$0.0000 & \textbf{0.6049}$\pm$0.0002 & 0.4636$\pm$0.0000 \\
				& NMI & 0.3342 & ME & ME & 0.1887$\pm$0.1483 & 0.3354$\pm$0.0000 & 0.3575$\pm$0.0000 & 0.0136$\pm$0.0120 & 0.3406$\pm$0.0000 & \textbf{0.3970}$\pm$0.0002 & 0.3098$\pm$0.0000 \\
				\midrule
				\multirow{2}{*}{Average (17)} & ARI & \textbf{0.5998} & 0.5550 & 0.5792 & 0.4106 & 0.5441 & 0.4644 & 0.1578 & 0.5022 & 0.4343 & 0.4983 \\
				& NMI & \textbf{0.6613} & 0.6213 & 0.6285 & 0.5135 & 0.6179 & 0.5547 & 0.2676 & 0.5892 & 0.5112 & 0.5746 \\
				\bottomrule
		\end{tabular}}
	\end{threeparttable}
	\vspace{0.2em}
	{\footnotesize
		Deterministic methods are reported by their unique outputs. Stochastic methods are reported as mean $\pm$ sample standard deviation over 20 independent runs. The best mean value for each dataset and metric is highlighted in bold.
		
		ME denotes memory exceeded. The Average row reports the mean of the dataset-level means over the 17 complete cases used in the common statistical comparison; standard deviations are not averaged across datasets.
	}
\end{table}

Among the region-based baselines, GBCT forms a reduced set of granular balls and clusters from relations among those balls, whereas MDL-GBTRSC retains the original observations as graph vertices and transfers regional information to sample-level affinities~\citep{Xia2025GBCT}. On the 17 complete real datasets, their mean ARI/NMI values are $0.1578/0.2676$ and $0.5998/0.6613$, respectively. MDMSC is substantially closer to MDL-GBTRSC, with means of $0.5022/0.5892$ and average ranks of $5.12/4.12$, and it gives the strongest result on Seeds; GMM-SC likewise attains the highest NMI on Iris. These results illustrate how different uses of regional structure lead to distinct behavior across datasets.

\subsubsection{Synthetic benchmark results}
\label{subsec:synthetic_results}

Table~\ref{tab:synthetic_ari_nmi_comparison} reports the comparison on the 20 synthetic datasets. MDL-GBTRSC attains mean ARI/NMI values of $0.9618/0.9688$, compared with the next-highest mean values of $0.7921$ for ARI and $0.8306$ for NMI. It is best or tied for the best value on 15 of the 20 datasets for both metrics and reaches $\mathrm{ARI}=\mathrm{NMI}=1$ on 12 datasets: zelnik6, zelnik1, 3-spiral, db2, jain, zelnik4, target, chainlink, fourty, N, A, and banana. This behavior is also reflected in the average ranks, where MDL-GBTRSC is close to rank two despite the presence of ten competing methods and numerous tied perfect results.

Exact recovery occurs on 3-spiral, db2, jain, target, chainlink, N, A, and banana, whose controlled geometries include intertwined, nested, or strongly non-convex structures, and also on fourty with 40 compact clusters. On aggregation, diamond9, and cure-t2-4k, the corresponding ARI/NMI values remain near one at $0.9783/0.9754$, $0.9992/0.9990$, and $0.9952/0.9886$, despite touching boundaries, bridging relations, overlap, or noise. These cases are consistent with the intended interaction between reciprocal continuity and leaf-scale affinity regularization; their individual contributions are examined separately in the later ablation study.

The highest synthetic scores on the remaining five datasets are obtained by other methods: SC-kNN on blobs, STSC-FK on zelnik2, MDMSC on wingnut and complex8, and GBCT on cluto-t8-8k. According to Table~\ref{tab:synthetic_dataset_details}, these cases include sparse-density, shape-diverse, and noisy structures. Their results indicate that different structural representations emphasize different aspects of the controlled geometries, with direct granular-ball relations or alternative representative structures proving competitive in several cases.

\begin{table}[htbp]
	\centering
	\scriptsize
	\setlength{\tabcolsep}{2.0pt}
	\renewcommand{\arraystretch}{1.06}
	\caption{Dataset-level comparison of ARI and NMI on the synthetic datasets.}
	\label{tab:synthetic_ari_nmi_comparison}
	\begin{threeparttable}
		\resizebox{\columnwidth}{!}{%
			\begin{tabular}{llcccccccccc}
				\toprule
				\multirow{2}{*}{Dataset} & \multirow{2}{*}{Metric}
				& \multirow{2}{*}{Ours}
				& \multicolumn{5}{c}{Spectral baselines}
				& \multirow{2}{*}{\makecell{GBCT\\(TNNLS 2025)}}
				& \multirow{2}{*}{\makecell{MDMSC\\(AAAI 2025)}}
				& \multirow{2}{*}{\makecell{GBSC\\(TKDE 2023)}}
				& \multirow{2}{*}{\makecell{GMM-SC\\(TNNLS 2025)}} \\
				\cmidrule(lr){4-8}
				& & & STSC-FK & N-DSC & FastSC & SC-kNN & SC-RBF & & & & \\
				\midrule
				\multirow{2}{*}{zelnik6} & ARI & \textbf{1.0000} & \textbf{1.0000} & 0.5899 & 0.4991$\pm$0.1825 & 0.5775$\pm$0.0047 & 0.6743$\pm$0.0035 & 0.6353$\pm$0.0383 & 0.6196$\pm$0.0418 & 0.6408$\pm$0.0056 & 0.7026$\pm$0.0067 \\
				& NMI & \textbf{1.0000} & \textbf{1.0000} & 0.6000 & 0.5148$\pm$0.1522 & 0.5192$\pm$0.0028 & 0.6610$\pm$0.0025 & 0.6392$\pm$0.0238 & 0.6467$\pm$0.0359 & 0.6340$\pm$0.0049 & 0.6927$\pm$0.0058 \\
				\midrule
				\multirow{2}{*}{zelnik1} & ARI & \textbf{1.0000} & \textbf{1.0000} & 0.1736 & 0.4854$\pm$0.2171 & 0.7406$\pm$0.2680 & 0.0389$\pm$0.0036 & 0.6930$\pm$0.2321 & 0.5531$\pm$0.0511 & 0.5151$\pm$0.0000 & 0.5091$\pm$0.0023 \\
				& NMI & \textbf{1.0000} & \textbf{1.0000} & 0.3027 & 0.6228$\pm$0.1750 & 0.8347$\pm$0.1702 & 0.1506$\pm$0.0029 & 0.7883$\pm$0.1598 & 0.6962$\pm$0.0264 & 0.6672$\pm$0.0000 & 0.6636$\pm$0.0046 \\
				\midrule
				\multirow{2}{*}{blobs} & ARI & 0.9215 & 0.9027 & 0.4784 & 0.7508$\pm$0.1733 & \textbf{0.9311}$\pm$0.0000 & 0.8902$\pm$0.0047 & 0.5795$\pm$0.2729 & 0.8547$\pm$0.0000 & 0.7889$\pm$0.0000 & 0.7573$\pm$0.0295 \\
				& NMI & 0.8853 & 0.8615 & 0.5348 & 0.7646$\pm$0.1104 & \textbf{0.8979}$\pm$0.0000 & 0.8483$\pm$0.0055 & 0.6456$\pm$0.2192 & 0.8167$\pm$0.0000 & 0.7811$\pm$0.0000 & 0.7370$\pm$0.0357 \\
				\midrule
				\multirow{2}{*}{zelnik2} & ARI & 0.9901 & \textbf{1.0000} & 0.5790 & 0.4495$\pm$0.1999 & 0.7292$\pm$0.0042 & 0.4818$\pm$0.0162 & 0.1152$\pm$0.1868 & 0.4776$\pm$0.0000 & 0.1614$\pm$0.0213 & 0.4951$\pm$0.1262 \\
				& NMI & 0.9831 & \textbf{1.0000} & 0.7104 & 0.5177$\pm$0.1759 & 0.7662$\pm$0.0055 & 0.5543$\pm$0.0143 & 0.2122$\pm$0.1774 & 0.5561$\pm$0.0000 & 0.3363$\pm$0.0207 & 0.5999$\pm$0.1031 \\
				\midrule
				\multirow{2}{*}{3-spiral} & ARI & \textbf{1.0000} & 0.0643 & 0.0192 & 0.1571$\pm$0.0989 & 0.4550$\pm$0.0000 & -0.0049$\pm$0.0010 & 0.9078$\pm$0.1547 & \textbf{1.0000}$\pm$0.0000 & 0.0040$\pm$0.0000 & -0.0005$\pm$0.0014 \\
				& NMI & \textbf{1.0000} & 0.0728 & 0.0249 & 0.2315$\pm$0.0888 & 0.5043$\pm$0.0000 & 0.0014$\pm$0.0009 & 0.9260$\pm$0.1150 & \textbf{1.0000}$\pm$0.0000 & 0.0116$\pm$0.0000 & 0.0052$\pm$0.0012 \\
				\midrule
				\multirow{2}{*}{db2} & ARI & \textbf{1.0000} & 0.2557 & 0.3045 & 0.2775$\pm$0.1317 & 0.3304$\pm$0.0028 & 0.2059$\pm$0.0000 & 0.7629$\pm$0.2650 & 0.9027$\pm$0.0950 & 0.2377$\pm$0.0000 & 0.2141$\pm$0.0050 \\
				& NMI & \textbf{1.0000} & 0.5326 & 0.5609 & 0.5005$\pm$0.1166 & 0.5776$\pm$0.0066 & 0.4208$\pm$0.0000 & 0.8685$\pm$0.1432 & 0.9085$\pm$0.0548 & 0.5227$\pm$0.0000 & 0.4306$\pm$0.0020 \\
				\midrule
				\multirow{2}{*}{jain} & ARI & \textbf{1.0000} & 0.5071 & 0.1642 & 0.1048$\pm$0.2557 & 0.5988$\pm$0.0040 & 0.5612$\pm$0.0000 & 0.4648$\pm$0.3256 & \textbf{1.0000}$\pm$0.0000 & 0.2563$\pm$0.0000 & -0.0281$\pm$0.0360 \\
				& NMI & \textbf{1.0000} & 0.5003 & 0.3008 & 0.2140$\pm$0.2154 & 0.5629$\pm$0.0029 & 0.5365$\pm$0.0000 & 0.4529$\pm$0.3297 & \textbf{1.0000}$\pm$0.0000 & 0.2463$\pm$0.0000 & 0.1841$\pm$0.0241 \\
				\midrule
				\multirow{2}{*}{zelnik4} & ARI & \textbf{1.0000} & \textbf{1.0000} & \textbf{1.0000} & 0.7777$\pm$0.1531 & \textbf{1.0000}$\pm$0.0000 & \textbf{1.0000}$\pm$0.0000 & 0.9026$\pm$0.1527 & \textbf{1.0000}$\pm$0.0000 & 0.7532$\pm$0.0000 & 0.9675$\pm$0.1000 \\
				& NMI & \textbf{1.0000} & \textbf{1.0000} & \textbf{1.0000} & 0.8423$\pm$0.1112 & \textbf{1.0000}$\pm$0.0000 & \textbf{1.0000}$\pm$0.0000 & 0.9530$\pm$0.0736 & \textbf{1.0000}$\pm$0.0000 & 0.8689$\pm$0.0000 & 0.9843$\pm$0.0482 \\
				\midrule
				\multirow{2}{*}{target} & ARI & \textbf{1.0000} & 0.7031 & 0.3552 & 0.3046$\pm$0.0902 & 0.3853$\pm$0.0001 & 0.6351$\pm$0.0006 & 0.6547$\pm$0.0140 & 0.6688$\pm$0.0000 & 0.0528$\pm$0.0070 & 0.6132$\pm$0.1112 \\
				& NMI & \textbf{1.0000} & 0.7467 & 0.5597 & 0.4433$\pm$0.1149 & 0.5704$\pm$0.0000 & 0.6364$\pm$0.0005 & 0.6505$\pm$0.0094 & 0.6659$\pm$0.0000 & 0.1835$\pm$0.0103 & 0.6582$\pm$0.0449 \\
				\midrule
				\multirow{2}{*}{aggregation} & ARI & \textbf{0.9783} & 0.9586 & 0.7252 & 0.6406$\pm$0.1475 & 0.5071$\pm$0.0728 & 0.5824$\pm$0.0306 & 0.8838$\pm$0.0565 & 0.9761$\pm$0.0000 & 0.8314$\pm$0.0238 & 0.8115$\pm$0.0719 \\
				& NMI & \textbf{0.9754} & 0.9628 & 0.8583 & 0.8040$\pm$0.0799 & 0.7470$\pm$0.0422 & 0.7471$\pm$0.0106 & 0.9265$\pm$0.0345 & 0.9726$\pm$0.0000 & 0.8935$\pm$0.0082 & 0.8858$\pm$0.0323 \\
				\midrule
				\multirow{2}{*}{chainlink} & ARI & \textbf{1.0000} & \textbf{1.0000} & \textbf{1.0000} & 0.0581$\pm$0.0636 & \textbf{1.0000}$\pm$0.0000 & 0.1226$\pm$0.0014 & \textbf{1.0000}$\pm$0.0000 & \textbf{1.0000}$\pm$0.0000 & 0.0037$\pm$0.0004 & 0.6527$\pm$0.0599 \\
				& NMI & \textbf{1.0000} & \textbf{1.0000} & \textbf{1.0000} & 0.0557$\pm$0.0667 & \textbf{1.0000}$\pm$0.0000 & 0.0911$\pm$0.0010 & \textbf{1.0000}$\pm$0.0000 & \textbf{1.0000}$\pm$0.0000 & 0.0034$\pm$0.0003 & 0.5508$\pm$0.0778 \\
				\midrule
				\multirow{2}{*}{fourty} & ARI & \textbf{1.0000} & \textbf{1.0000} & \textbf{1.0000} & 0.9768$\pm$0.0211 & 0.4362$\pm$0.0780 & 0.3362$\pm$0.0151 & 0.9939$\pm$0.0204 & 0.9674$\pm$0.0000 & 0.9546$\pm$0.0000 & 0.9552$\pm$0.0145 \\
				& NMI & \textbf{1.0000} & \textbf{1.0000} & \textbf{1.0000} & 0.9948$\pm$0.0047 & 0.8218$\pm$0.0291 & 0.7424$\pm$0.0100 & 0.9988$\pm$0.0038 & 0.9930$\pm$0.0000 & 0.9865$\pm$0.0000 & 0.9840$\pm$0.0027 \\
				\midrule
				\multirow{2}{*}{wingnut} & ARI & 0.9610 & 0.8269 & 0.0251 & 0.3032$\pm$0.2513 & 0.8531$\pm$0.0016 & 0.4087$\pm$0.0000 & 0.6341$\pm$0.4761 & \textbf{1.0000}$\pm$0.0000 & 0.9456$\pm$0.0000 & 0.5454$\pm$0.1593 \\
				& NMI & 0.9205 & 0.7340 & 0.0189 & 0.2814$\pm$0.2439 & 0.7663$\pm$0.0021 & 0.3197$\pm$0.0000 & 0.6457$\pm$0.4467 & \textbf{1.0000}$\pm$0.0000 & 0.8951$\pm$0.0000 & 0.4482$\pm$0.1458 \\
				\midrule
				\multirow{2}{*}{N} & ARI & \textbf{1.0000} & \textbf{1.0000} & \textbf{1.0000} & 0.3694$\pm$0.1648 & 0.4716$\pm$0.1118 & 0.3867$\pm$0.0020 & \textbf{1.0000}$\pm$0.0000 & \textbf{1.0000}$\pm$0.0000 & 0.6230$\pm$0.0014 & 0.3996$\pm$0.0748 \\
				& NMI & \textbf{1.0000} & \textbf{1.0000} & \textbf{1.0000} & 0.5560$\pm$0.1381 & 0.6848$\pm$0.0810 & 0.5553$\pm$0.0049 & \textbf{1.0000}$\pm$0.0000 & \textbf{1.0000}$\pm$0.0000 & 0.7990$\pm$0.0004 & 0.5514$\pm$0.0904 \\
				\midrule
				\multirow{2}{*}{A} & ARI & \textbf{1.0000} & 0.5710 & \textbf{1.0000} & 0.2215$\pm$0.0546 & 0.3011$\pm$0.0386 & 0.3557$\pm$0.0104 & \textbf{1.0000}$\pm$0.0000 & 0.5917$\pm$0.1535 & 0.4949$\pm$0.0000 & 0.3628$\pm$0.0533 \\
				& NMI & \textbf{1.0000} & 0.7619 & \textbf{1.0000} & 0.4610$\pm$0.0550 & 0.5796$\pm$0.0272 & 0.5838$\pm$0.0116 & \textbf{1.0000}$\pm$0.0000 & 0.7460$\pm$0.0676 & 0.7162$\pm$0.0000 & 0.6043$\pm$0.0332 \\
				\midrule
				\multirow{2}{*}{complex8} & ARI & 0.7318 & 0.6391 & 0.7001 & 0.3020$\pm$0.0830 & 0.5859$\pm$0.0330 & 0.4165$\pm$0.0075 & 0.7887$\pm$0.1643 & \textbf{0.8809}$\pm$0.0611 & 0.5013$\pm$0.0035 & 0.5181$\pm$0.0342 \\
				& NMI & 0.8277 & 0.8003 & 0.8032 & 0.4915$\pm$0.0702 & 0.7594$\pm$0.0153 & 0.6238$\pm$0.0067 & 0.8752$\pm$0.0756 & \textbf{0.9158}$\pm$0.0345 & 0.6989$\pm$0.0026 & 0.6856$\pm$0.0151 \\
				\midrule
				\multirow{2}{*}{diamond9} & ARI & \textbf{0.9992} & 0.9985 & 0.8307 & 0.4555$\pm$0.0738 & 0.9985$\pm$0.0000 & 0.7417$\pm$0.0157 & 0.8130$\pm$0.0941 & 0.8724$\pm$0.0409 & 0.9873$\pm$0.0000 & 0.8509$\pm$0.0191 \\
				& NMI & \textbf{0.9990} & 0.9981 & 0.9249 & 0.6460$\pm$0.0601 & 0.9981$\pm$0.0000 & 0.8303$\pm$0.0109 & 0.9233$\pm$0.0380 & 0.9391$\pm$0.0179 & 0.9844$\pm$0.0000 & 0.8866$\pm$0.0121 \\
				\midrule
				\multirow{2}{*}{cure-t2-4k} & ARI & \textbf{0.9952} & 0.6875 & 0.6730 & 0.2740$\pm$0.0960 & 0.9828$\pm$0.0000 & 0.5285$\pm$0.0215 & 0.7005$\pm$0.0782 & 0.0349$\pm$0.0000 & 0.6672$\pm$0.0002 & 0.5743$\pm$0.0793 \\
				& NMI & \textbf{0.9886} & 0.8480 & 0.8590 & 0.4350$\pm$0.0826 & 0.9697$\pm$0.0000 & 0.7571$\pm$0.0257 & 0.8045$\pm$0.0372 & 0.1013$\pm$0.0000 & 0.8485$\pm$0.0000 & 0.7915$\pm$0.0479 \\
				\midrule
				\multirow{2}{*}{banana} & ARI & \textbf{1.0000} & \textbf{1.0000} & \textbf{1.0000} & 0.0455$\pm$0.0548 & \textbf{1.0000}$\pm$0.0000 & 0.3756$\pm$0.0000 & 0.3519$\pm$0.2456 & 0.9892$\pm$0.0000 & \textbf{1.0000}$\pm$0.0000 & 0.2264$\pm$0.0003 \\
				& NMI & \textbf{1.0000} & \textbf{1.0000} & \textbf{1.0000} & 0.0465$\pm$0.0760 & \textbf{1.0000}$\pm$0.0000 & 0.2906$\pm$0.0000 & 0.3461$\pm$0.2371 & 0.9757$\pm$0.0000 & \textbf{1.0000}$\pm$0.0000 & 0.1699$\pm$0.0003 \\
				\midrule
				\multirow{2}{*}{cluto-t8-8k} & ARI & 0.6589 & 0.6589 & 0.6118 & 0.1460$\pm$0.0333 & 0.6225$\pm$0.0002 & 0.4271$\pm$0.0074 & \textbf{0.7447}$\pm$0.1671 & 0.4532$\pm$0.0125 & 0.5520$\pm$0.0056 & 0.4980$\pm$0.0324 \\
				& NMI & 0.7974 & 0.7927 & 0.7582 & 0.2843$\pm$0.0405 & 0.7652$\pm$0.0001 & 0.6287$\pm$0.0058 & \textbf{0.8671}$\pm$0.0744 & 0.6149$\pm$0.0178 & 0.7071$\pm$0.0031 & 0.6685$\pm$0.0219 \\
				\midrule
				\multirow{2}{*}{Average (20)} & ARI & \textbf{0.9618} & 0.7887 & 0.6115 & 0.3799 & 0.6753 & 0.4582 & 0.7313 & 0.7921 & 0.5486 & 0.5313 \\
				& NMI & \textbf{0.9688} & 0.8306 & 0.6908 & 0.4854 & 0.7663 & 0.5490 & 0.7762 & 0.8274 & 0.6392 & 0.6091 \\
				\bottomrule
		\end{tabular}}
	\end{threeparttable}
	\vspace{0.2em}
	{\footnotesize
		Deterministic methods are reported by their unique outputs. Stochastic methods are reported as mean $\pm$ sample standard deviation over 20 independent runs. The best mean value for each dataset and metric is highlighted in bold.
		
		The Average row reports the mean of the dataset-level means over all 20 synthetic datasets; standard deviations are not averaged across datasets.
	}
\end{table}

\begin{figure}
	\centering
	\setlength{\subfigcapskip}{0pt}
	\setlength{\subfigtopskip}{0pt}
	\setlength{\subfigbottomskip}{0pt}
	\setlength{\subfigcapmargin}{0pt}
	
	\subfigure[zelnik6]{
		\label{fig:ours_zelnik6}
		\includegraphics[width=0.15\textwidth]{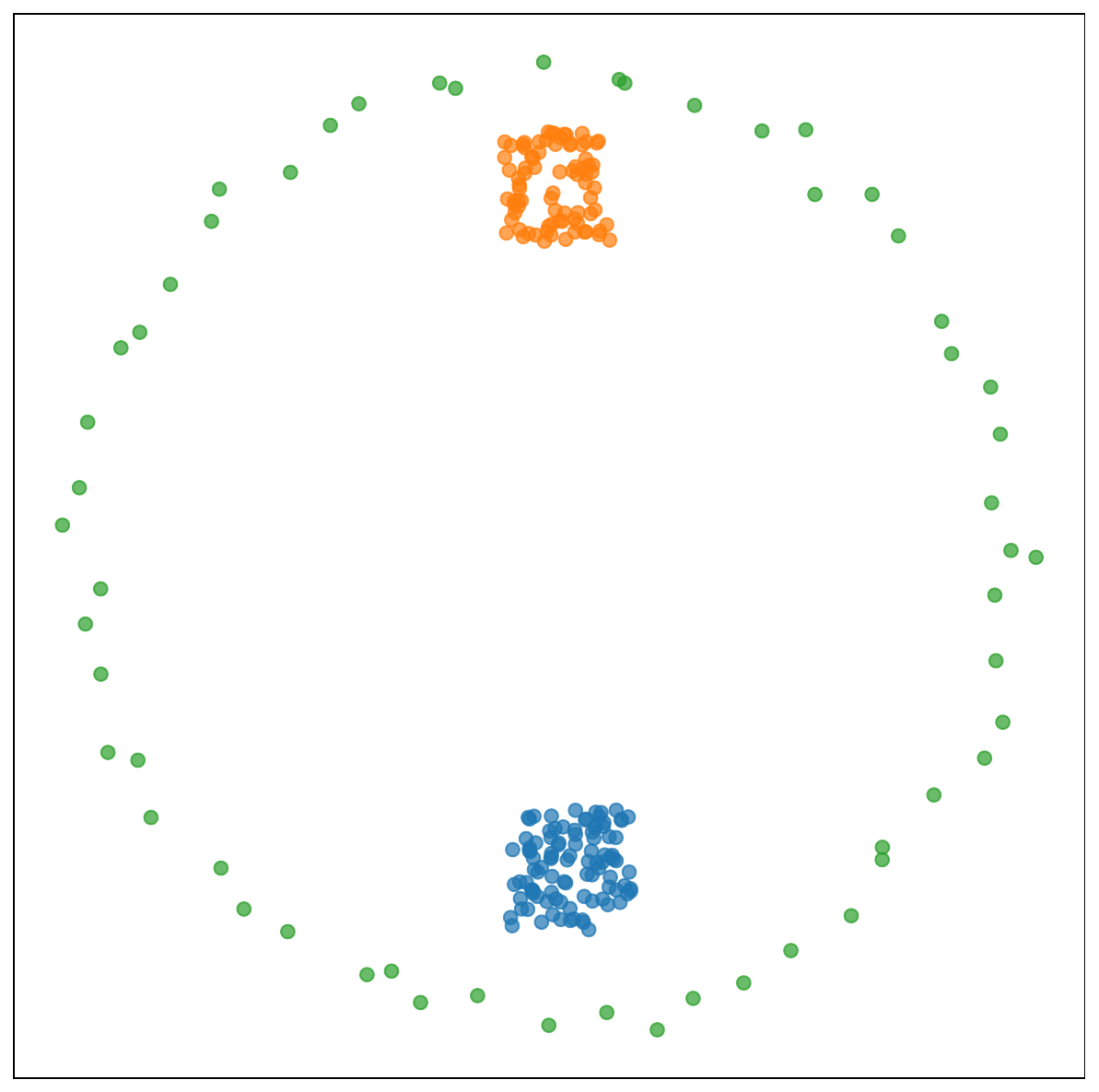}}
	\subfigure[zelnik1]{
		\label{fig:ours_zelnik1}
		\includegraphics[width=0.15\textwidth]{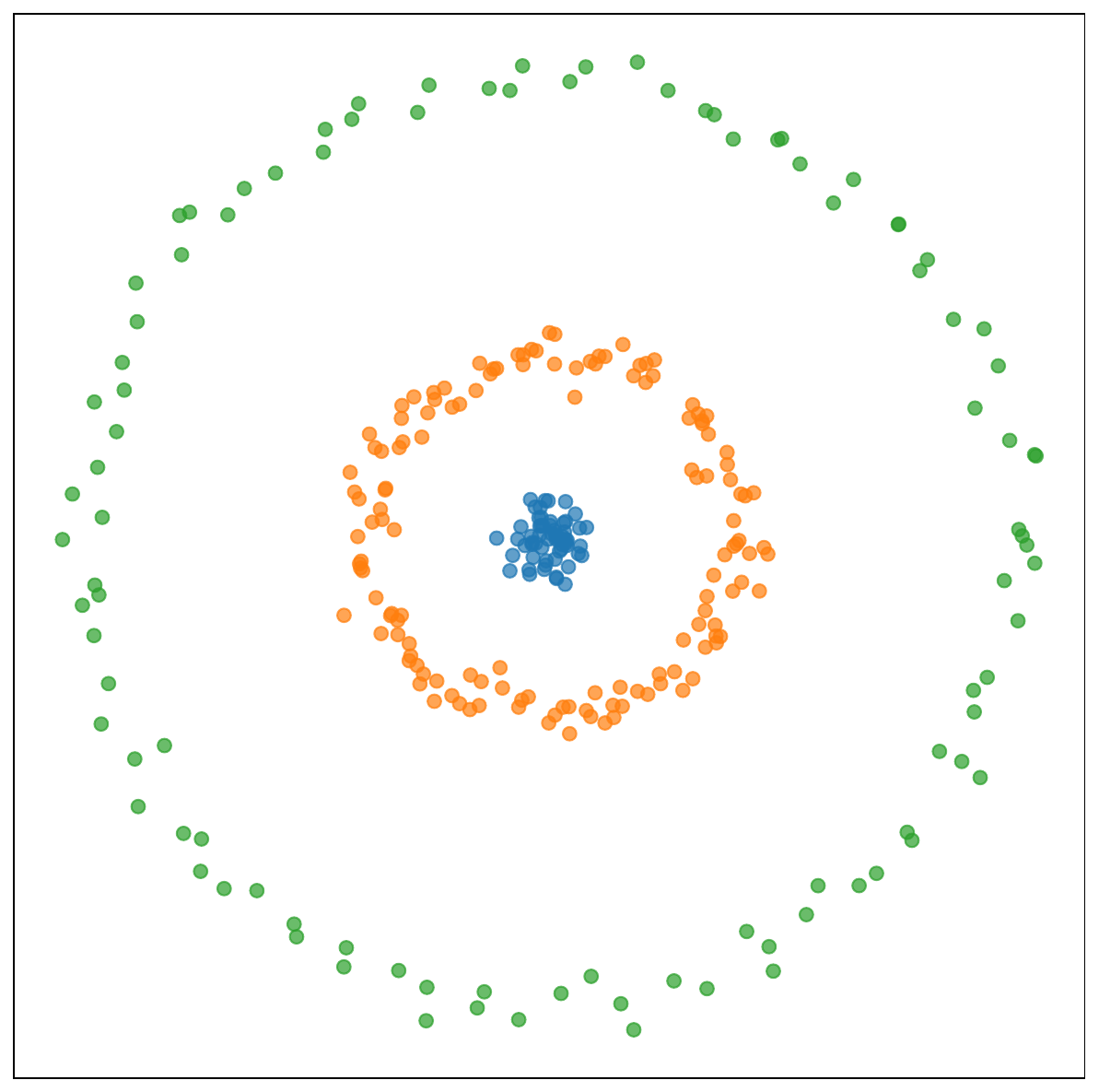}}
	\subfigure[blobs]{
		\label{fig:ours_blobs}
		\includegraphics[width=0.15\textwidth]{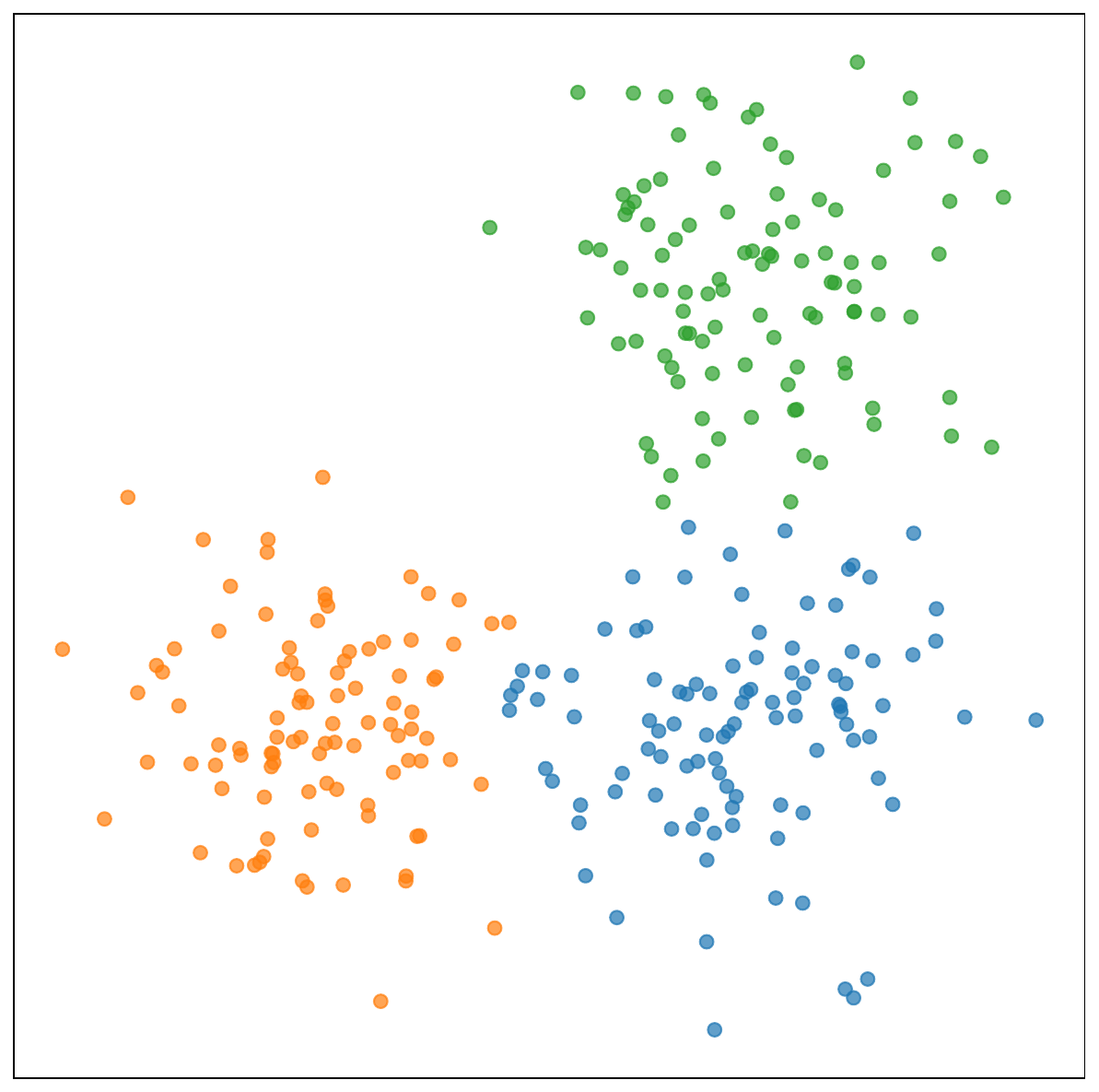}}
	\subfigure[zelnik2]{
		\label{fig:ours_zelnik2}
		\includegraphics[width=0.15\textwidth]{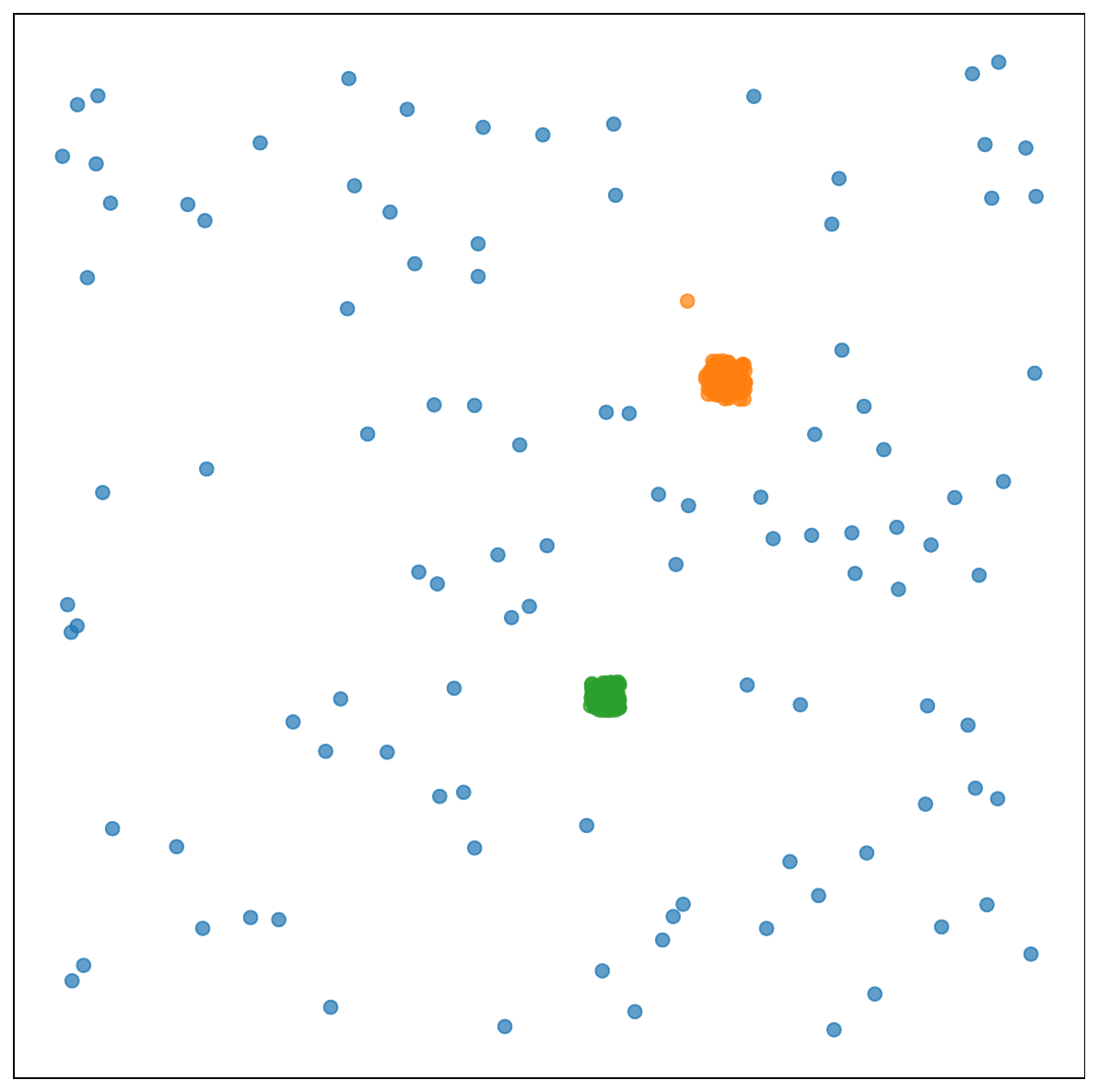}}
	\subfigure[3-spiral]{
		\label{fig:ours_3spiral}
		\includegraphics[width=0.15\textwidth]{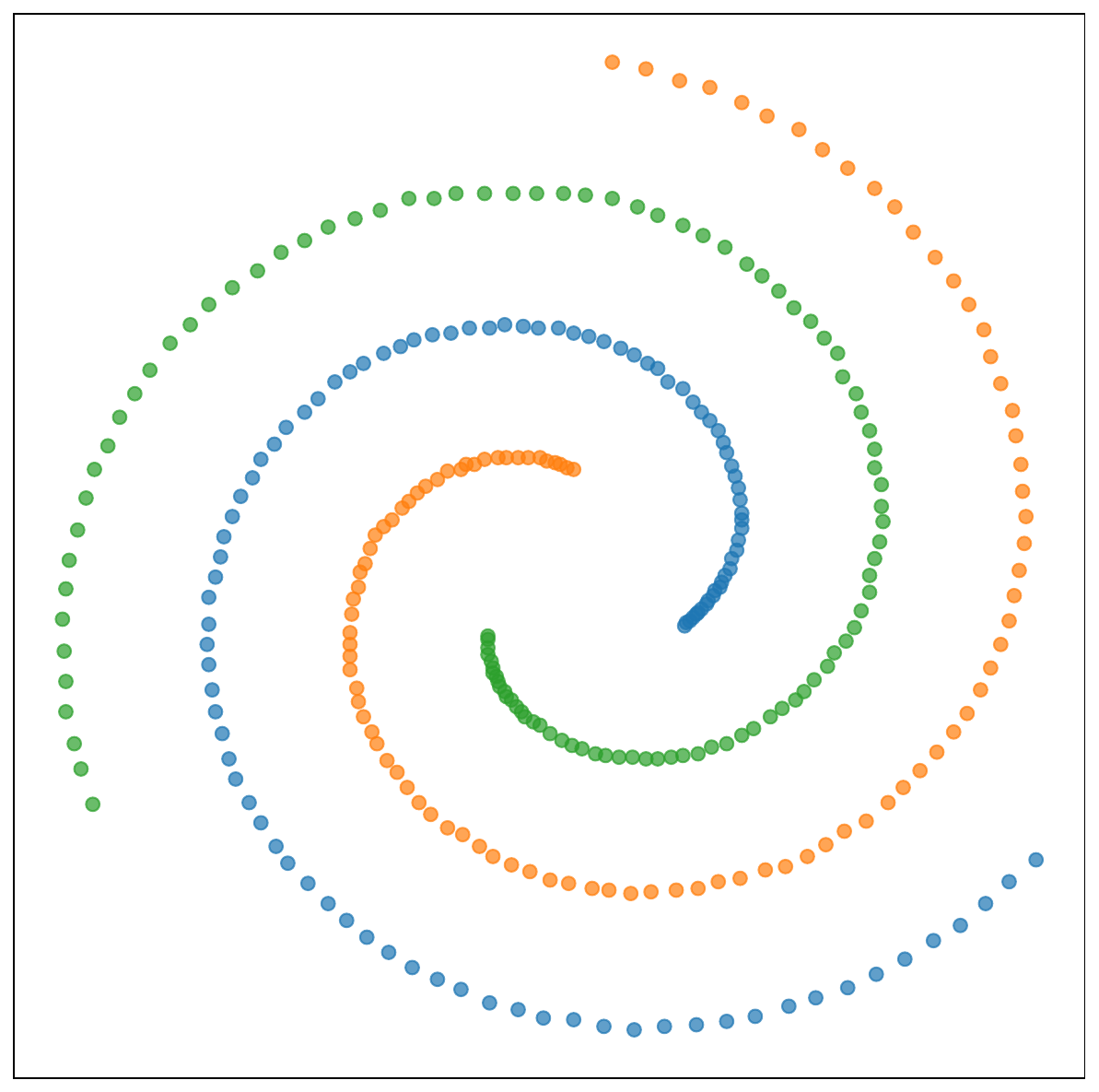}}
	
	\subfigure[db2]{
		\label{fig:ours_db2}
		\includegraphics[width=0.15\textwidth]{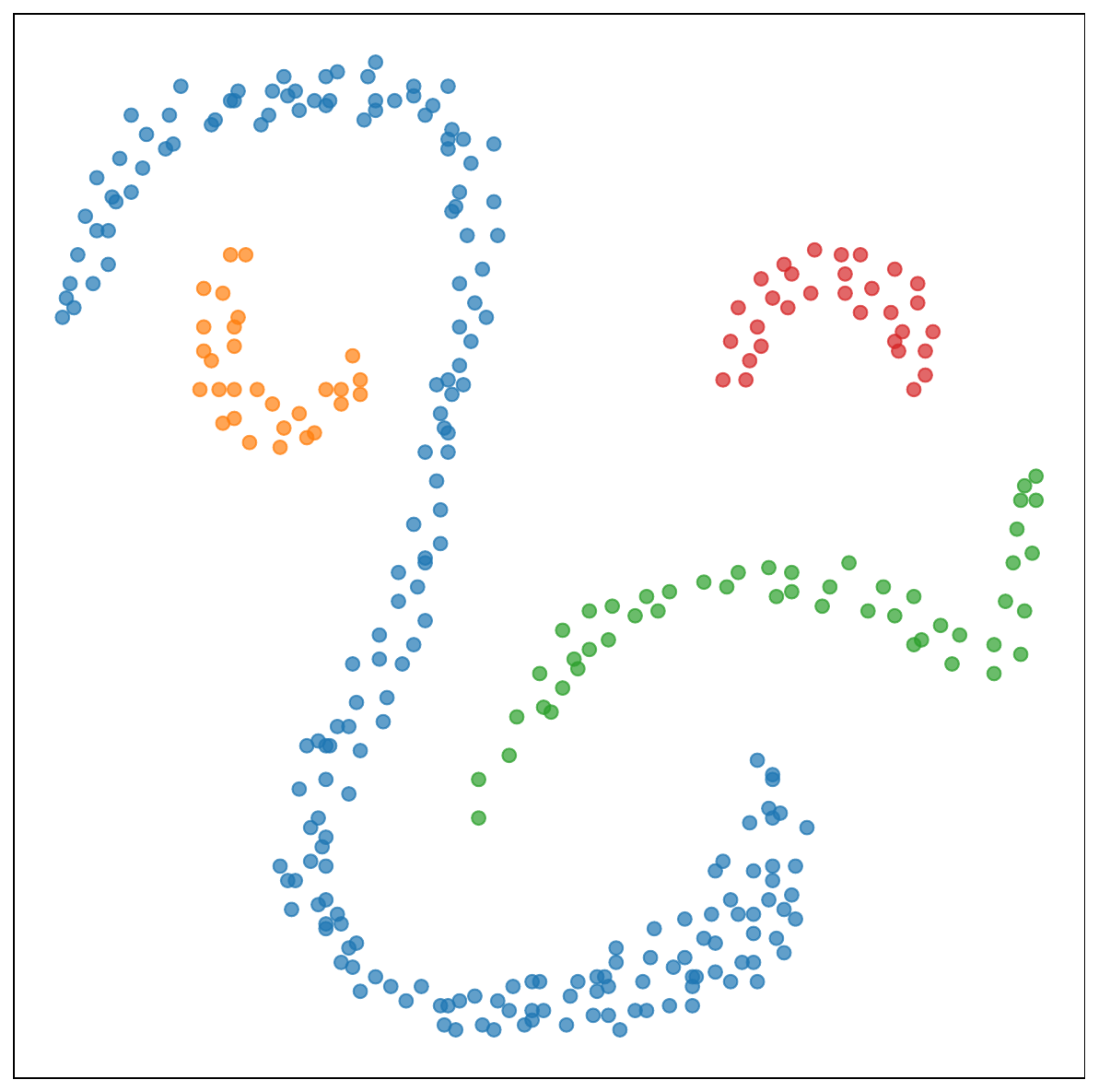}}
	\subfigure[jain]{
		\label{fig:ours_jain}
		\includegraphics[width=0.15\textwidth]{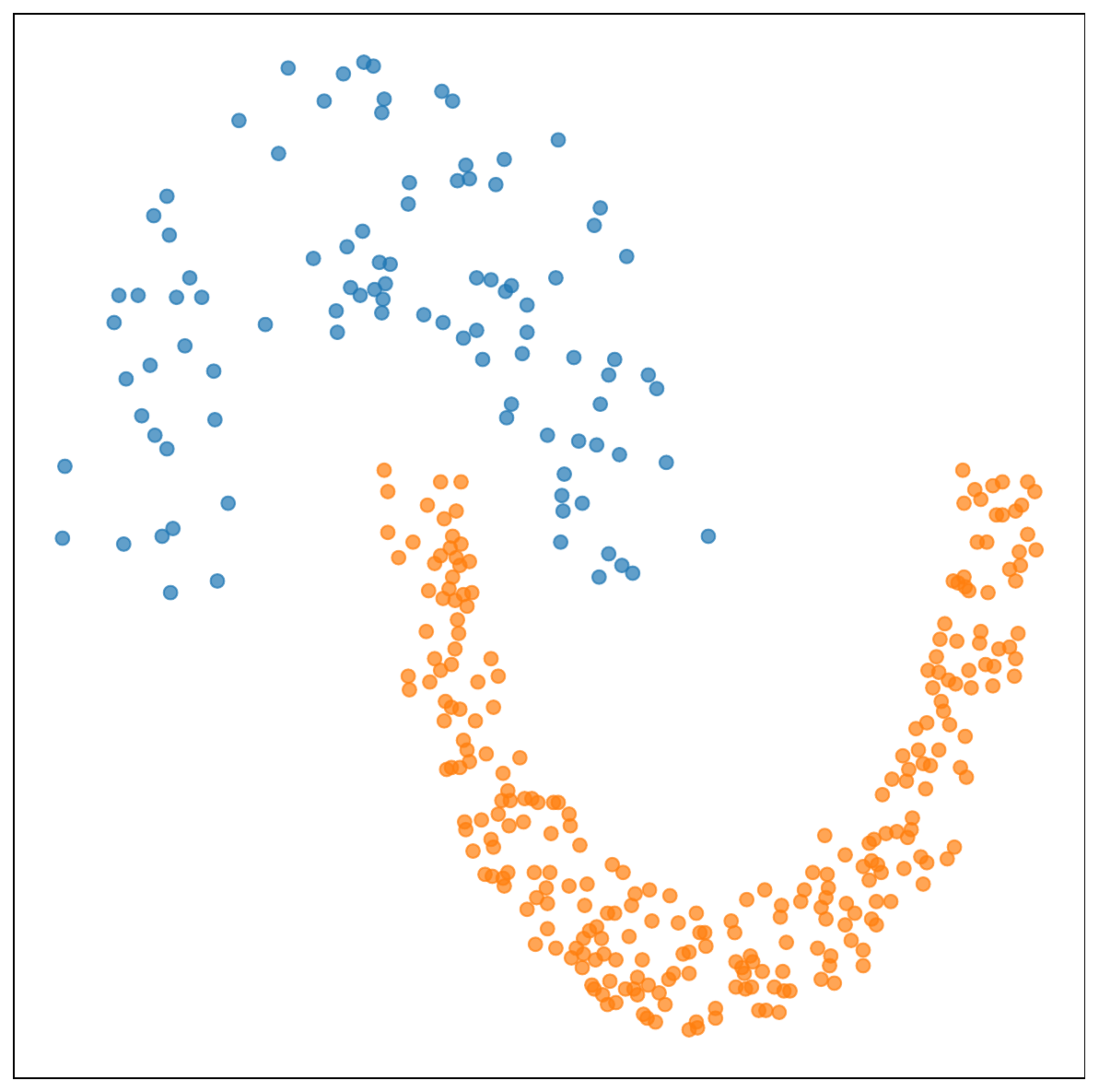}}
	\subfigure[zelnik4]{
		\label{fig:ours_zelnik4}
		\includegraphics[width=0.15\textwidth]{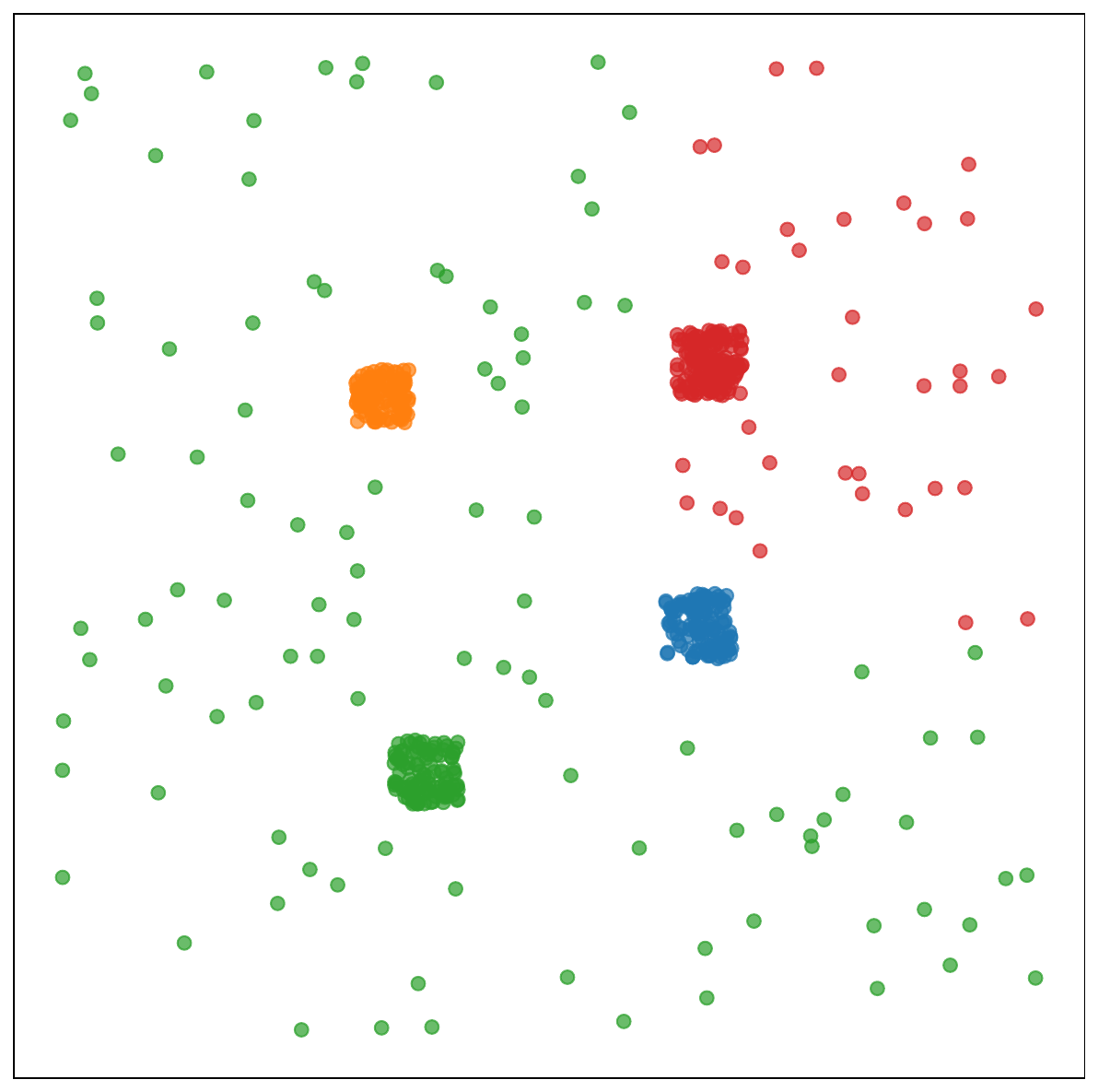}}
	\subfigure[target]{
		\label{fig:ours_target}
		\includegraphics[width=0.15\textwidth]{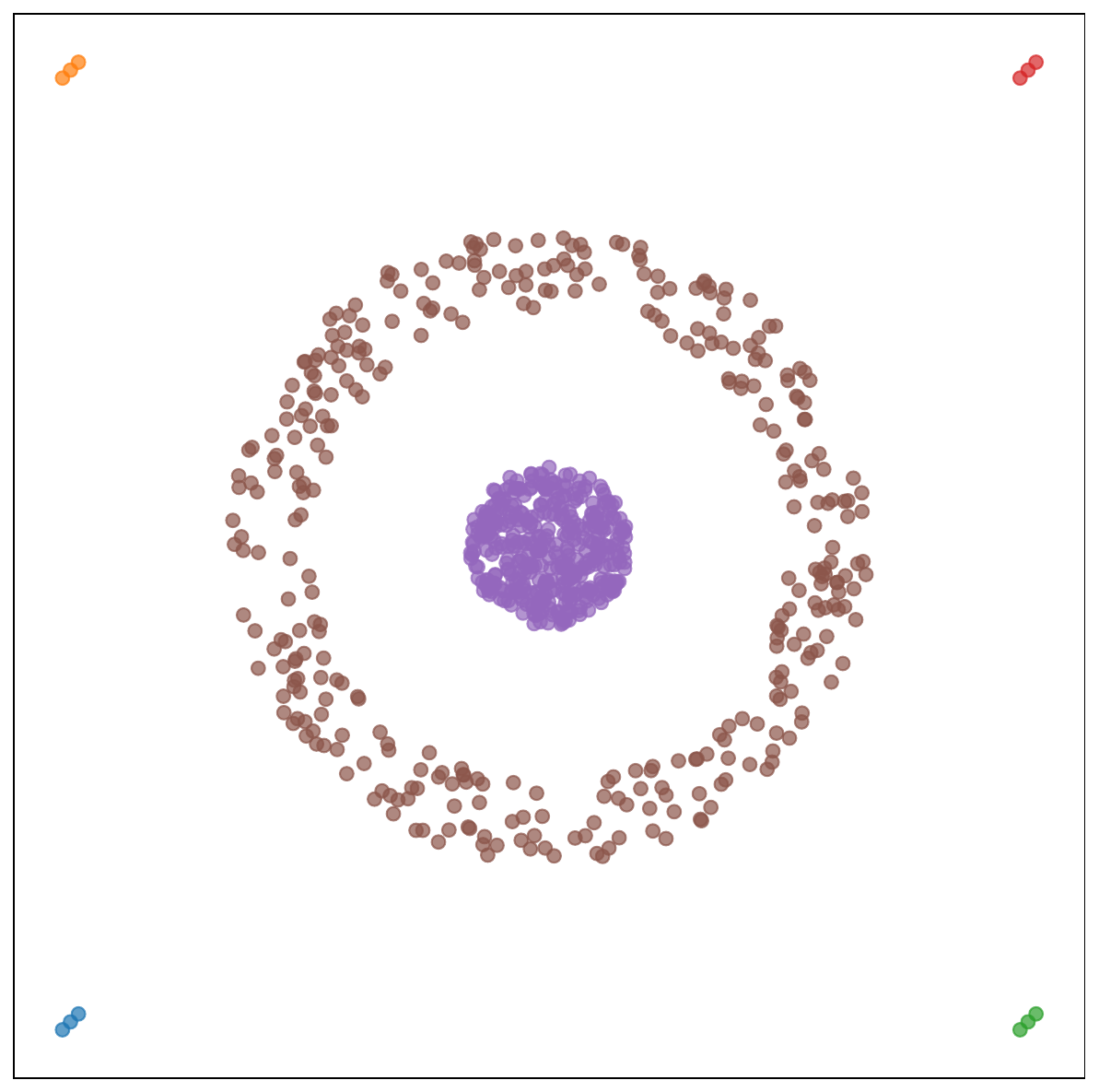}}
	\subfigure[aggregation]{
		\label{fig:ours_aggregation}
		\includegraphics[width=0.15\textwidth]{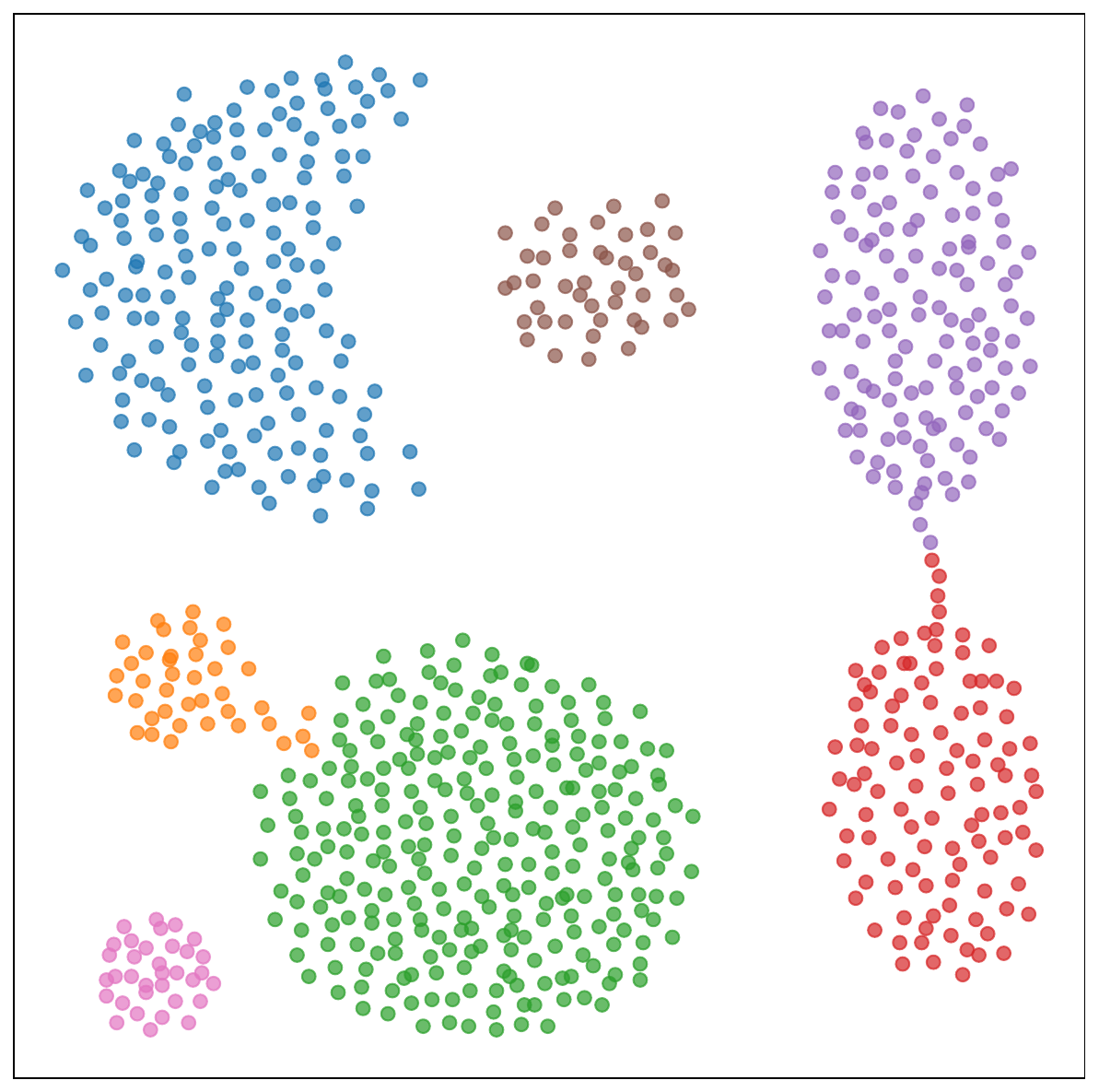}}
	
	\subfigure[chainlink]{
		\label{fig:ours_chainlink}
		\includegraphics[width=0.15\textwidth]{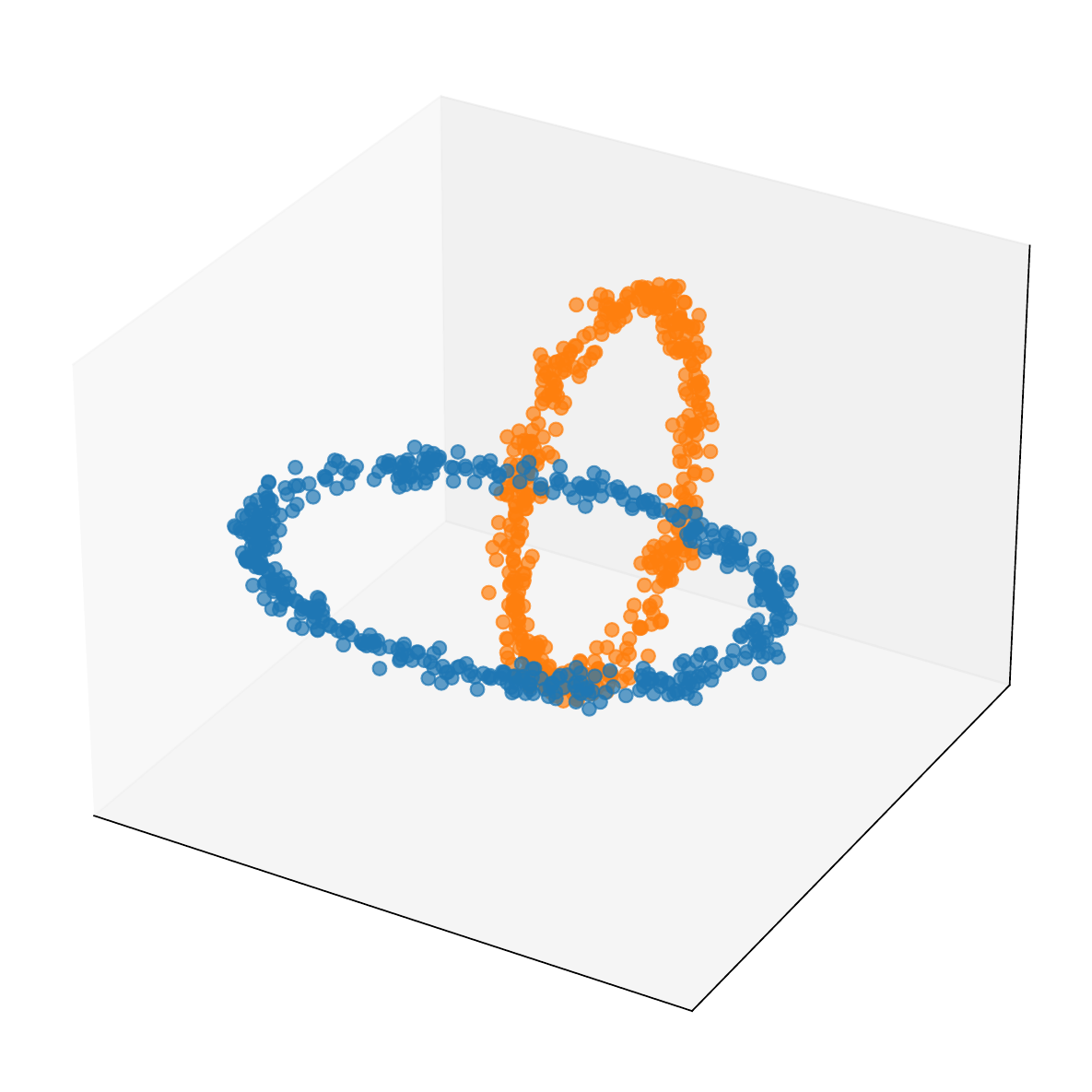}}
	\subfigure[fourty]{
		\label{fig:ours_fourty}
		\includegraphics[width=0.15\textwidth]{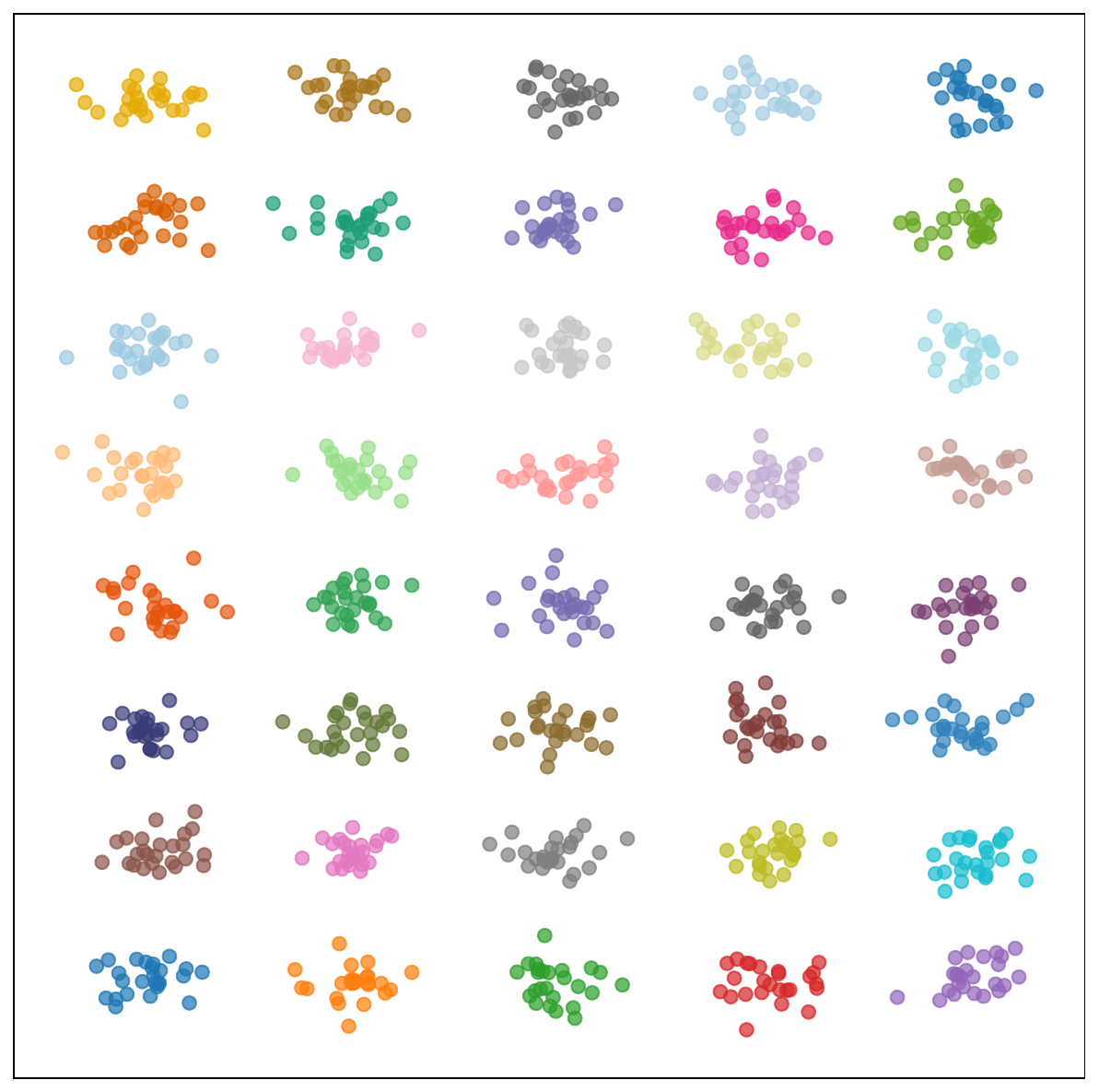}}
	\subfigure[wingnut]{
		\label{fig:ours_wingnut}
		\includegraphics[width=0.15\textwidth]{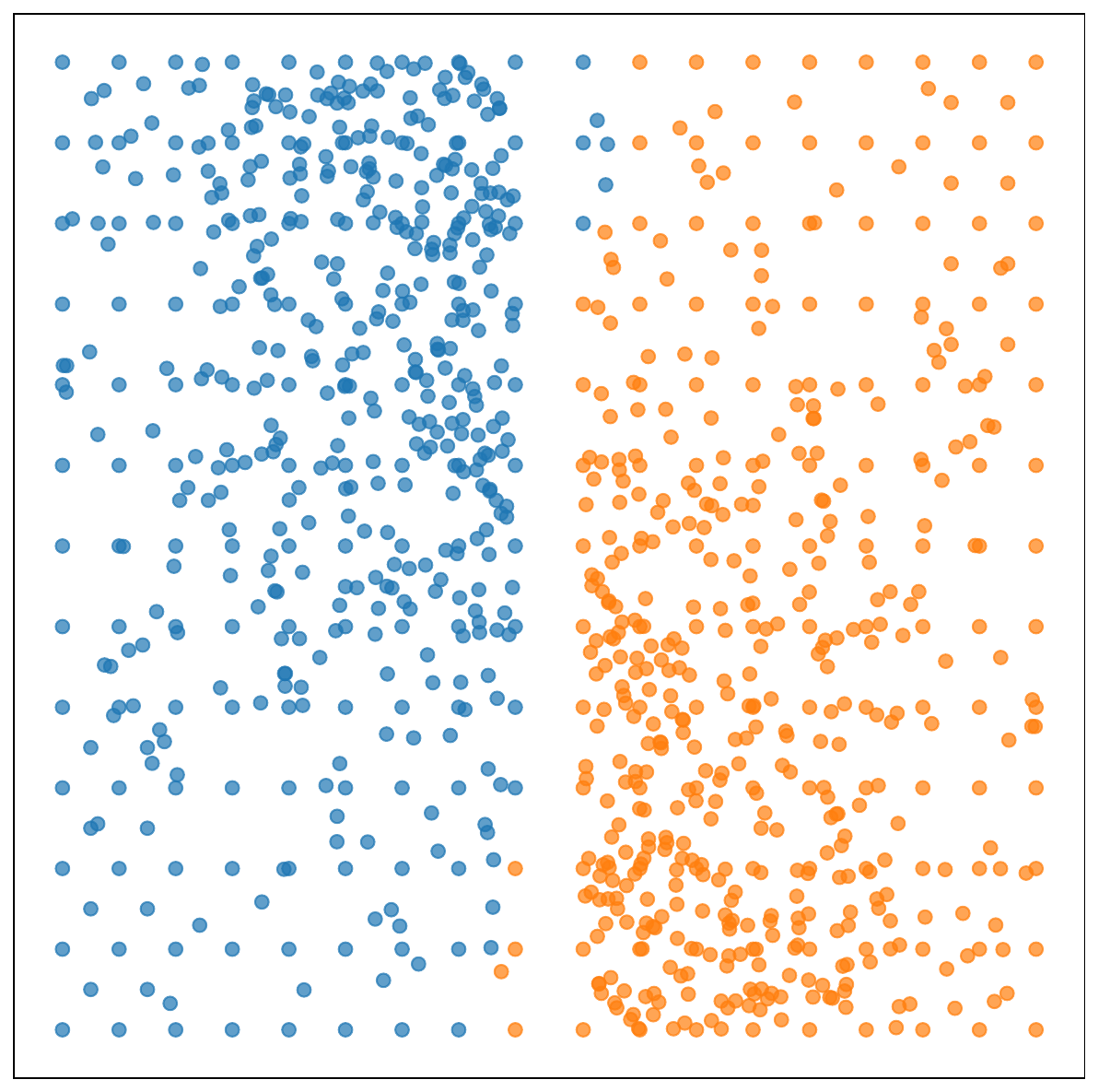}}
	\subfigure[N]{
		\label{fig:ours_n}
		\includegraphics[width=0.15\textwidth]{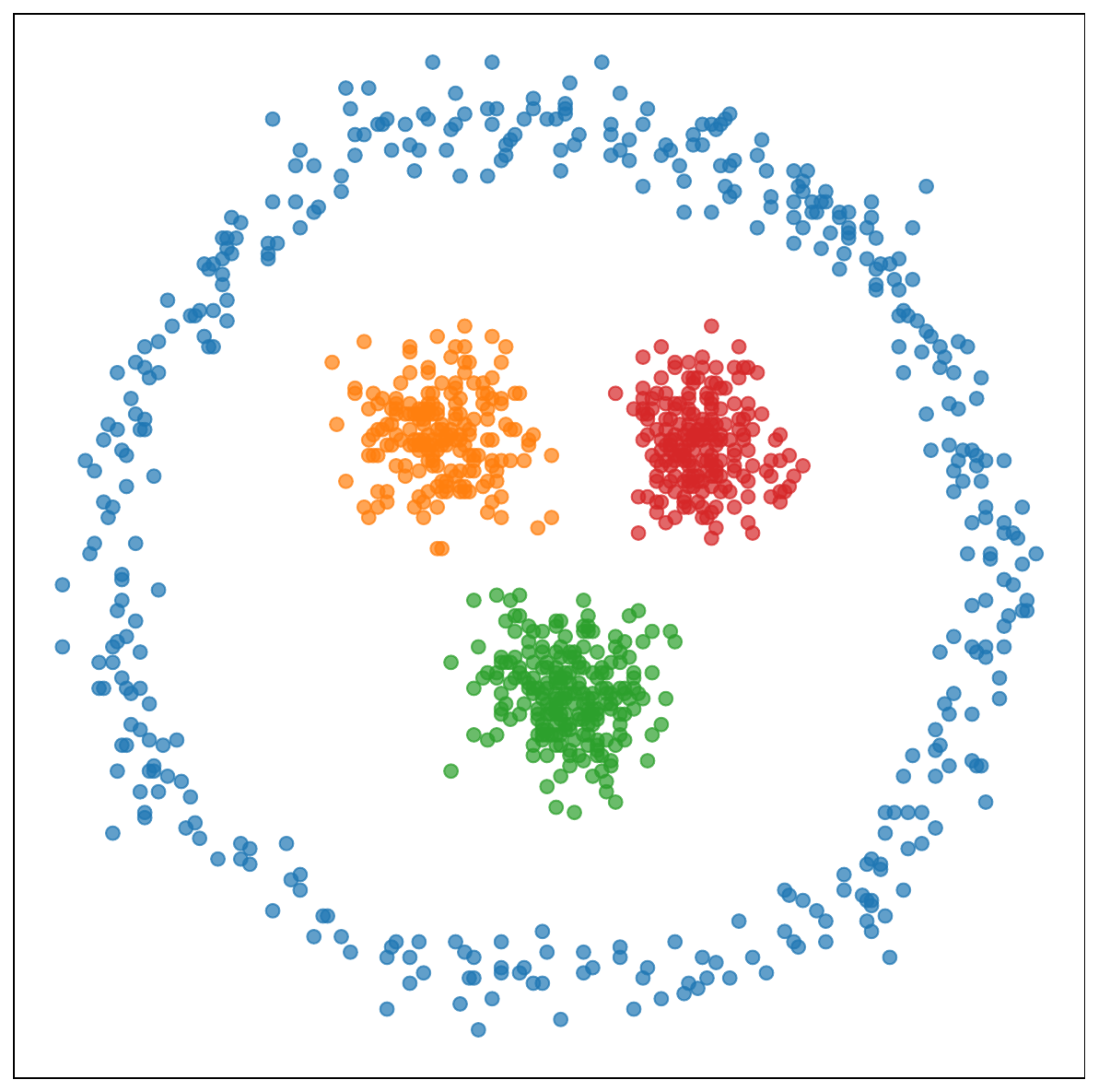}}
	\subfigure[A]{
		\label{fig:ours_a}
		\includegraphics[width=0.15\textwidth]{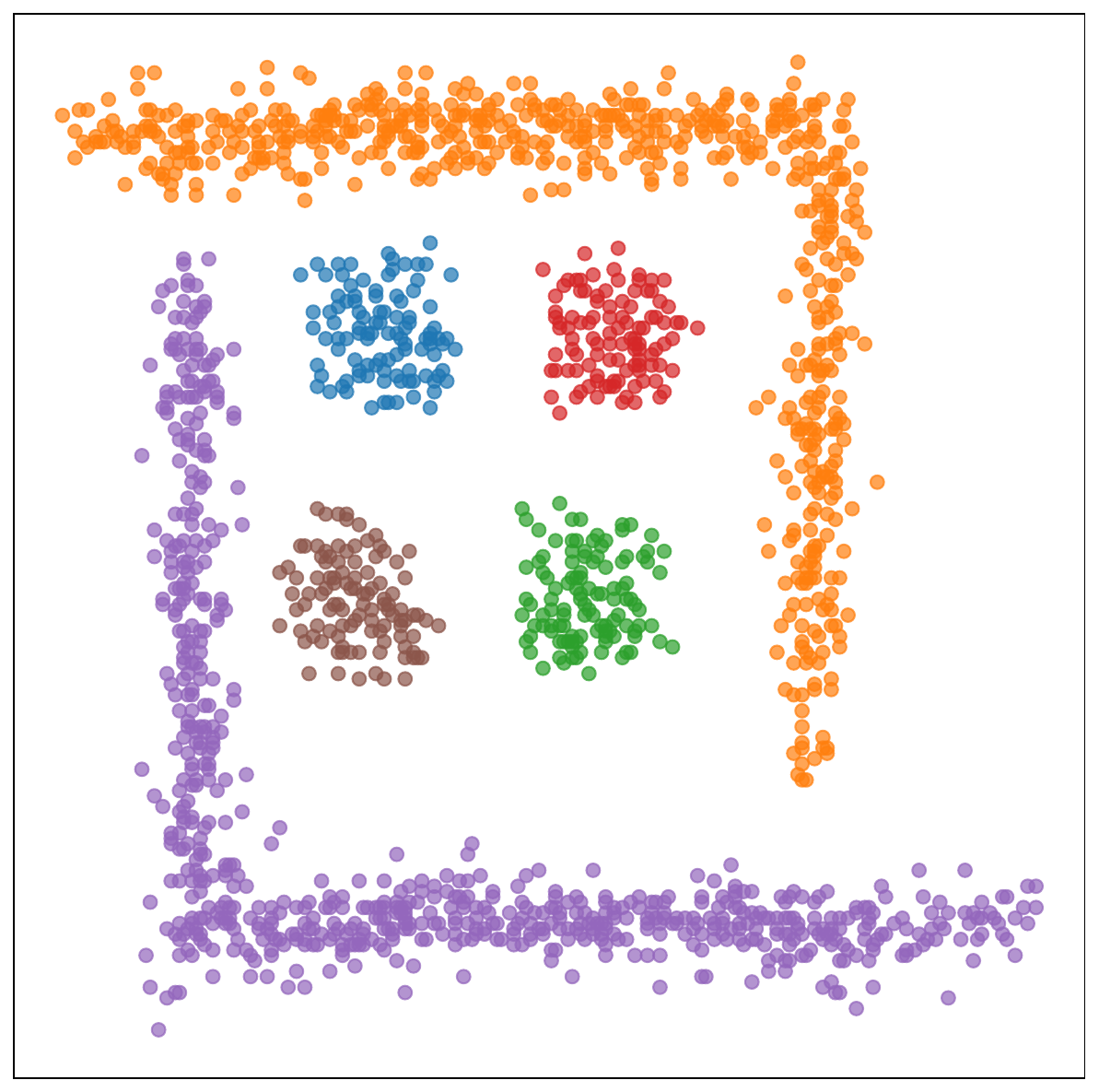}}
	
	\subfigure[complex8]{
		\label{fig:ours_complex8}
		\includegraphics[width=0.15\textwidth]{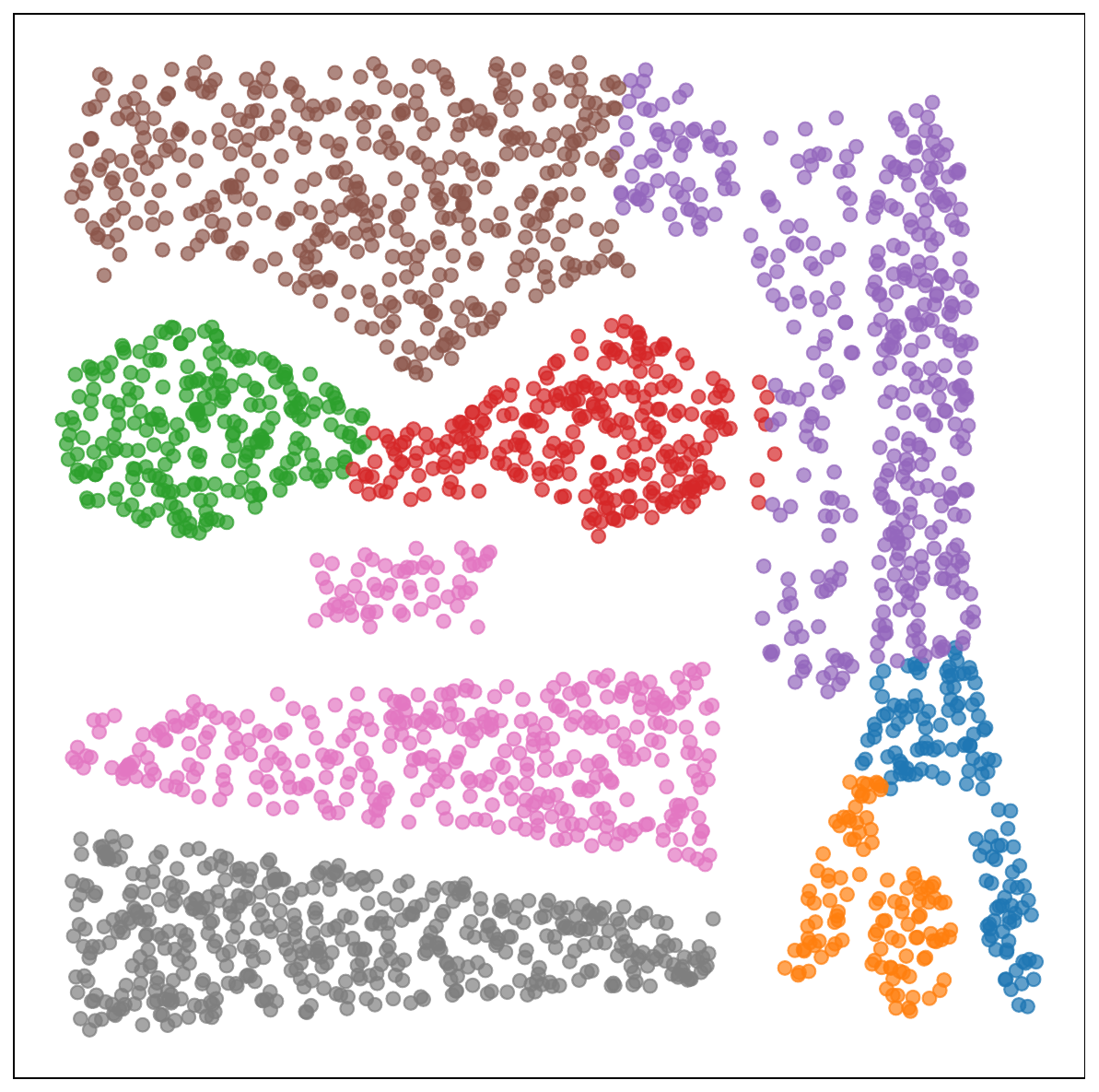}}
	\subfigure[diamond9]{
		\label{fig:ours_diamond9}
		\includegraphics[width=0.15\textwidth]{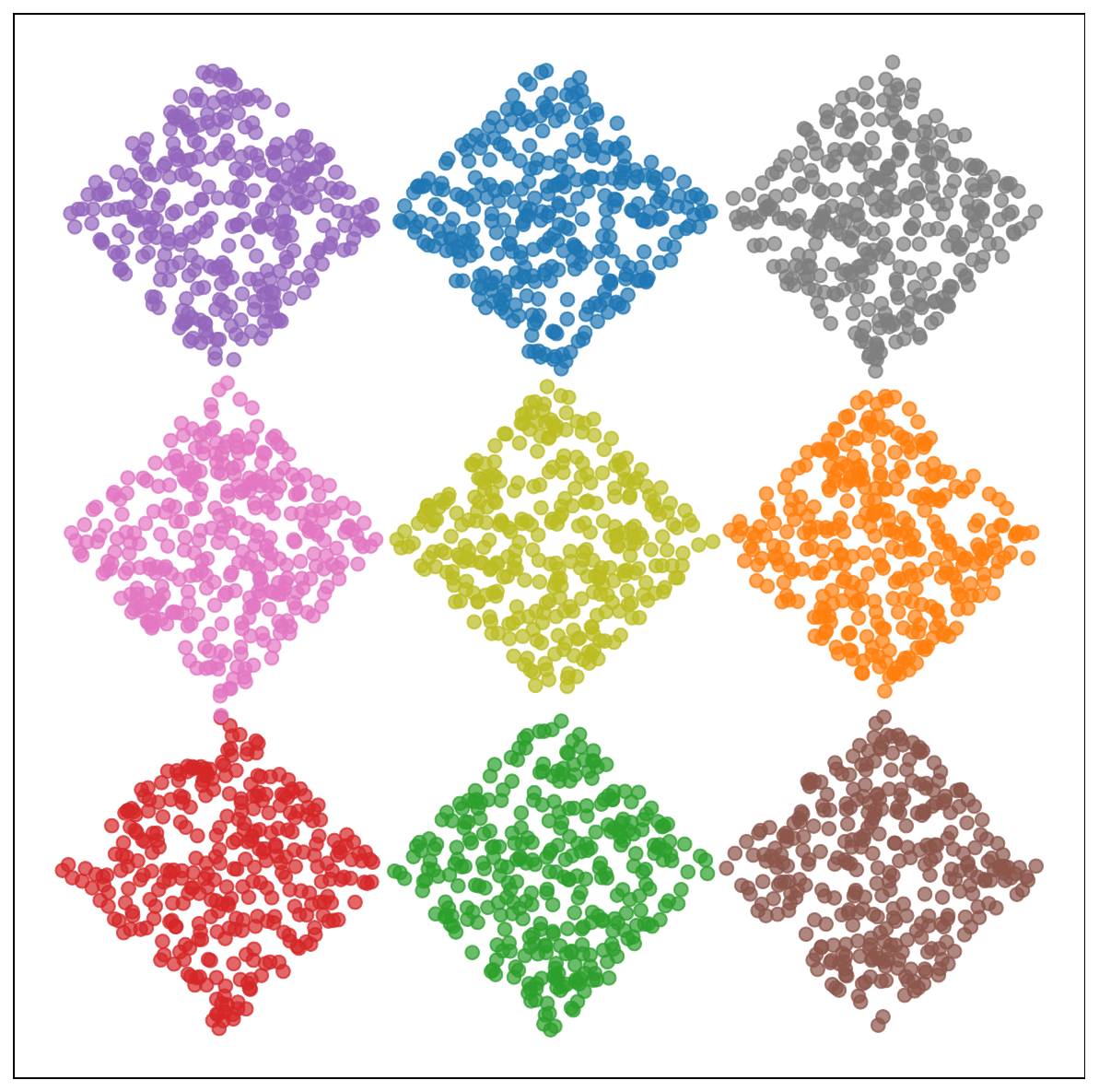}}
	\subfigure[cure-t2-4k]{
		\label{fig:ours_cure}
		\includegraphics[width=0.15\textwidth]{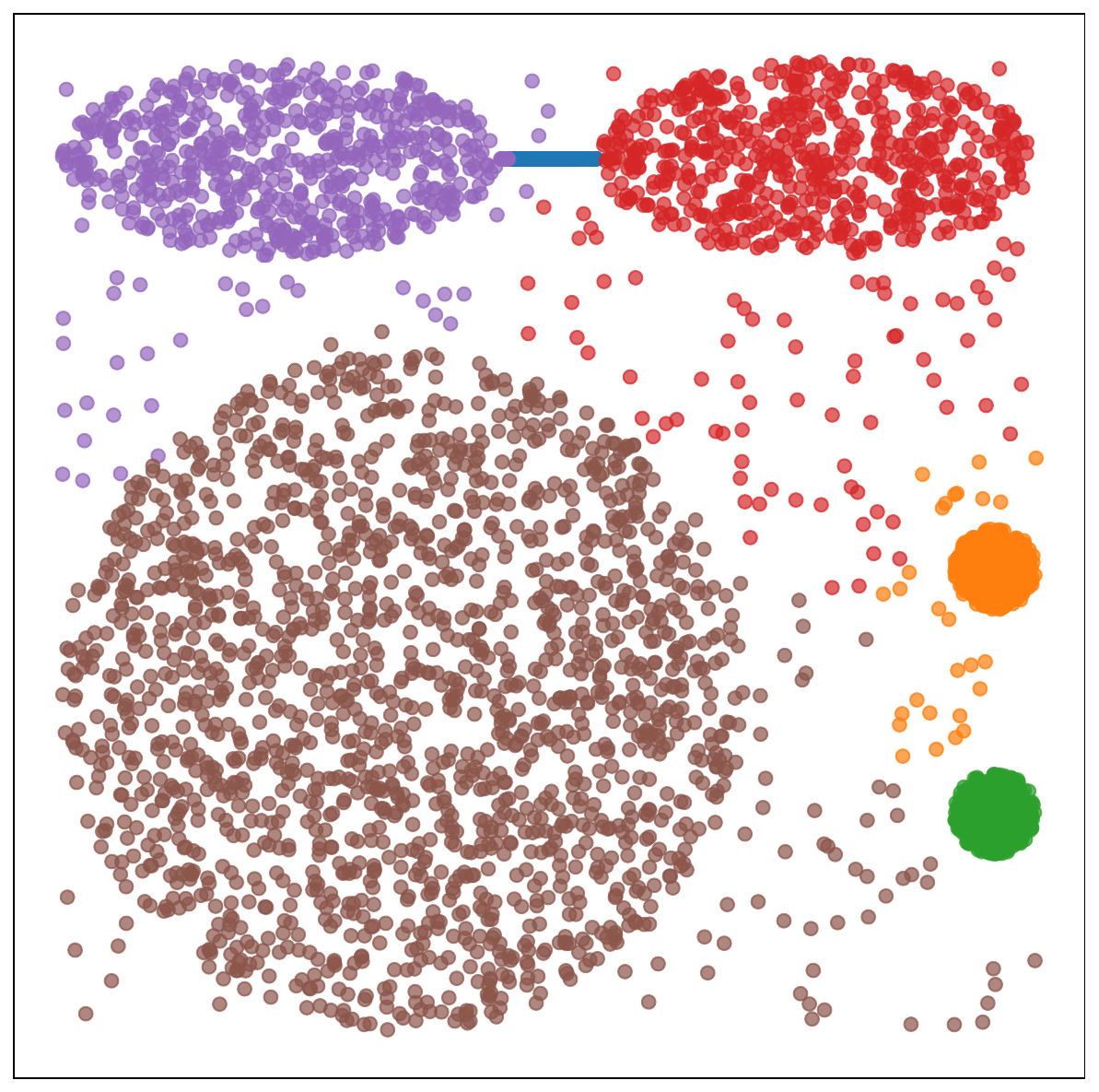}}
	\subfigure[banana]{
		\label{fig:ours_banana}
		\includegraphics[width=0.15\textwidth]{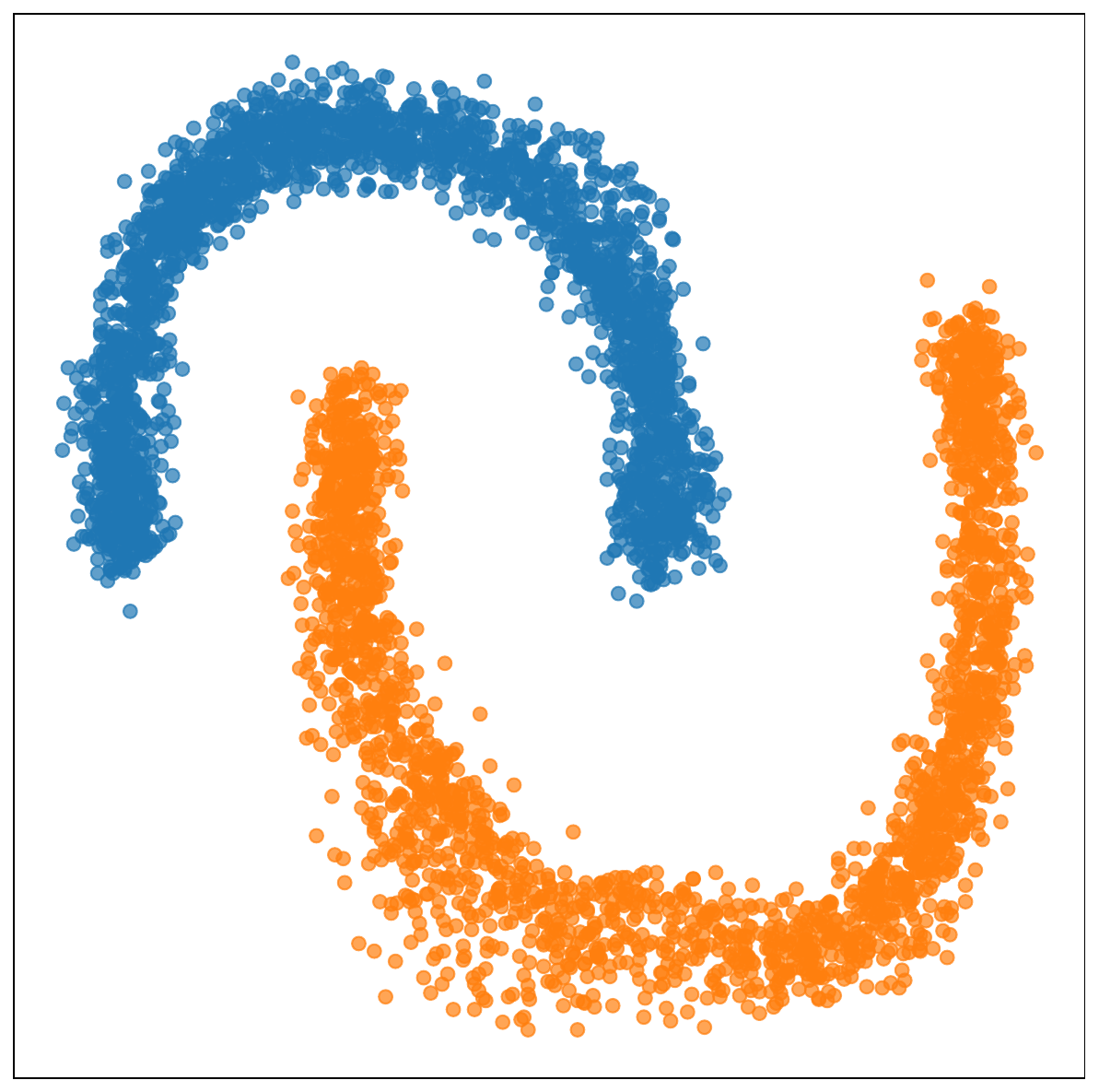}}
	\subfigure[cluto-t8-8k]{
		\label{fig:ours_cluto}
		\includegraphics[width=0.15\textwidth]{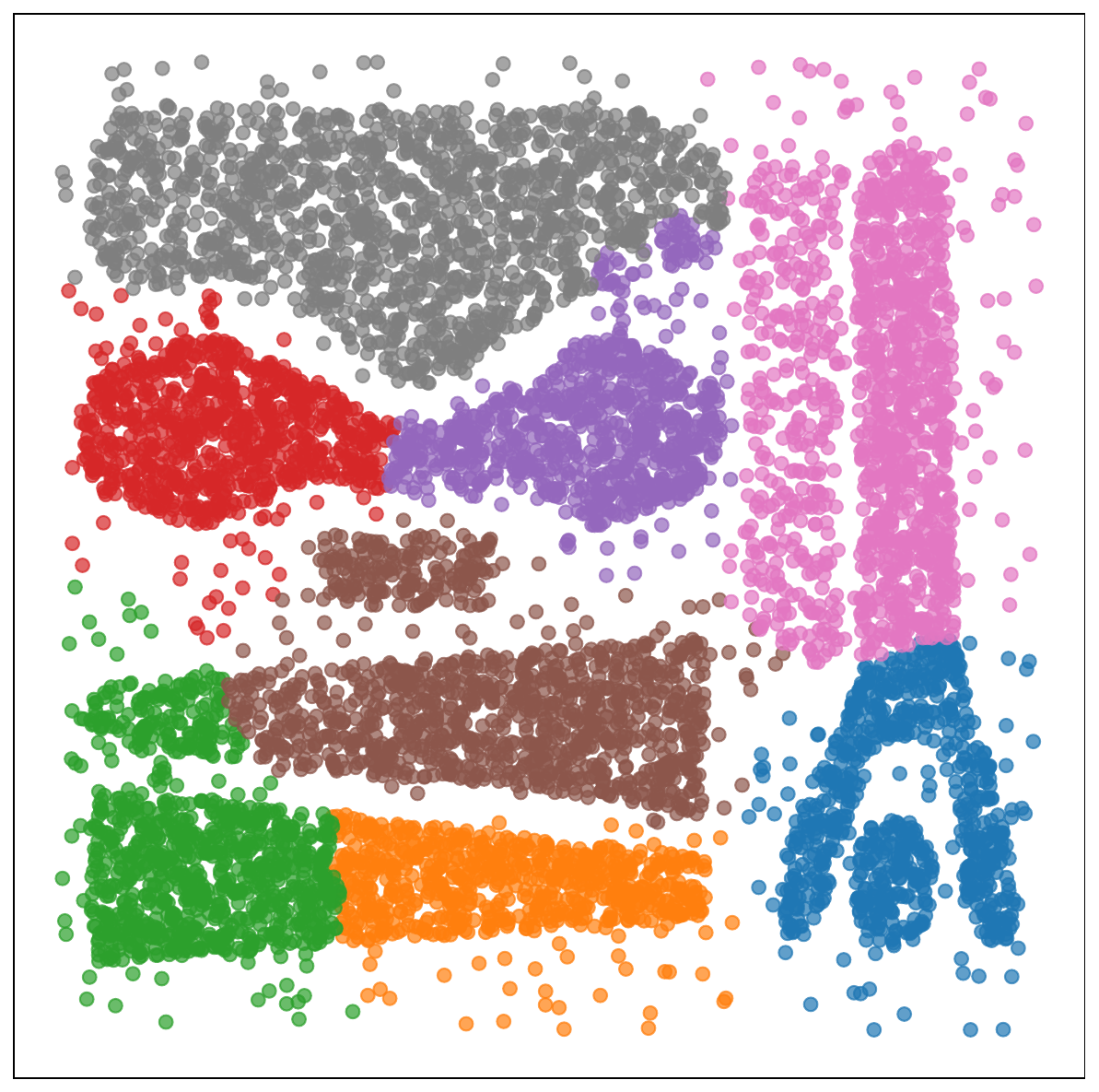}}
	
	\caption{Clustering partitions produced by MDL-GBTRSC on the 20 synthetic benchmark datasets.}
	\label{fig:mdl_gbtrsc_synthetic_all}
\end{figure}

Figure~\ref{fig:mdl_gbtrsc_synthetic_all} provides a geometric view of the MDL-GBTRSC partitions on the 20 synthetic datasets. The recovered partitions preserve a range of nested, intertwined, non-convex, and multi-cluster structures. The 12 datasets with $\mathrm{ARI}=\mathrm{NMI}=1$ are recovered exactly up to a permutation of cluster identifiers. On blobs, wingnut, complex8, and cluto-t8-8k, the remaining assignment differences are visible in the corresponding panels and are consistent with the lower numerical scores in Table~\ref{tab:synthetic_ari_nmi_comparison}. The corresponding partitions of the nine competing methods are provided in Appendix~\ref{app:synthetic_visualizations} under the same visualization protocol; stochastic methods are represented by their seed-0 runs.

\subsubsection{Statistical Comparison}
\label{subsec:statistical_significance}

Statistical comparisons are conducted separately for ARI and NMI on the real and synthetic benchmark suites. The Friedman test is applied first to the ten methods, followed by the Nemenyi post-hoc comparison for the overall rank structure when the omnibus null hypothesis is rejected. Pairwise comparisons between MDL-GBTRSC and each baseline are additionally performed using two-sided Wilcoxon signed-rank tests with Holm correction within each metric and benchmark suite. For $k$ methods evaluated on $N$ complete datasets, the Nemenyi critical difference is
\begin{equation}
		CD=q_{0.05}\sqrt{\frac{k(k+1)}{6N}},
	\label{eq:nemenyi_cd}
\end{equation}
where $k=10$ and $q_{0.05}=3.1637$. This gives $CD=3.2854$ for the 17 complete real datasets and $CD=3.0290$ for the 20 synthetic datasets. The Nemenyi analysis compares all methods through their average ranks, whereas the Wilcoxon-Holm analysis uses paired dataset-wise performance differences for the specific comparison between MDL-GBTRSC and each baseline.

\begin{table}[t]
	\centering
	\caption{Average ranks of the ten methods on the complete-case comparisons. Lower ranks indicate better performance.}
	\label{tab:summary_performance}
	\scriptsize
	\setlength{\tabcolsep}{4pt}
	\renewcommand{\arraystretch}{1.10}
	\begin{tabular}{lrrrr}
		\toprule
		& \multicolumn{2}{c}{Real (17)} & \multicolumn{2}{c}{Synthetic (20)} \\
		\cmidrule(lr){2-3}\cmidrule(lr){4-5}
		Method & ARI & NMI & ARI & NMI \\
		\midrule
		MDL-GBTRSC & \textbf{2.94} & \textbf{2.94} & \textbf{1.95} & \textbf{1.95} \\
		STSC-FK    & 3.71 & 3.59 & 3.48 & 3.45 \\
		N-DSC      & 4.35 & 4.88 & 5.70 & 5.50 \\
		FastSC     & 7.18 & 7.24 & 8.60 & 8.60 \\
		SC-kNN     & 3.53 & 3.41 & 4.88 & 4.85 \\
		SC-RBF     & 6.59 & 6.65 & 7.53 & 7.63 \\
		GBCT       & 9.24 & 9.35 & 4.88 & 4.98 \\
		MDMSC      & 5.12 & 4.12 & 4.25 & 4.20 \\
		GBSC       & 6.41 & 6.76 & 6.70 & 6.55 \\
		GMM-SC     & 5.94 & 6.06 & 7.05 & 7.30 \\
		\bottomrule
	\end{tabular}
\end{table}

\begin{figure}
	\centering
	\begin{tabular}{@{}c@{\hspace{0.02\textwidth}}c@{}}
		\includegraphics[width=0.47\textwidth]{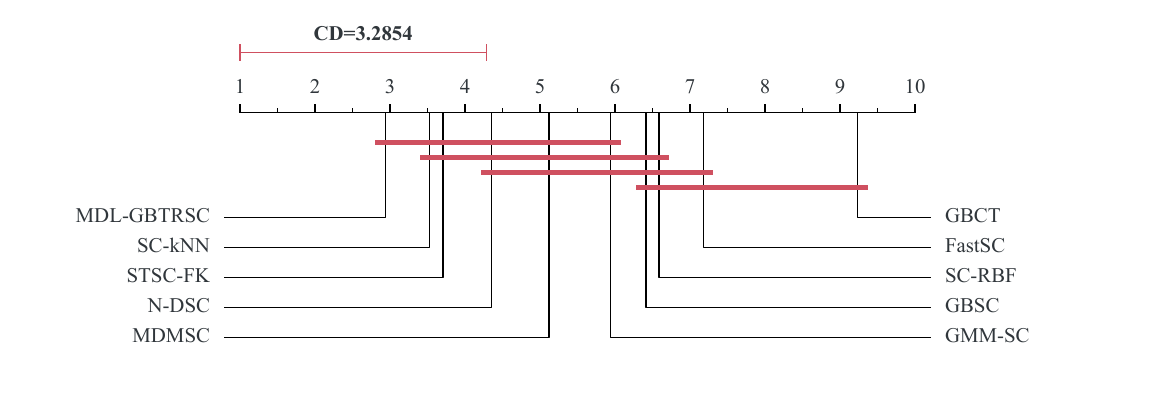} &
		\includegraphics[width=0.47\textwidth]{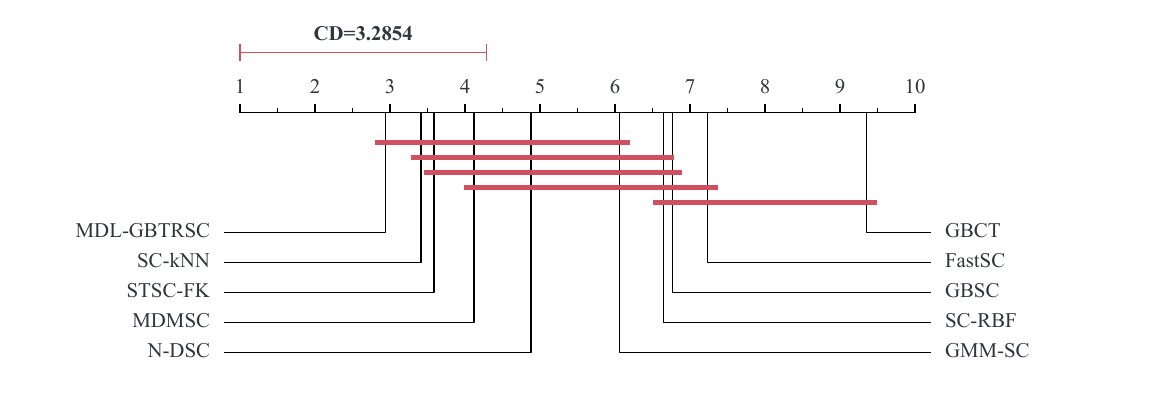} \\[0.6em]
		\includegraphics[width=0.47\textwidth]{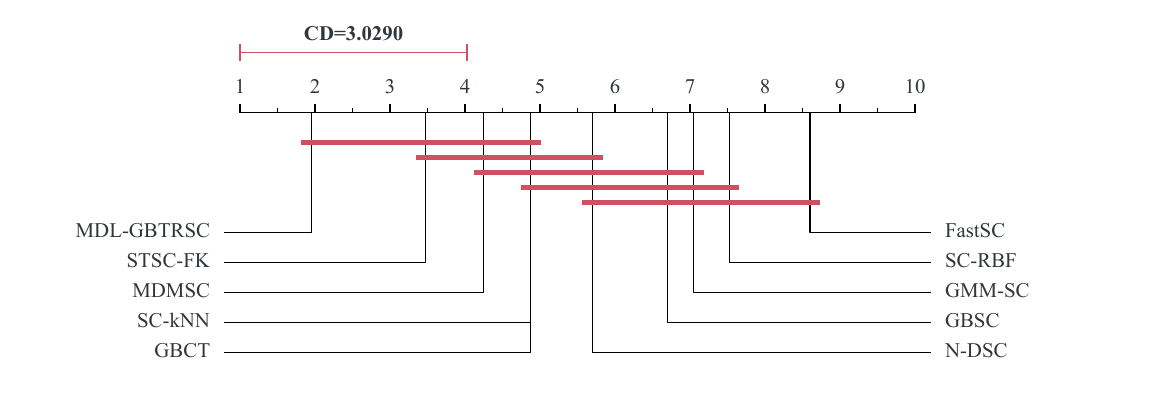} &
		\includegraphics[width=0.47\textwidth]{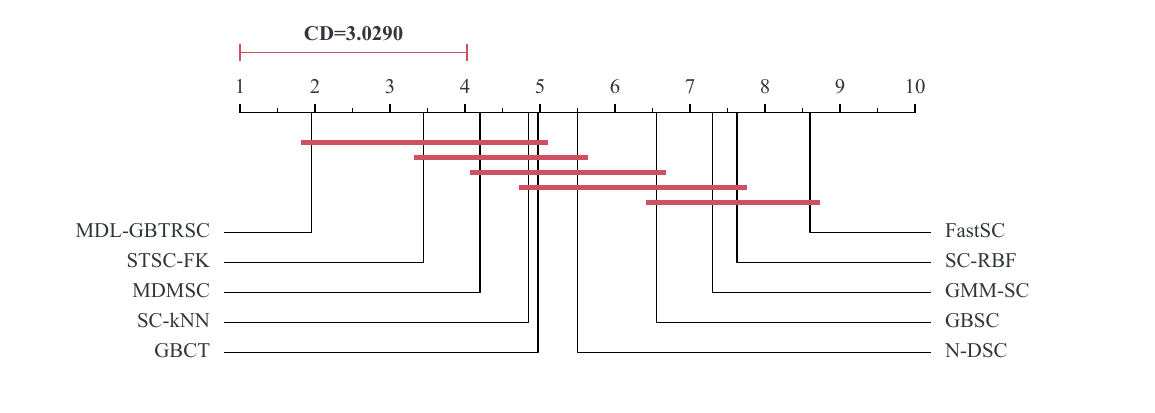}
	\end{tabular}
	\caption{Nemenyi critical-difference diagrams at $\alpha=0.05$ for ARI and NMI on the real and synthetic benchmark suites. Lower average ranks indicate better performance. Methods connected by a horizontal segment are not separated by more than the corresponding critical difference.}
	\label{fig:cd_diagrams}
\end{figure}

On the 17 complete real datasets, the Friedman tests reject equality among the ten methods for both ARI ($p=3.21\times10^{-10}$) and NMI ($p=1.29\times10^{-11}$). MDL-GBTRSC obtains the lowest average rank of $2.94$ for both metrics, followed by SC-kNN with ranks $3.53/3.41$ and STSC-FK with $3.71/3.59$. Relative to MDL-GBTRSC, the Nemenyi rank differences exceed $CD=3.2854$ for FastSC, SC-RBF, GBCT, and GBSC, while STSC-FK, N-DSC, SC-kNN, MDMSC, and GMM-SC remain within the critical distance. The Wilcoxon-Holm comparisons identify significant differences from FastSC, SC-RBF, GBCT, GBSC, and GMM-SC, whereas the comparisons with STSC-FK, N-DSC, SC-kNN, and MDMSC are not significant after correction.

On the 20 synthetic datasets, the Friedman tests give $p=4.02\times10^{-14}$ for ARI and $p=1.29\times10^{-14}$ for NMI. MDL-GBTRSC again has the lowest average rank, $1.95$ for both metrics, followed by STSC-FK at $3.48/3.45$ and MDMSC at $4.25/4.20$. With $CD=3.0290$, the ARI rank differences from MDL-GBTRSC exceed the critical distance for N-DSC, FastSC, SC-RBF, GBSC, and GMM-SC, while the difference to GBCT is just below the threshold; for NMI, GBCT also marginally exceeds the critical distance. The Holm-corrected Wilcoxon tests identify significant differences between MDL-GBTRSC and all nine baselines for both metrics. The difference between the two post-hoc analyses follows from the quantities they compare: Nemenyi operates on all-method average-rank differences, whereas the Wilcoxon tests use the paired performance differences for each specified method pair.

\subsubsection{Runtime Comparison}
\label{subsec:runtime_comparison}

\begin{figure}
	\centering
	\includegraphics[width=\columnwidth]{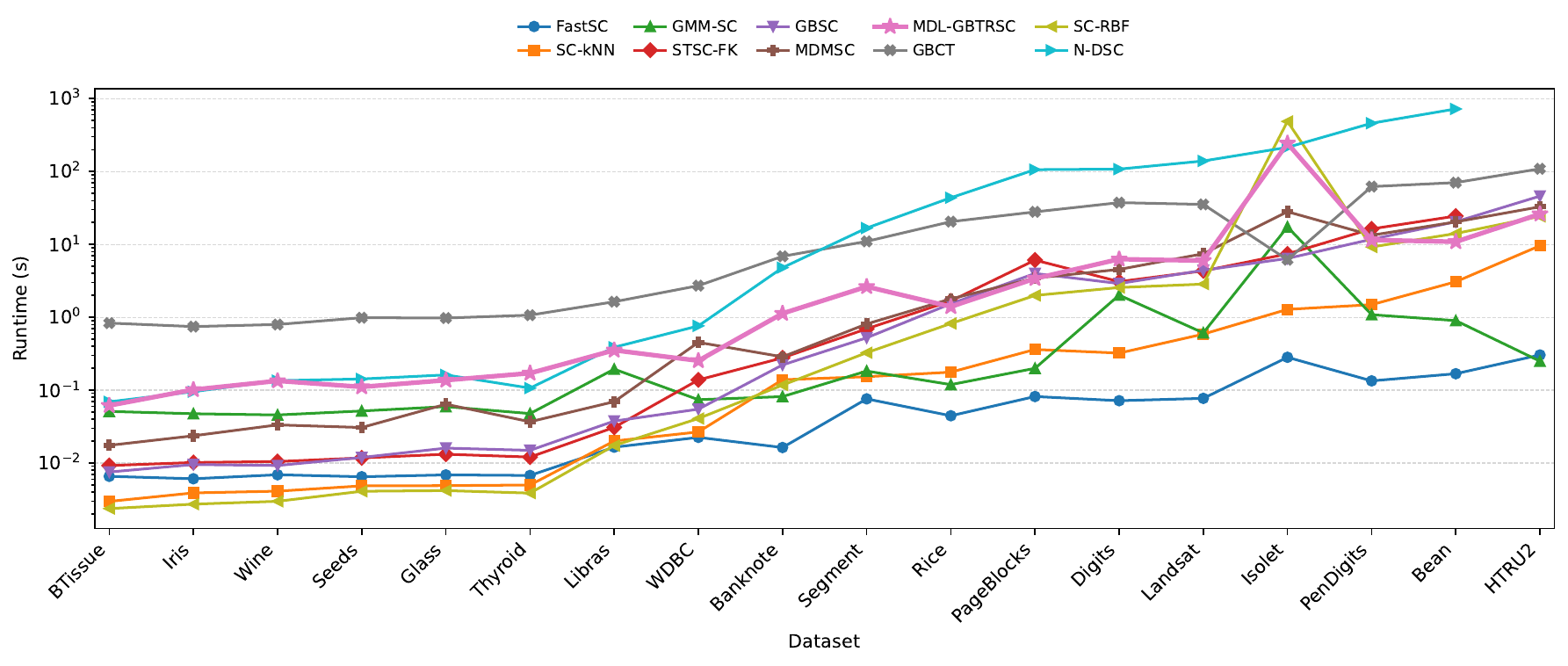}
	\caption{Dataset-level mean runtime on the 18 real benchmark datasets. The vertical axis is logarithmic. Runtime is unavailable for STSC-FK and N-DSC on HTRU2 because these runs exceed the available memory.}
	\label{fig:runtime_comparison_real}
\end{figure}

Runtime is evaluated on the 18 real benchmark datasets using the same implementations, preprocessing, and hardware described above. Each method is run 20 times on every dataset, and the mean runtime is reported. Table~\ref{tab:runtime_comparison} summarizes the method-wise averages, while Fig.~\ref{fig:runtime_comparison_real} shows the corresponding dataset-level results. STSC-FK and N-DSC exceed the available memory on HTRU2, so their reported averages are computed over the remaining 17 datasets.

\begin{table}[t]
	\centering
	\caption{Mean runtime on the real benchmark datasets. Completed denotes the number of datasets with an available runtime measurement.}
	\label{tab:runtime_comparison}
	\scriptsize
	\setlength{\tabcolsep}{7pt}
	\renewcommand{\arraystretch}{1.08}
	\begin{tabular}{lrr}
		\toprule
		Method & Runtime (s) & Completed \\
		\midrule
		FastSC      & 0.079   & 18/18 \\
		SC-kNN      & 1.115   & 18/18 \\
		GMM-SC      & 1.428   & 18/18 \\
		STSC-FK     & 4.509   & 17/18 \\
		GBSC        & 5.890   & 18/18 \\
		MDMSC       & 6.764   & 18/18 \\
		MDL-GBTRSC  & 18.595  & 18/18 \\
		GBCT        & 22.610  & 18/18 \\
		SC-RBF      & 36.356  & 18/18 \\
		N-DSC       & 111.283 & 17/18 \\
		\bottomrule
	\end{tabular}
\end{table}

The aggregate results in Table~\ref{tab:runtime_comparison} show a clear separation in computational scale. FastSC, SC-kNN, and GMM-SC have the smallest mean runtimes, followed by STSC-FK, GBSC, and MDMSC. MDL-GBTRSC requires $18.595$ s on average over the 18 real datasets, compared with $22.610$ s for GBCT and $36.356$ s for SC-RBF. N-DSC has the largest reported average runtime, based on the 17 datasets it completes.

The dataset-level curves in Fig.~\ref{fig:runtime_comparison_real} show that the relative cost is not determined by sample size alone. Runtime differences remain moderate on several low-dimensional datasets but become more pronounced on higher-dimensional problems, with Isolet providing a representative case. MDL-GBTRSC performs neighborhood construction and repeated local model evaluation in the original feature space before forming the final sample graph, so increasing dimension affects both pairwise geometric operations and the local descriptions evaluated during granular-ball refinement.

The generated hierarchy introduces an additional data-dependent factor. Different datasets can produce different numbers and depths of local refinement decisions, which changes the amount of description-length evaluation and the number of terminal regions whose scales are transferred to the sample-level affinity graph. The controlled scalability experiments in Section~\ref{subsec:scalability} separate the effects of sample size and feature dimension under a common data-generation setting.

\subsection{Structure of the Learned Graph and Final Partition}
\label{subsec:graph_behavior}

Across the 38 benchmark datasets, MDL-GBTRSC selects the reciprocal graph in 6 cases and the completed graph in 32 cases. The resulting selected graphs contain exactly $K$ connected components on 8 datasets, satisfy $1<c_s<K$ on another 8 datasets, and are connected on the remaining 22 datasets. These three connectivity states lead respectively to direct component assignment, component-wise spectral subdivision, and global spectral partitioning. Table~\ref{tab:graph_states} reports the corresponding component counts, selected graph, and final partitioning mode for each dataset; none of the selected graphs contains isolated vertices or more than $K$ connected components.

\begin{table*}[t]
	\centering
	\caption{Graph-selection and partitioning states of MDL-GBTRSC on the 38 benchmark datasets. $c_r$, $c_c$, and $c_s$ denote the numbers of connected components in the reciprocal, completed, and selected graphs, respectively. ``Rec.'' and ``Comp.'' denote the reciprocal and completed graphs; ``Direct'', ``Component'', and ``Global'' denote $c_s=K$, $1<c_s<K$ with component-wise spectral subdivision, and global spectral partitioning, respectively.}
	\label{tab:graph_states}
	\scriptsize
	\renewcommand{\arraystretch}{1.05}
	
	\resizebox{\textwidth}{!}{%
		\begin{tabular}{lrrrrll@{\hspace{1.2em}}lrrrrll}
			\toprule
			\multicolumn{7}{c}{\textbf{Real datasets}} &
			\multicolumn{7}{c}{\textbf{Synthetic datasets}} \\
			\cmidrule(lr){1-7}\cmidrule(lr){8-14}
			Dataset & $K$ & $c_r$ & $c_c$ & $c_s$ & Graph & Partition &
			Dataset & $K$ & $c_r$ & $c_c$ & $c_s$ & Graph & Partition \\
			\midrule
			BTissue    & 6  & 3  & 1 & 1 & Comp. & Global    & zelnik6      & 3  & 3  & 1  & 3  & Rec.  & Direct \\
			Iris       & 3  & 5  & 2 & 2 & Comp. & Component & zelnik1      & 3  & 3  & 3  & 3  & Rec.  & Direct \\
			Wine       & 3  & 6  & 1 & 1 & Comp. & Global    & blobs        & 3  & 2  & 1  & 1  & Comp. & Global \\
			Seeds      & 3  & 2  & 1 & 1 & Comp. & Global    & zelnik2      & 3  & 4  & 1  & 1  & Comp. & Global \\
			Glass      & 6  & 15 & 1 & 1 & Comp. & Global    & 3-spiral     & 3  & 3  & 1  & 3  & Rec.  & Direct \\
			Thyroid    & 3  & 7  & 1 & 1 & Comp. & Global    & db2          & 4  & 4  & 1  & 4  & Rec.  & Direct \\
			Libras     & 15 & 1  & 1 & 1 & Comp. & Global    & jain         & 2  & 2  & 1  & 2  & Rec.  & Direct \\
			WDBC       & 2  & 18 & 1 & 1 & Comp. & Global    & zelnik4      & 4  & 7  & 1  & 1  & Comp. & Global \\
			Banknote   & 2  & 10 & 1 & 1 & Comp. & Global    & target       & 6  & 6  & 2  & 6  & Rec.  & Direct \\
			Segment    & 7  & 30 & 2 & 2 & Comp. & Component & aggregation  & 7  & 5  & 5  & 5  & Comp. & Component \\
			Rice       & 2  & 7  & 1 & 1 & Comp. & Global    & chainlink    & 2  & 4  & 2  & 2  & Comp. & Direct \\
			PageBlocks & 5  & 27 & 2 & 2 & Comp. & Component & fourty       & 40 & 45 & 37 & 37 & Comp. & Component \\
			Digits     & 10 & 38 & 1 & 1 & Comp. & Global    & wingnut      & 2  & 1  & 1  & 1  & Comp. & Global \\
			Landsat    & 6  & 55 & 1 & 1 & Comp. & Global    & N            & 4  & 3  & 3  & 3  & Comp. & Component \\
			Isolet     & 26 & 34 & 1 & 1 & Comp. & Global    & A            & 6  & 9  & 5  & 5  & Comp. & Component \\
			PenDigits  & 10 & 60 & 1 & 1 & Comp. & Global    & complex8     & 8  & 4  & 3  & 3  & Comp. & Component \\
			Bean       & 7  & 22 & 1 & 1 & Comp. & Global    & diamond9     & 9  & 1  & 1  & 1  & Comp. & Global \\
			HTRU2      & 2  & 23 & 1 & 1 & Comp. & Global    & cure-t2-4k   & 6  & 10 & 1  & 1  & Comp. & Global \\
			{}         & {} & {} & {} & {} & {}   & {}        & banana       & 2  & 10 & 2  & 2  & Comp. & Direct \\
			{}         & {} & {} & {} & {} & {}   & {}        & cluto-t8-8k  & 8  & 11 & 1  & 1  & Comp. & Global \\
			\bottomrule
	\end{tabular}}
\end{table*}

For all 18 real datasets, the completed graph is selected. Fifteen of these graphs are connected and proceed to global spectral partitioning, whereas Iris, Segment, and PageBlocks retain two connected components and are partitioned through component-wise subdivision. Consequently, the real-data results are obtained from spectral partitioning of the leaf-regularized sample graph rather than direct assignment from $K$ disconnected components. The different performance levels observed across these datasets therefore reflect differences in the learned affinity weights and the associated spectral embeddings in addition to the coarse connected-component structure.

Among the 20 synthetic datasets, direct component assignment is used on zelnik6, zelnik1, 3-spiral, db2, jain, target, chainlink, and banana, and all eight obtain $\mathrm{ARI}=\mathrm{NMI}=1$. Aggregation, fourty, N, A, and complex8 satisfy $1<c_s<K$ and are processed through component-wise subdivision, while the other seven datasets use global spectral partitioning. Exact recovery is also obtained under the latter two regimes: fourty, N, and A are recovered exactly after component-wise subdivision, and zelnik4 is recovered exactly through global spectral partitioning. The high synthetic accuracy is therefore associated with several graph-connectivity configurations, with the final partitioning route adapting to the structure of the selected sample graph.

\subsection{Mechanism Validation}
\label{subsec:mechanism_validation}

\subsubsection{Screened versus exhaustive split evaluation}
\label{subsec:split_search_analysis}

At each granular-ball node, the screened search evaluates the complete split description only for candidate cuts retained by the preliminary isotropic criterion, whereas exhaustive search evaluates all admissible cuts from the same deterministic ordering family. Exhaustive evaluation remains computationally feasible on Iris, Wine, Seeds, zelnik6, 3-spiral, and aggregation. Across these six datasets, screening reduces the number of complete split evaluations by $84.6\%$ on average and decreases the corresponding runtime by a factor of $3.50$. The reduction ranges from $78.6\%$ on Iris to $92.2\%$ on aggregation, with runtime speedups ranging from $2.59\times$ to $4.51\times$. In aggregation, for example, the number of complete split evaluations decreases from $15{,}660$ to $1{,}228$, while runtime decreases from $1.44$ s to $0.32$ s.

The reduction in candidate evaluation is accompanied by close agreement between the resulting partitions. The mean mutual ARI between the screened and exhaustive results is $0.9747$, and the mean absolute differences in external ARI and NMI are $0.0129$ and $0.0125$, respectively. On 3-spiral and zelnik6, the two procedures produce identical final partitions, while aggregation retains a mutual ARI of $0.9941$ despite the $92.2\%$ reduction in complete evaluations. The largest difference appears on Seeds, where the mutual ARI is $0.9120$, indicating that the preliminary ranking can modify the refinement path when a candidate excluded during screening would receive a more favorable score under the complete description-length criterion. Overall, the screened evaluation preserves the final clustering structure closely while substantially reducing the number of expensive complete split evaluations.

\begin{figure}
	\centering
	\includegraphics[width=0.82\columnwidth]{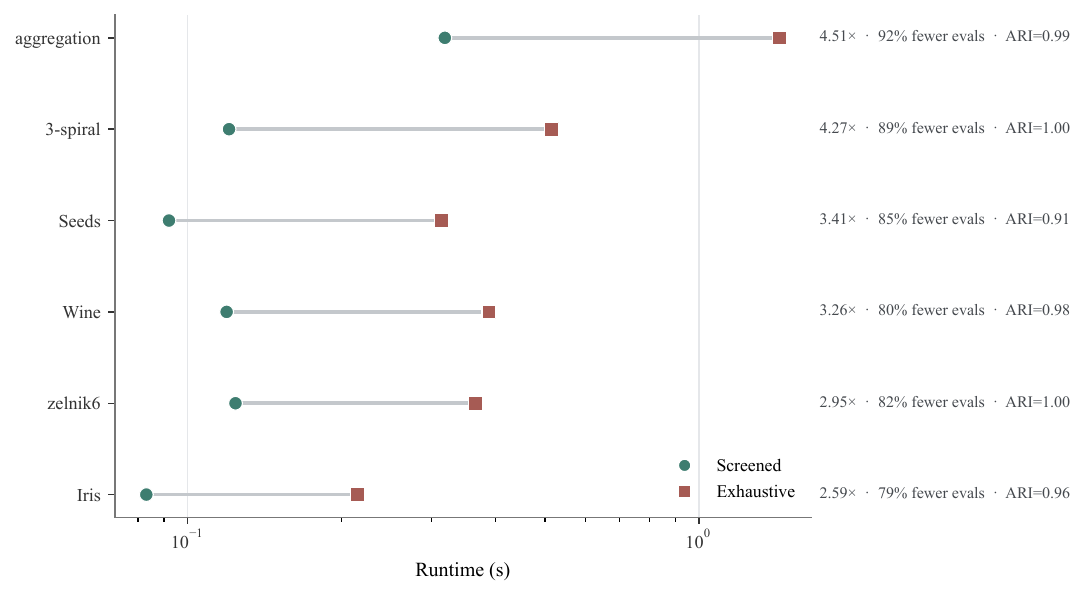}
	\caption{Runtime comparison between screened and exhaustive split evaluation on the six datasets for which exhaustive search is feasible. Annotations report the runtime speedup, reduction in complete split evaluations, and mutual ARI between the resulting partitions.}
	\label{fig:split_search_analysis}
\end{figure}

\subsubsection{Ablation Study}
\label{subsec:ablation}

\begin{figure}
	\centering
	\includegraphics[width=0.90\columnwidth]{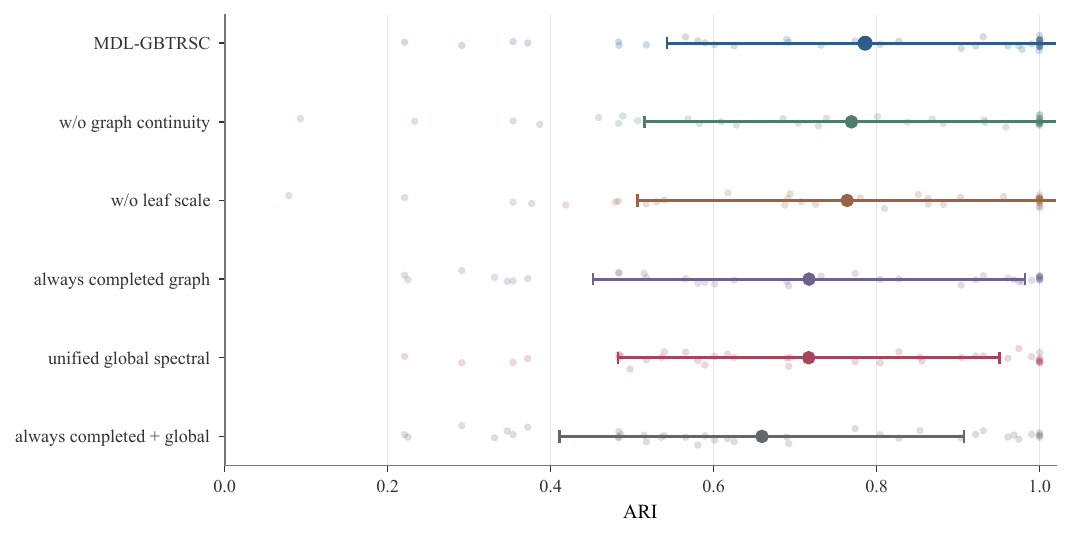}
	\caption{Dataset-level ARI values of MDL-GBTRSC and its five ablation variants. Translucent points represent individual datasets, while the filled circles and horizontal error bars denote the mean and one standard deviation, respectively.}
	\label{fig:ablation}
\end{figure}

Five ablation variants are constructed to examine how the main components of MDL-GBTRSC affect different stages of the clustering process. The graph-continuity term is first removed from granular-ball refinement, and the leaf-scale term is separately removed from sample-level affinity construction. The two downstream mechanisms are then examined by always using the completed graph or by replacing the connectivity-aware partitioning strategy with a global spectral partition. A final variant applies both downstream modifications simultaneously. Table~\ref{tab:ablation_summary} reports the corresponding changes in clustering performance and hierarchy statistics, while Fig.~\ref{fig:ablation} shows the dataset-level ARI distributions. The individual ARI changes for all 38 datasets are reported in Appendix~\ref{app:ablation_details}.

\begin{table*}[t]
	\centering
	\caption{Ablation results of the main components of MDL-GBTRSC. $\Delta$ARI and $\Delta$NMI denote changes relative to the full formulation. Leaves, Depth, and Samples/leaf are average statistics of the learned granular-ball hierarchy.}
	\label{tab:ablation_summary}
	\scriptsize
	\setlength{\tabcolsep}{7pt}
	\renewcommand{\arraystretch}{1.08}
	\begin{tabular}{lccccc}
		\toprule
		Variant & $\Delta$ARI & $\Delta$NMI & Leaves & Depth & Samples/leaf \\
		\midrule
		Full MDL-GBTRSC
		& -- & -- & 97.4 & 8.74 & 141.4 \\
		
		Without graph continuity
		& -0.0167 & -0.0134 & 639.7 & 11.53 & 65.3 \\
		
		Without leaf scale
		& -0.0220 & -0.0208 & 97.4 & 8.74 & 141.4 \\
		
		Completed graph only
		& -0.0688 & -0.0546 & 97.4 & 8.74 & 141.4 \\
		
		Global partition only
		& -0.0693 & -0.0386 & 97.4 & 8.74 & 141.4 \\
		
		Completed graph + global partition
		& -0.1266 & -0.0842 & 97.4 & 8.74 & 141.4 \\
		\bottomrule
	\end{tabular}
\end{table*}

Removing graph continuity produces a relatively moderate decrease in the final clustering scores, with mean ARI and NMI changing by $-0.0167$ and $-0.0134$, respectively, while the learned hierarchy changes substantially. The average number of terminal balls increases from $97.4$ to $639.7$, the average tree depth increases from $8.74$ to $11.53$, and the average number of samples per leaf decreases from $141.4$ to $65.3$. After the resulting leaf structures are transferred to the sample graph, an average of 20,464 affinity values differ from those of the full formulation. This behavior is consistent with the role of $L_{\rm cut}$ in the split description, where reciprocal-neighborhood continuity constrains refinements that would otherwise be determined solely by the local geometric description.

The leaf-scale ablation isolates the subsequent transfer from regional structure to sample-level affinities because the learned hierarchy remains unchanged. Removing the scale term from Eq.~\eqref{eq:tree_regularized_cost} modifies an average of 7,627 affinity values and changes mean ARI and NMI by $-0.0220$ and $-0.0208$, respectively. As shown by the dataset-level distributions in Fig.~\ref{fig:ablation}, the magnitude of this effect varies across datasets, reflecting the direct influence of the terminal-ball scale $\eta_b$ on pairwise affinities while leaving tree induction unchanged.

The two downstream ablations lead to larger average changes. Always using the completed graph results in $\Delta\mathrm{ARI}=-0.0688$ and $\Delta\mathrm{NMI}=-0.0546$, whereas replacing the connectivity-aware strategy with a global spectral partition gives $\Delta\mathrm{ARI}=-0.0693$ and $\Delta\mathrm{NMI}=-0.0386$. Their effects occur under different graph states: the former mainly changes datasets for which the reciprocal graph is selected by the full method, while the latter affects datasets whose selected graph contains multiple connected components and therefore invokes component-wise subdivision. Applying both modifications simultaneously produces the largest average change, with $\Delta\mathrm{ARI}=-0.1266$ and $\Delta\mathrm{NMI}=-0.0842$. The distribution in Fig.~\ref{fig:ablation} further shows that these aggregate changes are accompanied by substantial dataset-level variation, consistent with the graph-selection and partitioning states reported in Table~\ref{tab:graph_states}.

\subsection{Effect of the Prescribed Cluster Number}
\label{subsec:k_analysis}

The granular-ball hierarchy and leaf-scale affinities are learned independently of the prescribed cluster number $K$, whereas $K$ enters the graph-selection and final partitioning stage. To isolate this dependence, the learned hierarchy and affinity weights are fixed and only the final stage is recomputed using $K-1$, $K$, and $K+1$. Across the 38 benchmark datasets, the mean ARI/NMI values are $0.4157/0.5038$, $0.7873/0.8146$, and $0.6056/0.7016$, respectively. Relative to the reference $K$, decreasing the prescribed cluster number by one changes ARI/NMI by $-0.3716/-0.3108$, whereas increasing it by one changes them by $-0.1817/-0.1130$.

The stronger reduction under $K-1$ is partly associated with the eight binary datasets in the benchmark suite, for which $K-1=1$ produces a single-cluster partition. Nevertheless, both directions lead to lower average performance than the reference setting. Since the hierarchy and sample-level affinities remain unchanged throughout this experiment, the observed variation arises from the $K$-dependent graph-selection and spectral partitioning stage.

\begin{figure}
	\centering
	\includegraphics[width=0.72\columnwidth]{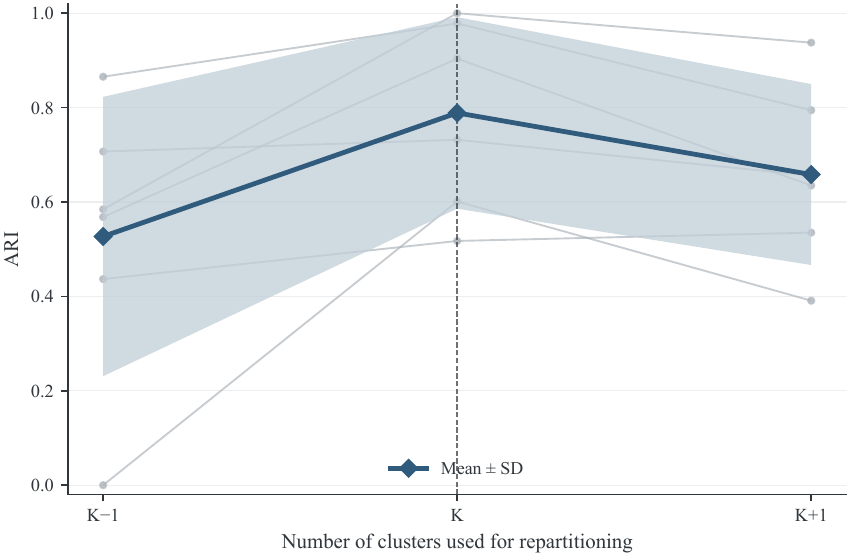}
	\caption{ARI obtained with prescribed cluster numbers $K-1$, $K$, and $K+1$ over the 38 benchmark datasets. The learned granular-ball hierarchy and sample-level affinity weights are fixed, and only the final graph-selection and partitioning stage is recomputed.}
	\label{fig:k_sensitivity}
\end{figure}

\subsection{Scalability Analysis}
\label{subsec:scalability}

The scalability experiments use controlled Gaussian-cluster data generated separately from the benchmark datasets in the preceding experiments. For each configuration $(n,d)$, $n$ samples are generated from $K=5$ isotropic Gaussian clusters with a fixed within-cluster standard deviation. The five cluster centers are constructed deterministically from equally spaced angular positions, and their cosine and sine coordinates are alternated across the feature dimensions. As $d$ increases, these coordinates are repeated instead of appending noise-only dimensions. The resulting between-center distances and the norm of isotropic within-cluster variation both increase approximately with $\sqrt{d}$, so the experiments increase dimensional workload without progressively replacing the original cluster structure by uninformative features. The random state and all data-generation parameters are fixed across configurations.

The prescribed cluster number is fixed at $K=5$ throughout the scalability experiments, while $n$ and $d$ are varied independently or jointly. For sample-size scaling, $d=10$ is fixed and $n$ increases from $1{,}000$ to $200{,}000$. For dimensional scaling, $n=2{,}000$ is fixed and $d$ increases from $2$ to $20{,}000$. In the joint setting, both quantities increase, with the input matrix size $nd$ ranging from $10^{4}$ to $10^{8}$ elements and the largest configuration containing $100{,}000$ samples and $1{,}000$ features. Runtime and the increase in peak resident-set size (RSS) are recorded for each configuration.

\begin{figure}
	\centering
	\begin{tabular}{@{}cc@{}}
		\includegraphics[width=0.47\textwidth]{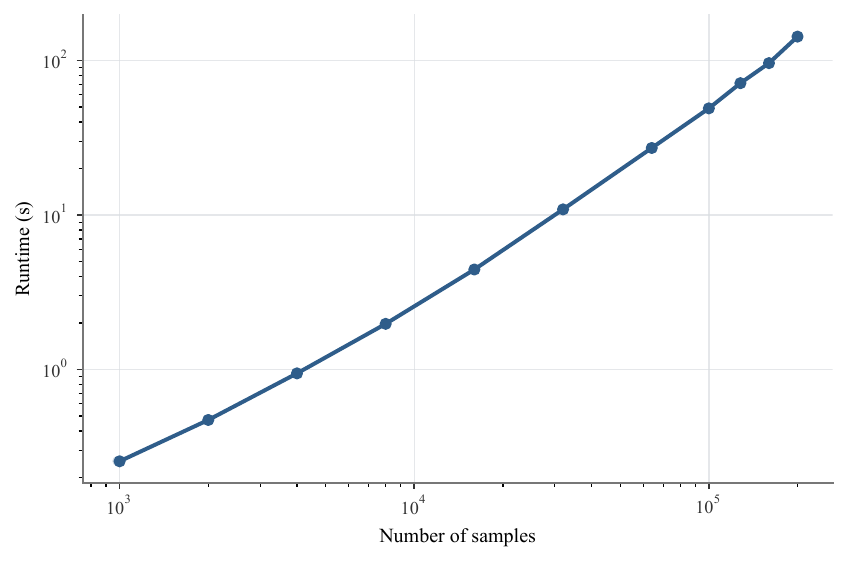} &
		\includegraphics[width=0.47\textwidth]{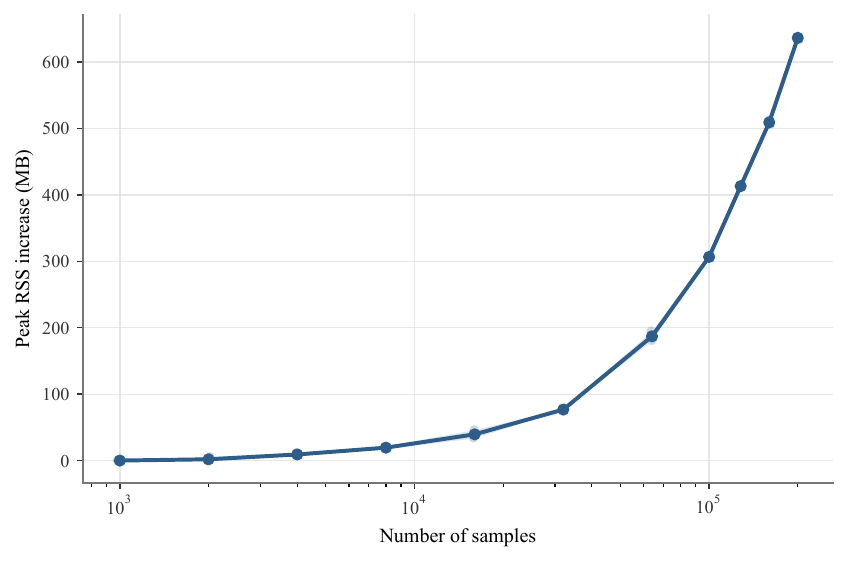} \\[-0.2em]
		{\scriptsize (a) Sample-size scaling: runtime} &
		{\scriptsize (b) Sample-size scaling: memory} \\[0.7em]
		
		\includegraphics[width=0.47\textwidth]{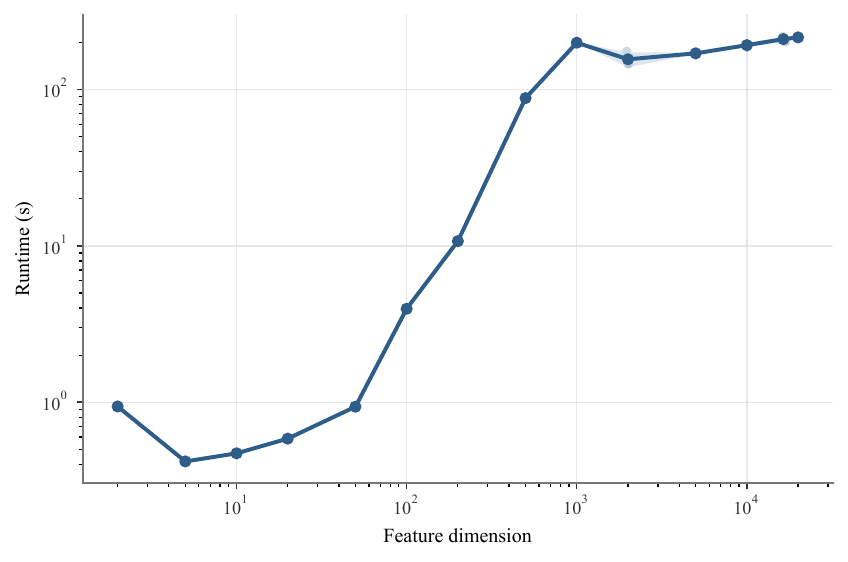} &
		\includegraphics[width=0.47\textwidth]{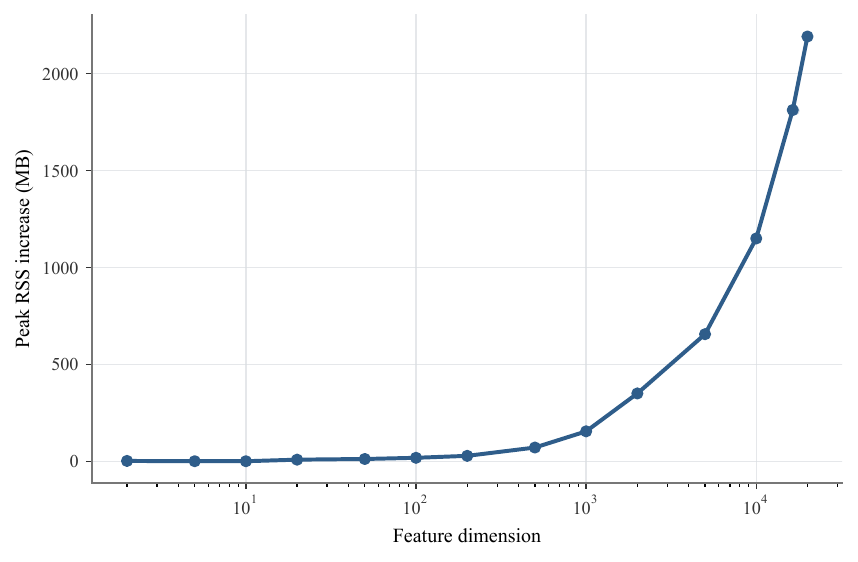} \\[-0.2em]
		{\scriptsize (c) Dimensional scaling: runtime} &
		{\scriptsize (d) Dimensional scaling: memory} \\[0.7em]
		
		\includegraphics[width=0.47\textwidth]{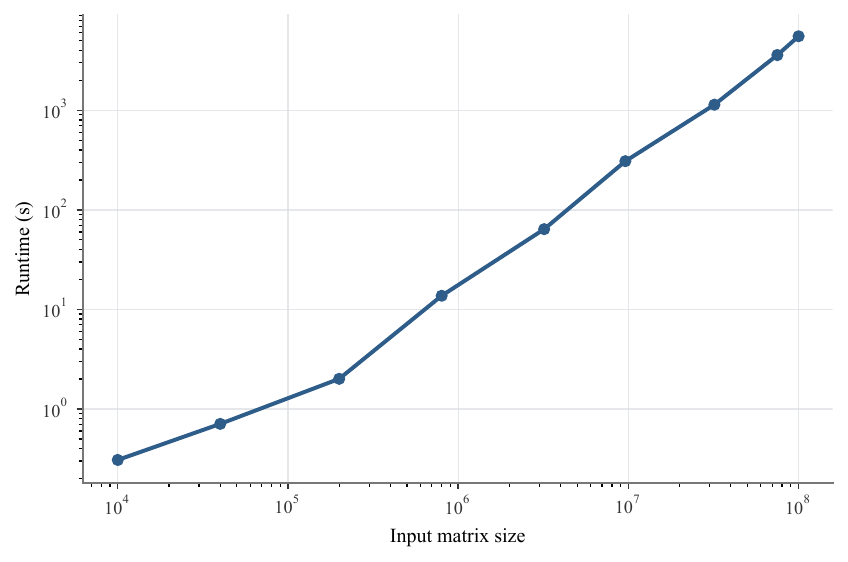} &
		\includegraphics[width=0.47\textwidth]{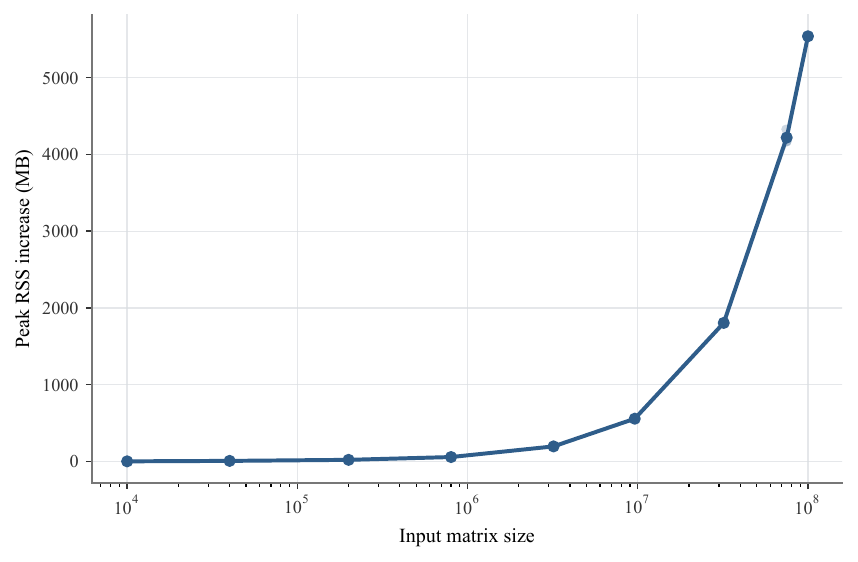} \\[-0.2em]
		{\scriptsize (e) Joint scaling: runtime} &
		{\scriptsize (f) Joint scaling: memory}
	\end{tabular}
	\caption{Runtime and peak-RSS increase of MDL-GBTRSC on the controlled five-cluster Gaussian data. Sample-size scaling fixes $d=10$, dimensional scaling fixes $n=2{,}000$, and $K=5$ is used in all configurations.}
	\label{fig:scalability}
\end{figure}

With $d=10$, runtime increases from approximately $0.25$ s at $n=1{,}000$ to $143.02$ s at $n=200{,}000$. The corresponding increase in peak RSS reaches approximately $636.3$ MB at the largest sample size. Intermediate configurations show the same overall growth trend, with runtimes of $10.88$ s at $n=32{,}000$ and $49.07$ s at $n=100{,}000$. The memory increase remains below $200$ MB up to $n=64{,}000$ and rises to approximately $307$ MB, $509$ MB, and $636$ MB at $n=100{,}000$, $160{,}000$, and $200{,}000$, respectively.

With $n=2{,}000$, increasing $d$ has a more pronounced effect on memory usage. The increase in peak RSS is approximately $350.1$ MB at $d=2{,}000$, $1{,}150.0$ MB at $d=10{,}000$, and $2{,}192.0$ MB at $d=20{,}000$. Runtime rises rapidly up to $d=1{,}000$, reaching approximately $199.2$ s, and then varies between roughly $150$ and $215$ s over the higher-dimensional configurations. The nonmonotonic runtime curve indicates that computational cost is affected not only by the input dimension but also by the hierarchy induced for a particular dataset, whereas memory usage follows a more regular increase with $d$.

When $n$ and $d$ increase together, the larger input matrices lead to substantially higher computational requirements. Runtime increases from approximately $0.31$ s at the smallest configuration to $1{,}138.48$ s when the input contains about $3.2\times10^{7}$ elements, and to $5{,}541.88$ s at $n=100{,}000$ and $d=1{,}000$. The corresponding peak-RSS increase at the largest configuration is approximately $5.54$ GB. At $n=100{,}000$ and $d=750$, runtime is $3{,}589.82$ s and the increase in peak RSS is approximately $4.22$ GB. These results show how the sample-size and dimensional costs accumulate when both quantities grow simultaneously.

The generated cluster structure remains recoverable over the tested scaling range, with ARI and NMI equal to $1$ for all configurations. The reported curves therefore describe the computational behavior of MDL-GBTRSC under increasing input size on this controlled Gaussian setting.

\section{Discussion and Implications}
\label{sec:discussion}

\subsection{Discussion of Results}
\label{subsec:discussion_results}

Across the real and synthetic benchmarks, MDL-GBTRSC achieves the lowest complete-case average ranks for both ARI and NMI, although the degree of statistical separation differs between the two suites. On the real datasets, several strong spectral clustering methods remain statistically comparable after Holm correction. The synthetic benchmarks show a clearer distinction, with MDL-GBTRSC attaining the best or tied performance on 15 of the 20 datasets for both metrics and showing significant paired differences from all competing methods. The learned graph states also differ between the two suites. All real datasets are resolved through global or component-wise spectral partitioning, whereas the synthetic datasets cover direct component assignment, component-wise subdivision, and global spectral partitioning. Exact recovery occurs under each of these regimes, suggesting that accurate partitions can arise from different connectivity structures of the learned sample graph.

Removing reciprocal-neighborhood continuity produces a substantially finer granular-ball hierarchy, whereas removing the leaf-scale term preserves the hierarchy and changes the sample-level affinities. The downstream interventions show a different pattern: graph selection becomes influential when the reciprocal graph is retained by the full model, while connectivity-aware partitioning matters when the selected graph contains multiple connected components. These effects correspond closely to the stages at which the respective components enter the algorithm and help distinguish the roles of hierarchy refinement, affinity construction, and final graph partitioning.

The screened split experiment provides a complementary view of the refinement process. A large proportion of complete candidate evaluations can be omitted while the resulting partitions remain close to those obtained by exhaustive evaluation. The small discrepancies observed on some datasets reflect differences in the refinement paths selected after screening. Together with the scalability experiments, this result shows that part of the computational burden can be reduced during hierarchy construction without altering the subsequent sample-level graph formulation.

Varying the prescribed cluster number affects only the final graph-selection and partitioning stage, since the learned hierarchy and affinity weights remain unchanged. The resulting performance changes under $K-1$ and $K+1$ indicate that the final partition remains sensitive to model order even when the underlying structural representation is fixed. The scalability experiments further show different responses to increasing sample size and dimensionality: memory usage grows relatively regularly with dimension, whereas runtime is also influenced by changes in the learned hierarchy. When large $n$ and large $d$ occur simultaneously, neighborhood construction, local modeling, and spectral computation jointly account for a substantial computational load.

\subsection{Theoretical Implications}
\label{subsec:theoretical_implications}

Within MDL-GBTRSC, granular balls describe regional structure without replacing the original observations as vertices of the final spectral graph. The hierarchy captures how samples are organized locally, and the information learned at this regional level is subsequently transferred to pairwise sample relations through the affinity function. Regional and sample-level representations are therefore linked within the same clustering process: the former provides structural context, while the latter retains the resolution required for spectral partitioning.

Reciprocal continuity, regional scale, and graph connectivity characterize different aspects of this structure. Reciprocal continuity influences whether a local region is retained or further refined, the terminal-ball scale describes the extent of the resulting region, and graph connectivity reflects the organization of the regularized sample-level relations. Their distinct behavior in the ablation experiments is consistent with these different roles. In particular, the connected-component structure contributes to the choice of partitioning strategy without serving as the sole basis for cluster assignment in every dataset.

The prescribed cluster number is introduced only after hierarchy construction and affinity estimation. Consequently, the same learned structural representation can be retained when different values of $K$ are considered, with model order affecting only the graph-selection and partitioning stage. This separation distinguishes the description of local data structure from the specification of how many clusters are ultimately extracted from that structure.

\subsection{Practical Implications}
\label{subsec:practical_implications}

Retaining the original observations as graph vertices keeps the final clustering result directly at the sample level. Regional structure is used internally to regularize local relations, so no additional assignment from granular-ball representatives back to individual observations is required. The connectivity state of the selected graph also provides a direct indication of how the final partition is formed, distinguishing cases in which connected components already determine the clusters from those that require further spectral subdivision.

Screened refinement reduces the number of complete description-length evaluations required during hierarchy construction, while sparse graph storage avoids explicitly forming a dense $n\times n$ affinity matrix. These choices support the use of the method on datasets beyond the small benchmark setting. The joint scaling experiments, however, show that exact neighborhood computation and repeated high-dimensional local descriptions remain important contributors to runtime and memory consumption as the data size increases.

The prescribed-$K$ formulation is most naturally suited to settings in which the number of clusters is supplied by the task definition or by prior knowledge. Since hierarchy construction and affinity estimation are independent of $K$, the learned structure can also be reused when several candidate model orders are examined. Applications without a known cluster number require an additional mechanism for determining the model order from the learned graph.

\subsection{Limitations and Future Work}
\label{subsec:limitations_future_work}

The real-data results are heterogeneous: MDL-GBTRSC does not attain the highest score on every dataset, and several leading methods remain statistically comparable after correction. The contribution of regional regularization therefore varies with the local geometry represented by the learned hierarchy and affinity graph.

Although hierarchy construction and affinity estimation do not depend on $K$, the final partition still requires a prescribed cluster number. The sensitivity observed under $K-1$ and $K+1$ suggests that the connectivity and spectral structure of the learned graph may contain useful information for model-order determination. Incorporating such information into the final stage would extend the current formulation to settings in which the number of clusters is not known in advance.

The screened split procedure also introduces an approximation between preliminary candidate selection and complete description-length evaluation. The two stages agree closely in the experiments but can produce different refinement paths when their candidate preferences differ. A closer approximation to the complete split criterion could improve this agreement while retaining the computational benefit of screening.

Each terminal ball currently contributes a scalar regional scale to the sample-level affinity function. This compact representation is effective on average, but the dataset-dependent ablation behavior suggests that some local structures may require richer information. Directional, anisotropic, or multi-scale regional descriptors are possible extensions when they can be defined directly from the learned granular-ball structure and incorporated without introducing a separate modeling branch.

The largest computational costs arise when sample size and dimensionality increase together. Exact neighborhood construction, repeated local descriptions, and the final spectral computation all contribute to this growth. Approximate neighborhood search, more efficient local description models, and sparse iterative eigensolvers provide natural directions for improving scalability while retaining the overall sequence of hierarchy construction, regional-to-sample affinity transfer, and connectivity-aware partitioning.

\section{Conclusion}
\label{sec:conclusion}

This paper presented MDL-GBTRSC, which integrates an MDL-guided granular-ball hierarchy with sample-level spectral clustering. Reciprocal-neighborhood continuity contributes to hierarchy refinement, terminal balls provide regional information for affinity construction, and the connectivity of the selected graph guides the final partitioning process. The granular-ball hierarchy thus provides an intermediate structural representation while the original observations remain the vertices of the spectral graph.

Experiments on 18 real and 20 synthetic benchmark datasets show favorable overall performance and provide complementary evidence for the roles of the main components through statistical comparison, graph-state analysis, screened-versus-exhaustive evaluation, ablation studies, prescribed-$K$ analysis, and scalability experiments. The analysis also points to model-order determination, more accurate split screening, richer regional descriptions, and more efficient large-scale graph computation as directions for extending the current framework.
	
\section*{CRediT authorship contribution statement}
\textbf{Zeqiang Xian}: Conceptualization, Methodology, Writing – original draft, Software, Validation. \textbf{Caihui Liu}: Conceptualization, Methodology, Writing – review \& editing, Validation, Supervision. \textbf{Yong Zhang}: Software, Data curation. \textbf{Wenjing Qiu}: Software, Data curation. 

\section*{Declaration of Competing Interest}
The authors declare that they have no known competing financial interests or personal relationships that could have appeared to influence the work reported in this paper.

\section*{Data availability}
Data will be made available on request.

\section*{Acknowledgment}
The research is supported by the National Natural Science Foundation of China under Grant No. 62566003, Graduate Innovation Funding Program of Jiangxi Province under Grant No. YC2025-S224.

\bibliographystyle{cas-model2-names}
\bibliography{References}  

\appendix
\renewcommand*{\theHtable}{app.\arabic{table}}

\section{Dataset-Level Ablation Results}
\label{app:ablation_details}
Table~\ref{tab:ablation_dataset_level} reports the dataset-level ARI changes corresponding to the aggregate ablation results in Section~\ref{subsec:ablation}.

\begin{table*}[htbp]
	\centering
	\caption{Dataset-level ARI change relative to full MDL-GBTRSC for the five ablation variants. Negative values indicate degradation.}
	\label{tab:ablation_dataset_level}
	\scriptsize
	\begin{tabular}{lrrrrr}
		\toprule
		Dataset&\makecell{Without graph\\continuity}&\makecell{Without leaf\\scale}&\makecell{Completed graph\\only}&\makecell{Global partition\\only}&\makecell{Completed graph with\\global partition}\\
		\midrule
		BTissue & +0.015 & +0.005 & +0.000 & +0.000 & +0.000 \\
		Banknote & -0.142 & -0.183 & +0.000 & +0.000 & +0.000 \\
		Bean & +0.003 & -0.026 & +0.000 & +0.000 & +0.000 \\
		Digits & -0.003 & +0.005 & +0.000 & +0.000 & +0.000 \\
		Glass & +0.012 & +0.000 & +0.000 & +0.000 & +0.000 \\
		HTRU2 & +0.005 & -0.004 & +0.000 & +0.000 & +0.000 \\
		Iris & -0.036 & -0.053 & +0.000 & +0.000 & +0.000 \\
		Isolet & -0.000 & -0.000 & +0.000 & +0.000 & +0.000 \\
		Landsat & +0.020 & -0.059 & +0.000 & +0.000 & +0.000 \\
		Libras & +0.000 & +0.000 & +0.000 & +0.000 & +0.000 \\
		PageBlocks & -0.198 & -0.212 & +0.000 & +0.000 & +0.000 \\
		PenDigits & +0.002 & +0.110 & +0.000 & +0.000 & +0.000 \\
		Rice & -0.007 & +0.002 & +0.000 & +0.000 & +0.000 \\
		Seeds & +0.014 & -0.002 & +0.000 & +0.000 & +0.000 \\
		Segment & -0.011 & -0.000 & +0.000 & +0.000 & +0.000 \\
		Thyroid & -0.099 & -0.120 & +0.000 & +0.000 & +0.000 \\
		WDBC & -0.006 & +0.006 & +0.000 & +0.000 & +0.000 \\
		Wine & -0.049 & -0.049 & +0.000 & +0.000 & +0.000 \\
		3-spiral & +0.000 & +0.000 & -0.653 & +0.000 & -0.653 \\
		A & +0.000 & +0.000 & +0.000 & -0.464 & -0.464 \\
		N & +0.000 & +0.000 & +0.000 & -0.147 & -0.147 \\
		aggregation & -0.020 & -0.022 & +0.000 & -0.439 & -0.439 \\
		banana & +0.000 & +0.000 & +0.000 & +0.000 & +0.000 \\
		blobs & +0.010 & -0.019 & +0.000 & +0.000 & +0.000 \\
		chainlink & +0.000 & +0.000 & +0.000 & +0.000 & +0.000 \\
		cluto-t8-8k & +0.003 & -0.008 & +0.000 & +0.000 & +0.000 \\
		complex8 & +0.007 & -0.007 & +0.000 & -0.246 & -0.246 \\
		cure-t2-4k & -0.041 & -0.112 & +0.000 & +0.000 & +0.000 \\
		db2 & +0.000 & +0.000 & -0.775 & -0.503 & -0.775 \\
		diamond9 & -0.001 & -0.001 & +0.000 & +0.000 & +0.000 \\
		fourty & +0.000 & +0.000 & +0.000 & -0.383 & -0.383 \\
		jain & +0.000 & +0.000 & -0.485 & +0.000 & -0.485 \\
		target & +0.000 & +0.000 & -0.669 & -0.306 & -0.669 \\
		wingnut & -0.123 & -0.098 & +0.000 & +0.000 & +0.000 \\
		zelnik1 & +0.000 & +0.000 & +0.000 & -0.144 & -0.517 \\
		zelnik2 & +0.010 & +0.010 & +0.000 & +0.000 & +0.000 \\
		zelnik4 & +0.000 & +0.000 & +0.000 & +0.000 & +0.000 \\
		zelnik6 & +0.000 & +0.000 & -0.032 & +0.000 & -0.032 \\
		\bottomrule
	\end{tabular}
\end{table*}

\section{Synthetic Clustering Visualizations of the Compared Methods}
\label{app:synthetic_visualizations}

Figures~\ref{fig:app_syn_stscfk}--\ref{fig:app_syn_gmmsc} show the partitions produced by the nine competing methods on the same 20 synthetic datasets used in the main experiments. Deterministic methods are shown by their unique outputs, while stochastic methods use the seed-0 run for visualization; the quantitative results in Table~\ref{tab:synthetic_ari_nmi_comparison} remain the 20-run mean $\pm$ sample standard deviation. The same plotting settings are used for a given dataset across methods.

\newcommand{\syntheticvisgrid}[1]{%
	\setlength{\subfigcapskip}{0pt}%
	\setlength{\subfigtopskip}{0pt}%
	\setlength{\subfigbottomskip}{0pt}%
	\setlength{\subfigcapmargin}{0pt}%
	\subfigure[zelnik6]{\includegraphics[width=0.15\textwidth]{zelnik6-#1.pdf}}
	\subfigure[zelnik1]{\includegraphics[width=0.15\textwidth]{zelnik1-#1.pdf}}
	\subfigure[blobs]{\includegraphics[width=0.15\textwidth]{blobs-#1.pdf}}
	\subfigure[zelnik2]{\includegraphics[width=0.15\textwidth]{zelnik2-#1.pdf}}
	\subfigure[3-spiral]{\includegraphics[width=0.15\textwidth]{3-spiral-#1.pdf}}
	
	\subfigure[db2]{\includegraphics[width=0.15\textwidth]{db2-#1.pdf}}
	\subfigure[jain]{\includegraphics[width=0.15\textwidth]{jain-#1.pdf}}
	\subfigure[zelnik4]{\includegraphics[width=0.15\textwidth]{zelnik4-#1.pdf}}
	\subfigure[target]{\includegraphics[width=0.15\textwidth]{target-#1.pdf}}
	\subfigure[aggregation]{\includegraphics[width=0.15\textwidth]{aggregation-#1.pdf}}
	
	\subfigure[chainlink]{\includegraphics[width=0.15\textwidth]{chainlink-#1.pdf}}
	\subfigure[fourty]{\includegraphics[width=0.15\textwidth]{fourty-#1.pdf}}
	\subfigure[wingnut]{\includegraphics[width=0.15\textwidth]{wingnut-#1.pdf}}
	\subfigure[N]{\includegraphics[width=0.15\textwidth]{N-#1.pdf}}
	\subfigure[A]{\includegraphics[width=0.15\textwidth]{A-#1.pdf}}
	
	\subfigure[complex8]{\includegraphics[width=0.15\textwidth]{complex8-#1.pdf}}
	\subfigure[diamond9]{\includegraphics[width=0.15\textwidth]{diamond9-#1.pdf}}
	\subfigure[cure-t2-4k]{\includegraphics[width=0.15\textwidth]{cure-t2-4k-#1.pdf}}
	\subfigure[banana]{\includegraphics[width=0.15\textwidth]{banana-#1.pdf}}
	\subfigure[cluto-t8-8k]{\includegraphics[width=0.15\textwidth]{cluto-t8-8k-#1.pdf}}
}

\newcommand{\appendixsyntheticfigure}[3]{%
	\begin{figure}[p]
		\centering
		\syntheticvisgrid{#1}
		\caption{Clustering partitions produced by #2 on the 20 synthetic benchmark datasets.}
		\label{fig:app_syn_#3}
	\end{figure}
}

\appendixsyntheticfigure{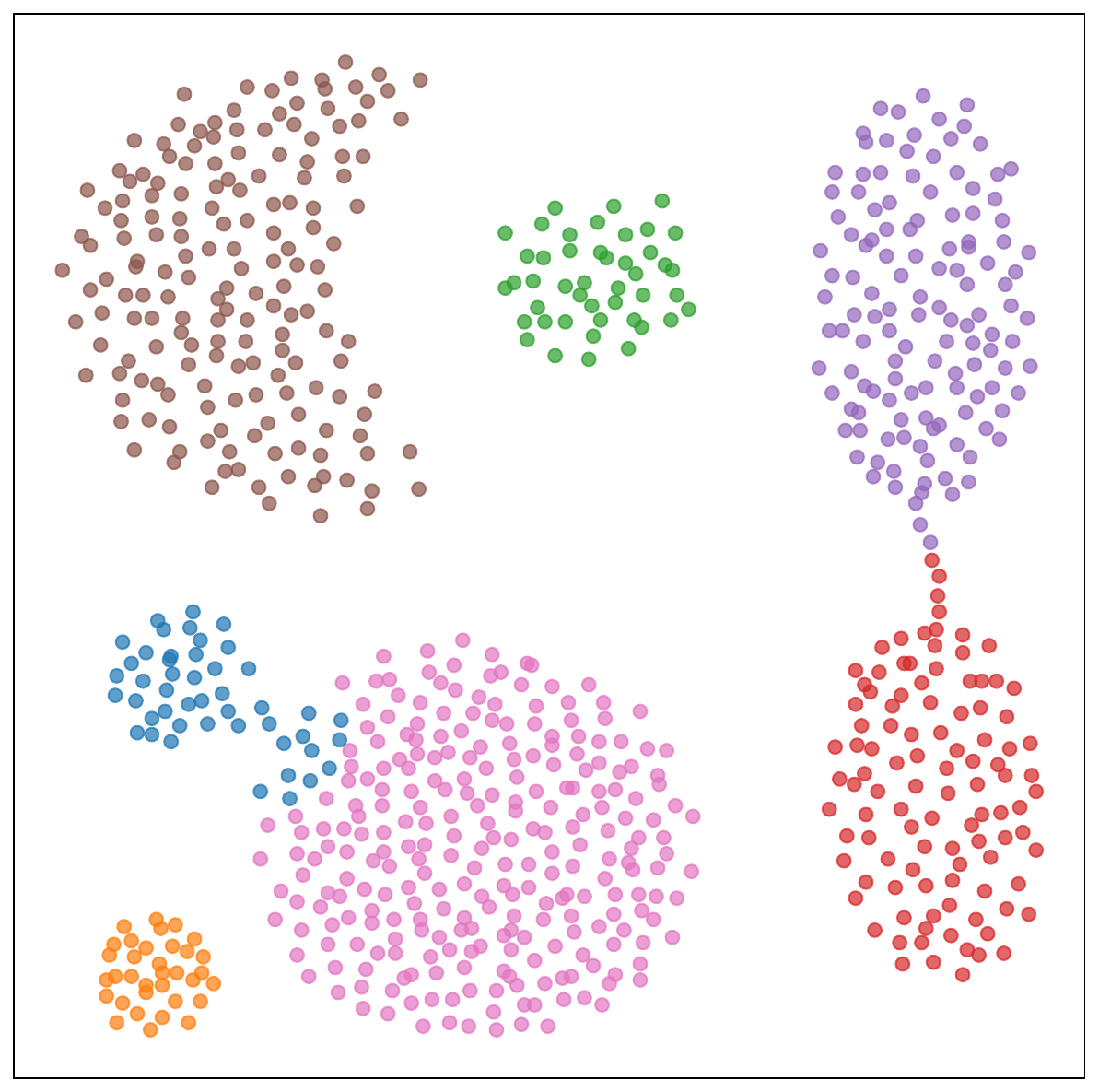}{STSC-FK}{stscfk}
\appendixsyntheticfigure{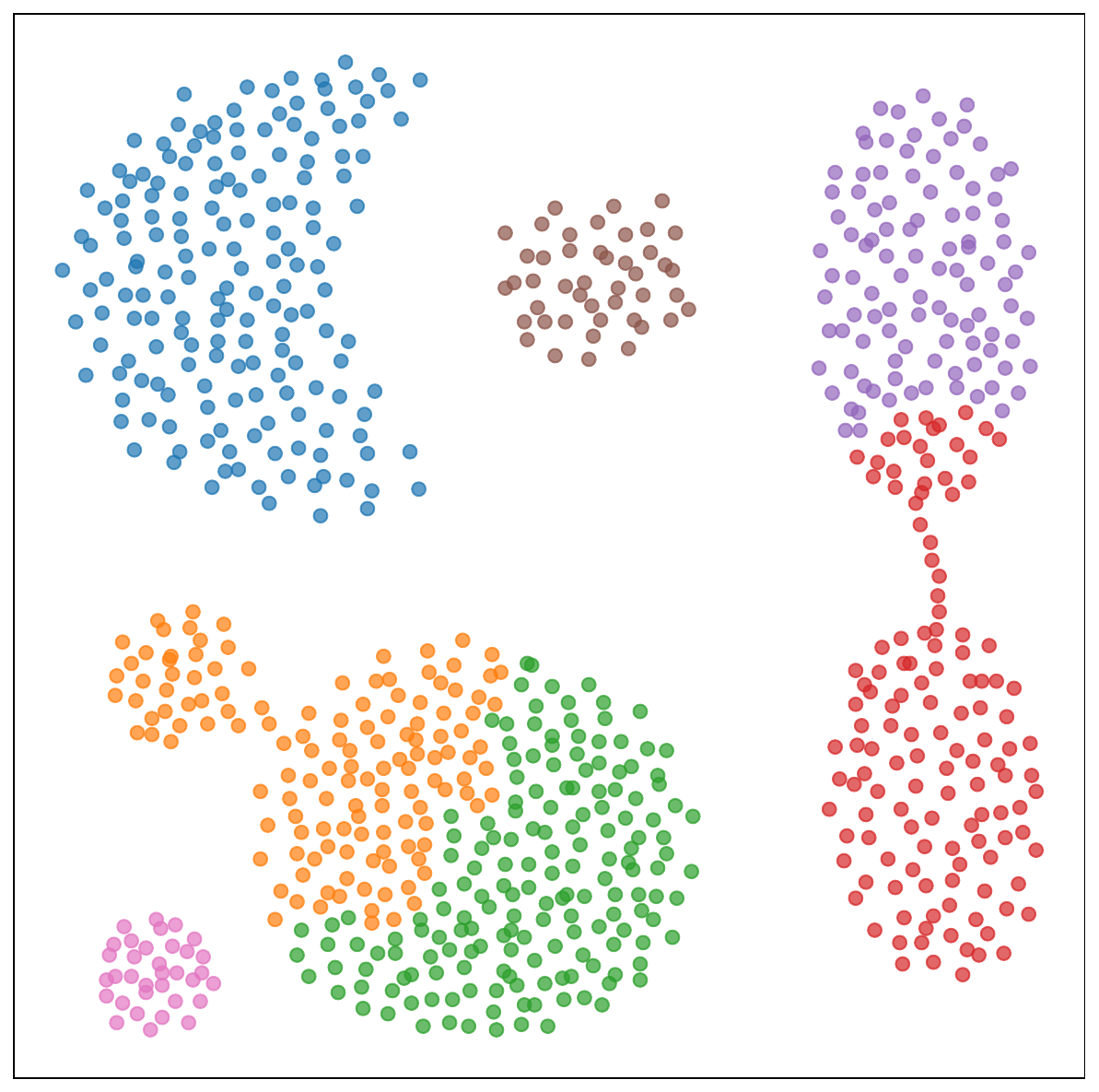}{N-DSC}{ndsc}
\appendixsyntheticfigure{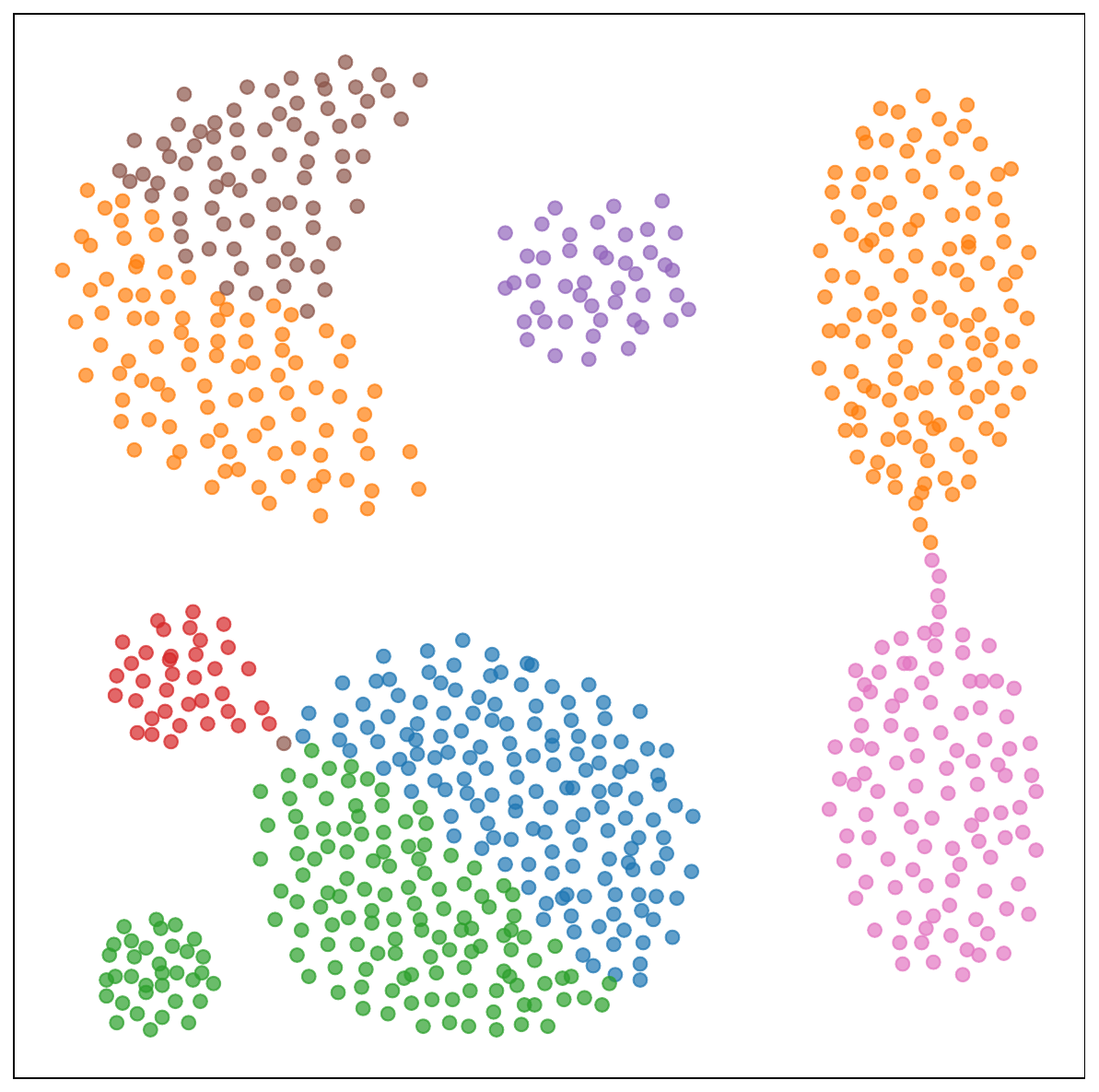}{FastSC}{fastsc}
\appendixsyntheticfigure{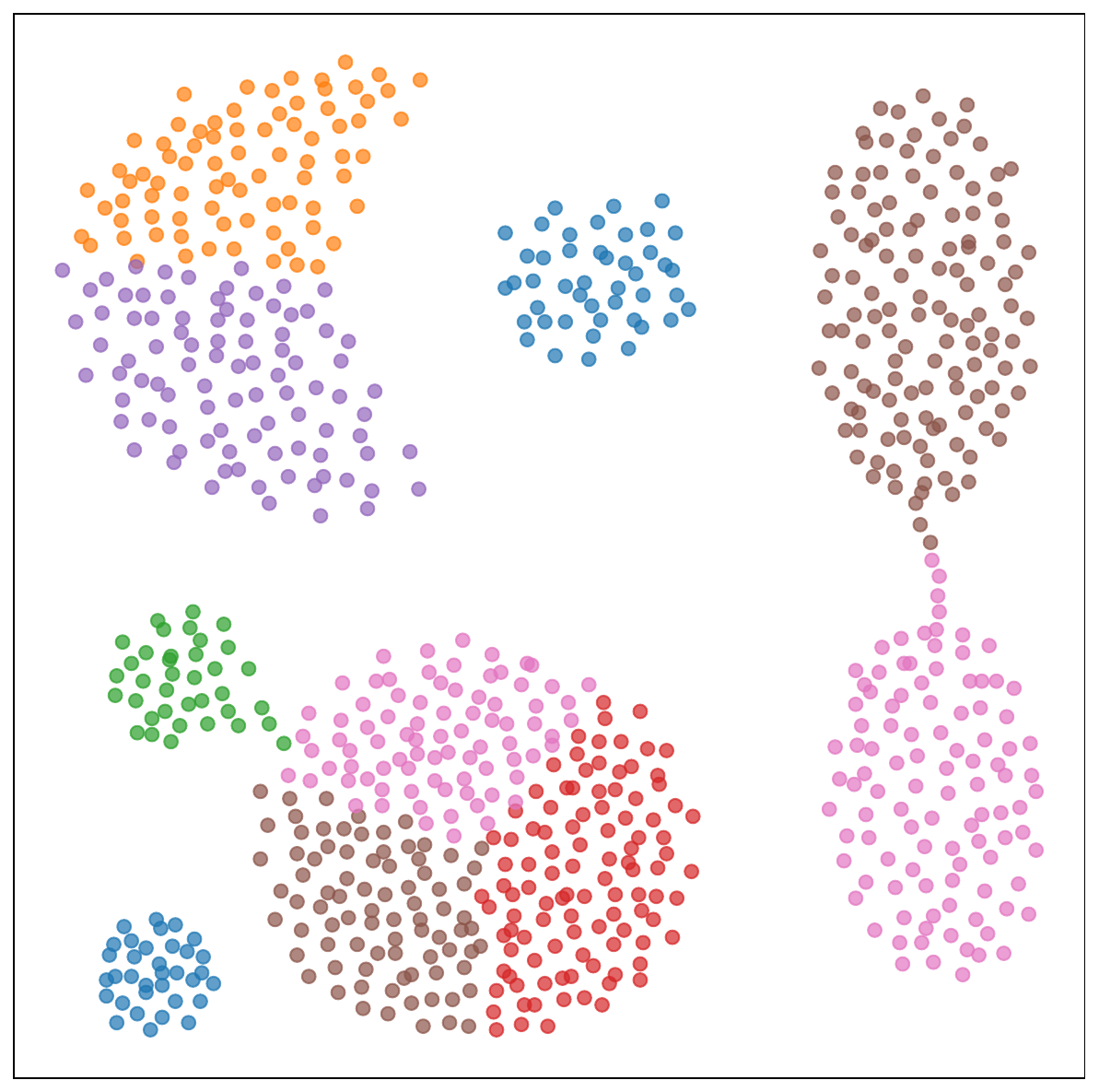}{SC-kNN}{scknn}
\appendixsyntheticfigure{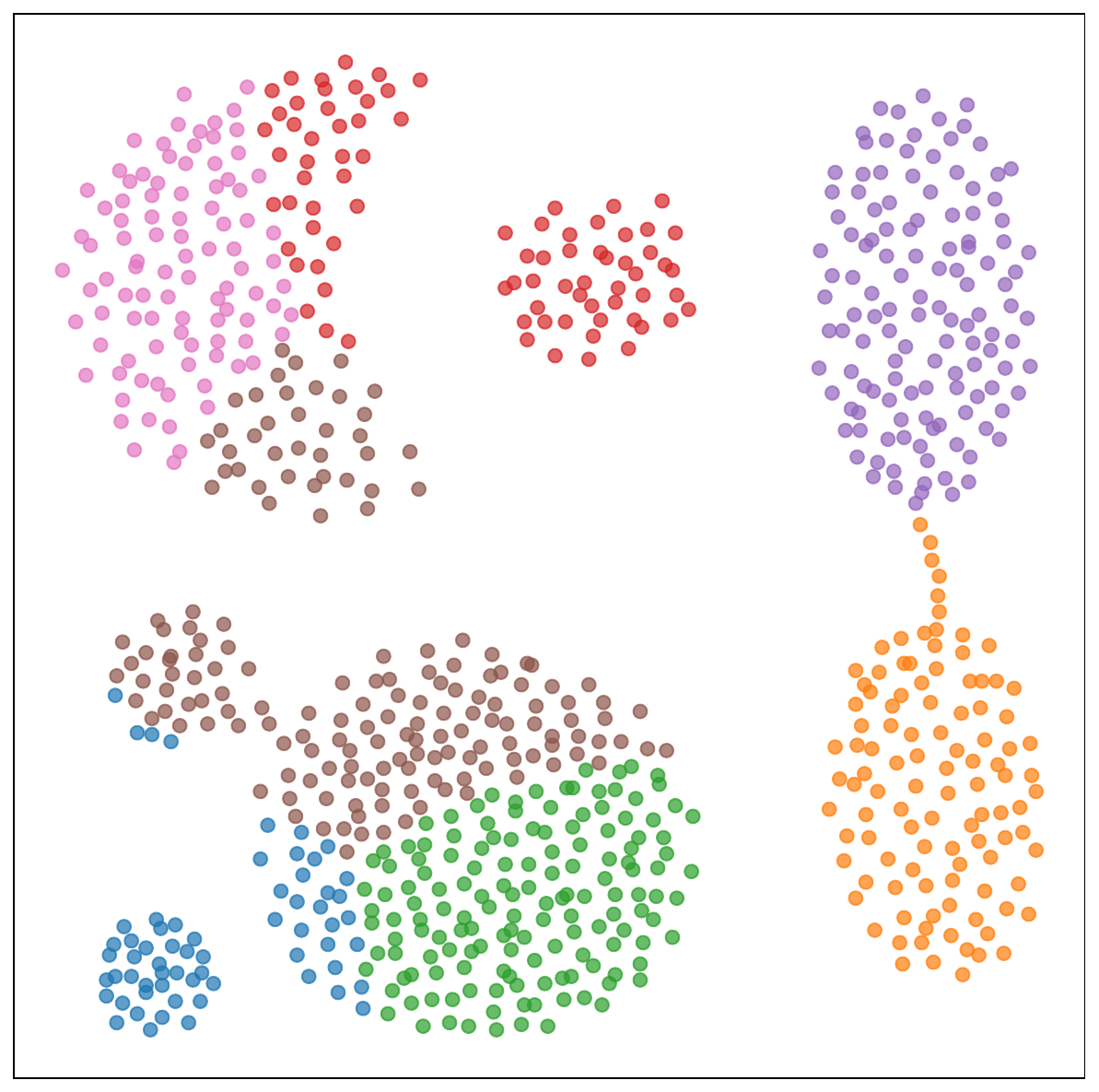}{SC-RBF}{scrbf}
\appendixsyntheticfigure{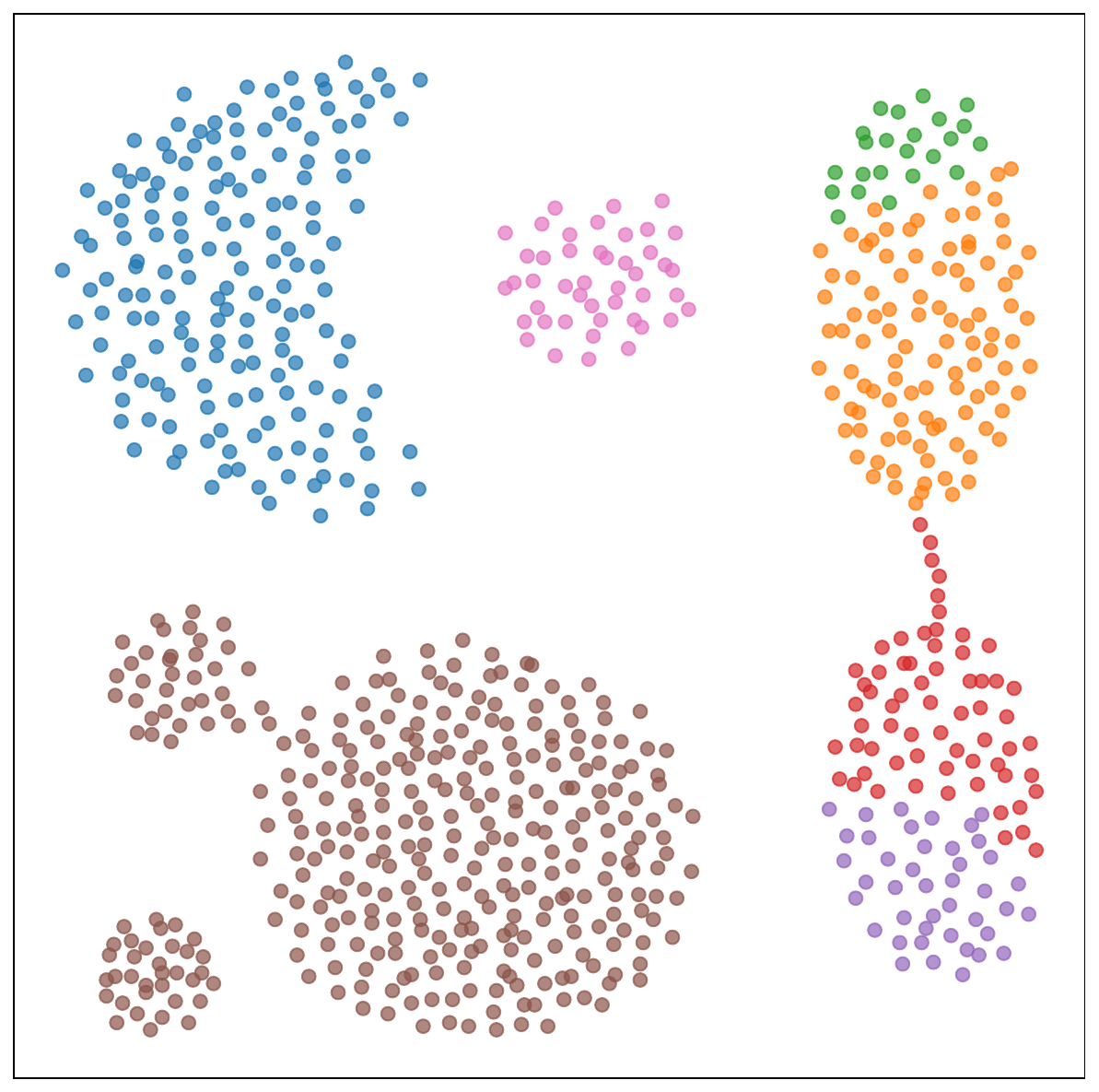}{GBCT}{gbct}
\appendixsyntheticfigure{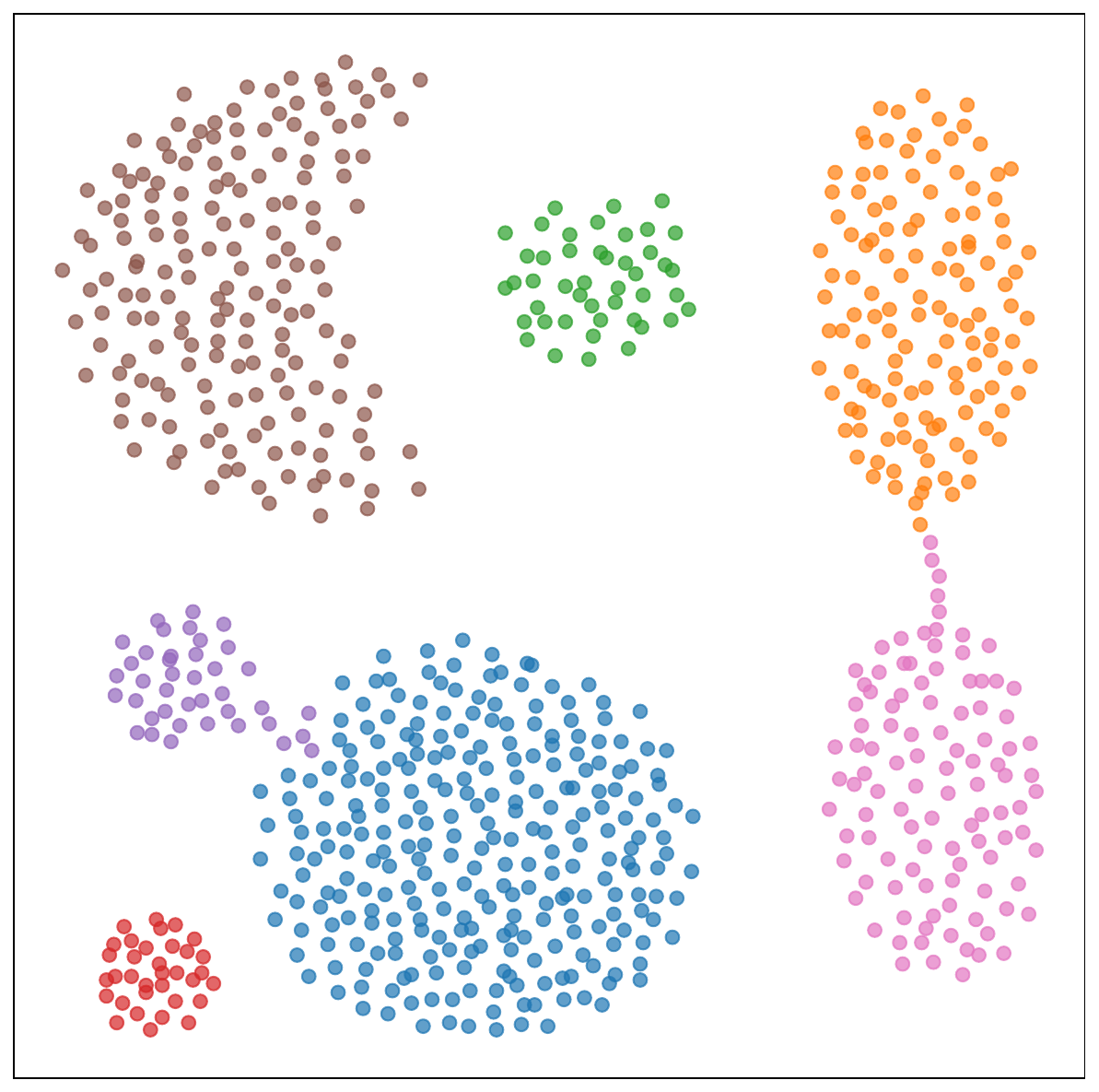}{MDMSC}{mdmsc}
\appendixsyntheticfigure{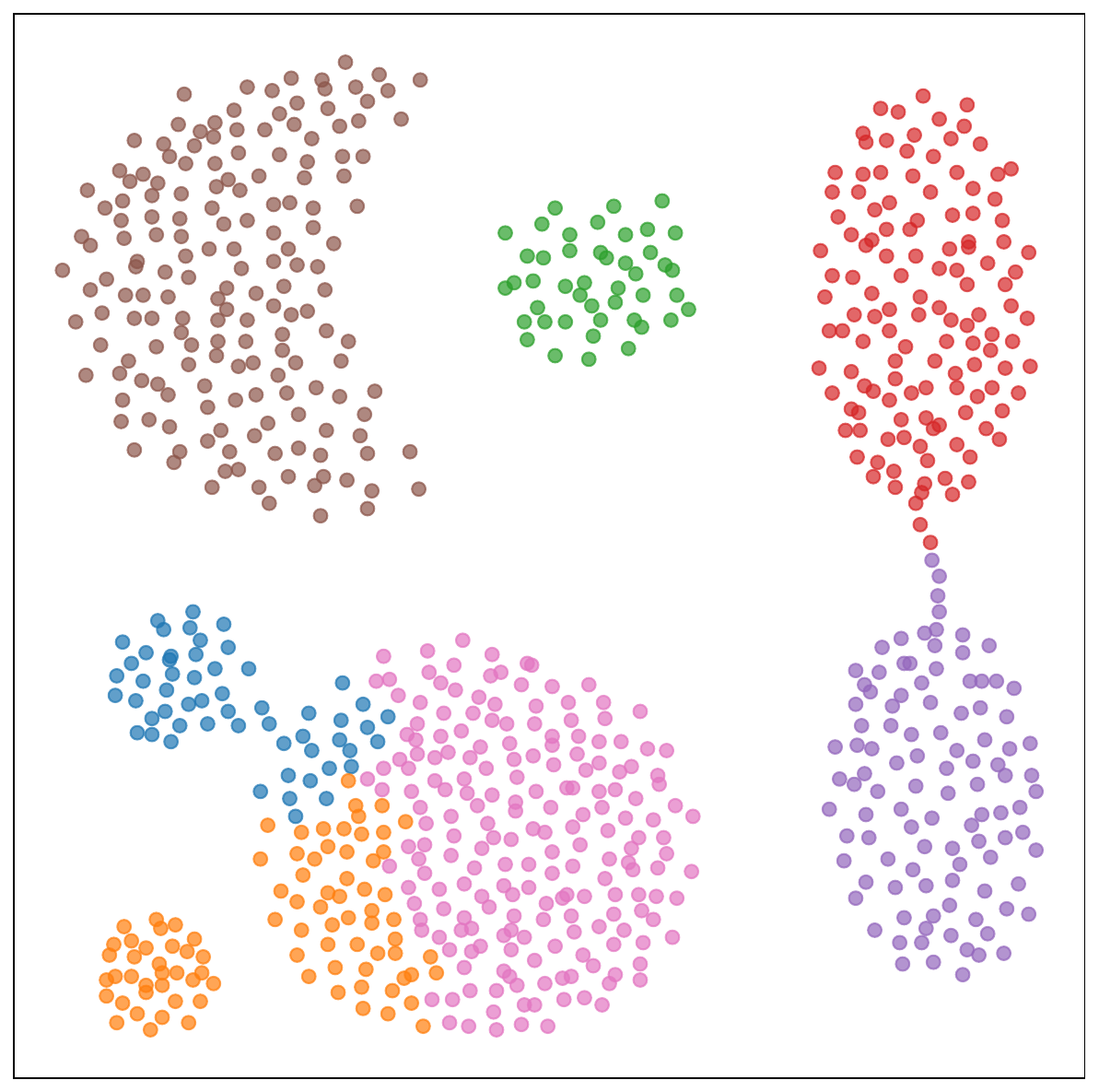}{GBSC}{gbsc}
\appendixsyntheticfigure{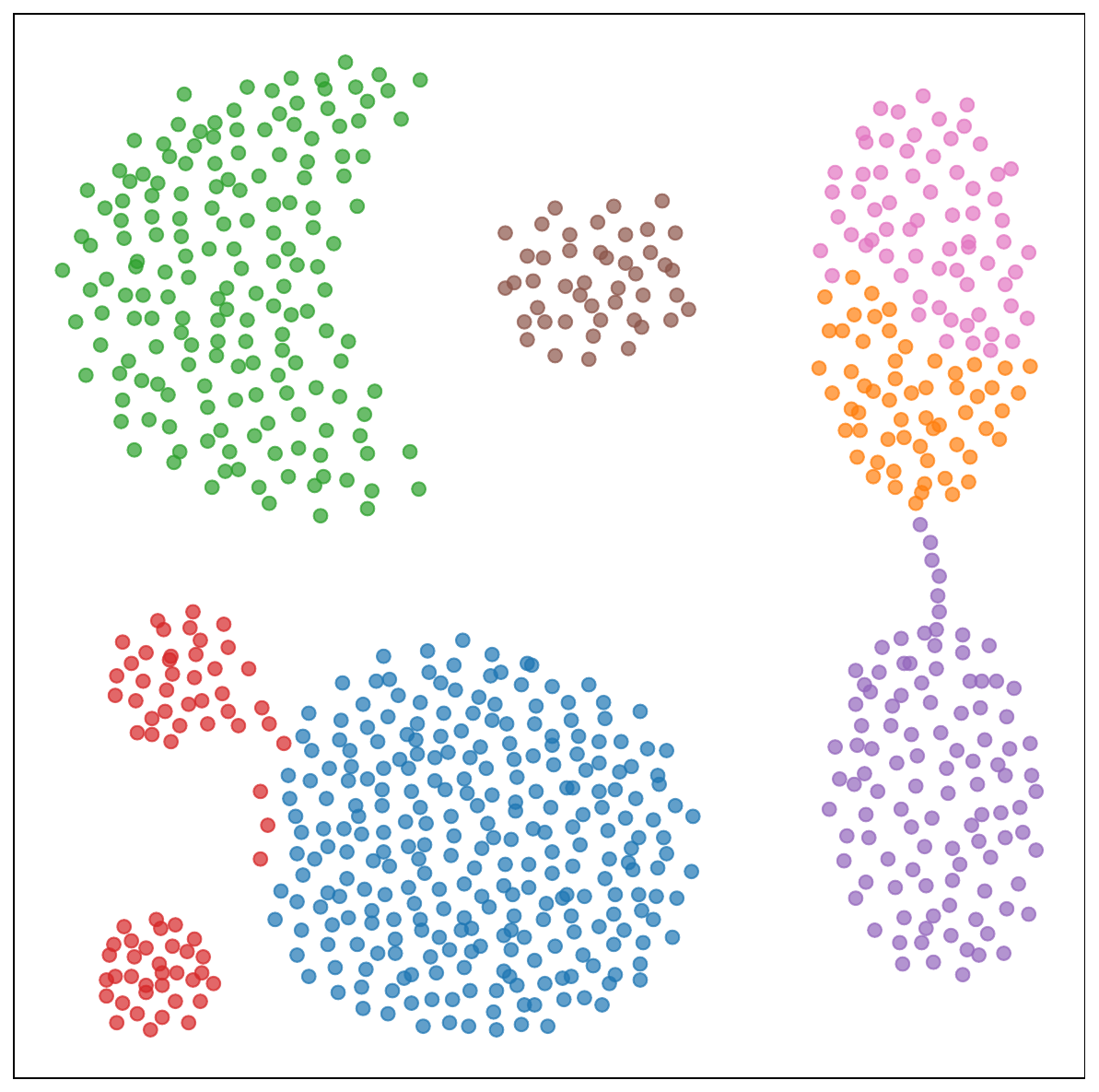}{GMM-SC}{gmmsc}

\end{document}